\documentclass{article}
\pdfoutput=1
\usepackage[preprint]{neurips_2023}
\usepackage[utf8]{inputenc} %
\usepackage[T1]{fontenc}    %
\usepackage{xurl}
\usepackage{hyperref}            %
\usepackage{breakurl}          %
\usepackage{amsfonts}       %
\usepackage{amsmath}       %
\usepackage{nicefrac}       %
\usepackage{microtype}      %
\usepackage{lipsum}
\usepackage{graphicx}
\usepackage{wrapfig}
\usepackage[size=small]{caption}
\usepackage{mathtools}
\usepackage{amssymb}
\usepackage{amsthm}
\usepackage[algoruled,boxed,lined,noend]{algorithm2e}
\usepackage{adjustbox} 
\usepackage{float}
\usepackage{amssymb}
\usepackage{pifont}
\usepackage{subcaption}
\usepackage{fontawesome}
\usepackage{multirow}
\usepackage{rotating}
\usepackage{array}    
\usepackage{lineno}
\usepackage{longtable}
\usepackage[table,dvipsnames]{xcolor}
\usepackage[numbers]{natbib}
\usepackage{textcomp}
\usepackage{ulem}
\usepackage{color}

\usepackage[dvipsnames]{xcolor}
\usepackage{booktabs}
\usepackage{multirow}
\usepackage{tabularx}

\definecolor{cvprblue}{rgb}{0.21,0.49,0.74}
\definecolor{citecolor}{HTML}{0071BC}
\definecolor{linkcolor}{HTML}{ED1C24}
\usepackage[capitalize]{cleveref}
\crefname{section}{Sec.}{Secs.}
\crefname{table}{Table}{Tables}
\crefname{figure}{Fig.}{Figs.}
\usepackage[most]{tcolorbox}
\usepackage{float}
\usepackage{xspace}
\tcbset{
  aibox/.style={
    width=394.18663pt,
    top=10pt,
    colback=white,
    colframe=black,
    colbacktitle=black,
    enhanced,
    center,
    attach boxed title to top left={yshift=-0.1in,xshift=0.15in},
    boxed title style={boxrule=0pt,colframe=white,},
  }
}
\newtcolorbox{AIbox}[2][]{aibox,title=#2,#1}

\usepackage{colortbl}
\definecolor{qual-fig-green}{RGB}{0,144,11}
\definecolor{qual-fig-red}{RGB}{238,0,0}
\definecolor{qual-fig-purple}{RGB}{153,51,255}

\title{Unifying Biomedical Vision-Language Expertise: Towards a Generalist Foundation Model via Multi-CLIP Knowledge Distillation}

\author{
Shansong Wang$^{1}$  \quad Zhecheng Jin$^{2}$  \quad Mingzhe Hu$^{1,3}$ \quad Mojtaba Safari$^{1}$  \quad Feng Zhao$^{4}$\\  \quad \textbf{Chih-Wei Chang}$^{1}$ \quad \textbf{Richard LJ Qiu$^{1}$}  \quad \textbf{Justin Roper}$^{1}$ \quad \textbf{David S. Yu}$^{1}$\quad \\ \textbf{Xiaofeng Yang}$\textsuperscript{1,2,3 \faEnvelope}$ 
\vspace{3mm}
\\
$^1$Department of Radiation Oncology, Winship Cancer Institute, Emory University School of Medicine\\
$^2$Department of Biomedical Engineering, College of Engineering, Georgia Institute of Technology\\
$^3$Department of Computer Science and Mathematics, Laney Graduate School, Emory University\\
$^4$School of Electrical and Computer Engineering, College of Engineering, Georgia Institute of Technology\\
\faEnvelope \quad Corresponding author: xiaofeng.yang@emory.edu \\
}

\begin{document}
\maketitle
\begin{abstract}
CLIP models pretrained on natural images with billion-scale image-text pairs have demonstrated impressive capabilities in zero-shot classification, cross-modal retrieval, and open-ended visual answering. However, transferring this success to biomedicine is hindered by the scarcity of large-scale biomedical image-text corpora, the heterogeneity of image modalities, and fragmented data standards across institutions. These limitations hinder the development of a unified and generalizable biomedical foundation model trained from scratch. To overcome this, we introduce MMKD-CLIP, a generalist biomedical foundation model developed via Multiple Medical CLIP Knowledge Distillation. Rather than relying on billion-scale raw data, MMKD-CLIP distills knowledge from nine state-of-the-art domain-specific or generalist biomedical CLIP models, each pretrained on millions of biomedical image-text pairs. Our two-stage training pipeline first performs CLIP-style pretraining on over 2.9 million biomedical image-text pairs from 26 image modalities, followed by feature-level distillation using over 19.2 million feature pairs extracted from teacher models. We evaluate MMKD-CLIP on 58 diverse biomedical datasets, encompassing over 10.8 million biomedical images across nine image modalities. The evaluation spans six core task types: zero-shot classification, linear probing, cross-modal retrieval, visual question answering, survival prediction, and cancer diagnosis. MMKD-CLIP consistently outperforms all teacher models while demonstrating remarkable robustness and generalization across image domains and task settings. These results underscore that multi-teacher knowledge distillation is a scalable and effective paradigm for building high-performing biomedical foundation models under the practical constraints of real-world data availability.

\end{abstract}
\section{Introduction}
\label{sec:intro}

\begin{figure}[!t] 
	\centering
	\includegraphics[width=0.98\textwidth]{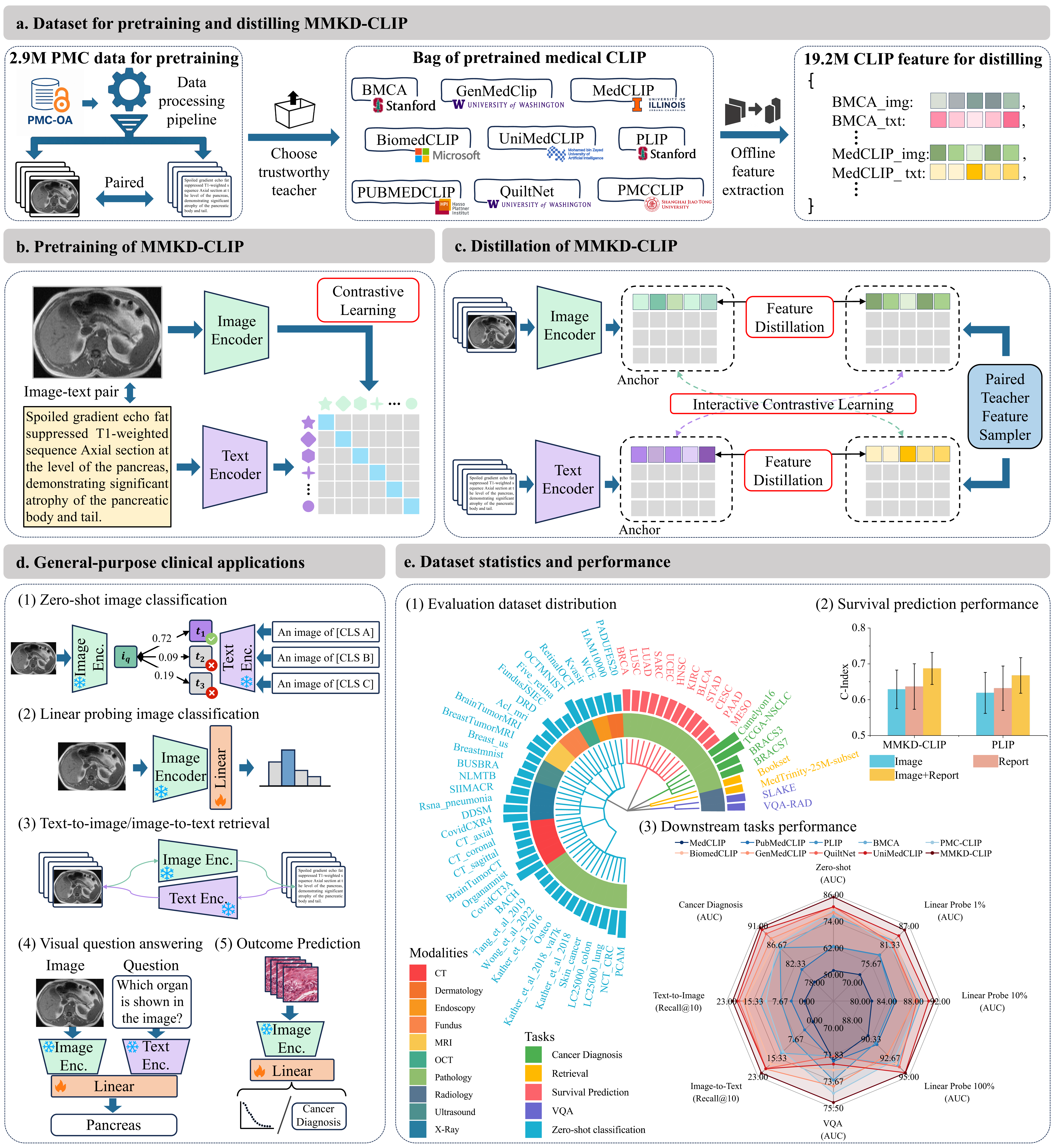}
	\caption{\textbf{Overview of MMKD-CLIP dataset curation, model training, and evaluation.} a. Dataset assembly for pretraining and distillation. We extracted over 2.9 million paired image-text data from the PMC-OA database. Nine pretrained medical CLIP teacher models including BMCA, GenMedCLIP, MedCLIP, BiomedCLIP, UniMedCLIP, PLIP, PUBMEDCLIP, and QuiltNet supplied over 19.2 million image and text feature vectors for student distillation. b. Pretraining of MMKD-CLIP based on contrastive learning. Each raw image-text pair is encoded via separate image and text  encoder. Representations are aligned by standard contrastive learning to establish an initial multimodal embedding space. c. Knowledge distillation pipeline. The student model is trained using both feature distillation (FD) loss and interactive contrastive learning (ICL) loss. The optimization details of the two losses are shown in section \ref{44}. d. General and clinical applications. Post-training, MMKD-CLIP supports (1) zero-shot image classification, (2) linear probing image classification, (3) text-to-image/image-to-text retrieval, (4) visual question answering and (5) outcome prediction. e. Evaluation dataset distribution and downstream performance. Concentric rings depict modality coverage across 58 benchmark datasets spanning CT, MRI, ultrasound, pathology, dermatology, fundus, OCT, endoscopy, radiography and X-ray. Radar chart and bar charts compare MMKD-CLIP against state-of-the-art vision–language models on different metrics.}
	\label{fig:overview} 
\end{figure}

Foundation models trained with large-scale vision-language data have achieved remarkable progress in general-domain applications, enabling robust zero-shot classification, cross-modal retrieval, and open-ended visual question answering (VQA)~\cite{hafner2021clip, radford2021learning, Liu_2024_CVPR, moor2023foundation,li2022blip}. In parallel, the medical AI community has adapted these foundation models into domain-specific CLIP-style~\cite{radford2021learning} architectures to effectively learn visual-semantic correspondences across radiology~\cite{you2023cxr, blankemeier2024merlin, lipkova2022artificial, wang2025triad}, pathology~\cite{xu2024whole, huang2023visual}, ophthalmology~\cite{zhou2023foundation, hu2024ophclip}, and other medical specialties. These biomedical vision–language models (VLMs) demonstrate promising capabilities in tasks such as disease classification, outcome prediction, and visual question answering~\cite{moor2023foundation, kim2024transparent}. However, current biomedical CLIP models are usually designed for specific medical domains~\cite{zhang2024generalist}. As a result, these models often struggle to generalize effectively when applied to data from other medical specialties or different image modalities~\cite{zhang2024generalist}.

Real-world clinical reasoning, in practice, rarely confines itself to isolated specialties~\cite{qiu2025quantifying,Safari_2024,10.1088/1361-6560/ade049}. Physicians integrate multimodal information from diverse sources, including radiographs, microscopic slides, endoscopy, and even molecular-level imaging, while grounding their interpretations in clinical notes, guidelines, and broader biomedical knowledge~\cite{yang2025multi,duan2024deep}. This inherently interdisciplinary nature of medicine calls for a generalist biomedical foundation model that can unify visual and textual concepts across multiple specialties. However, developing such models demands the collection of large-scale multimodal datasets from diverse sources. This effort is complicated by inconsistencies in data management standards across institutions, as well as the fragmented nature of available biomedical image–text corpora~\cite{simon2024future}. These challenges significantly hinder the effective training of comprehensive biomedical foundation models. Although several efforts have collected biomedical image-text pairs from the PubMed Central Open Access (PMC-OA) database~\cite{pmc_open_access}, yielding datasets ranging from approximately 1.6 to 6.2 million pairs~\cite{lin2023pmc,lozano2025biomedica,nie2025conceptclip}, these numbers remain substantially lower than the billions of image-text pairs typically utilized to train natural image domain vision-language models~\cite{schuhmann2022laion}. This restricts their ability to capture the full semantic richness and modality diversity of biomedical data. 

In natural image domains, distilling knowledge from a single teacher CLIP model pretrained on billions of image-text pairs into a compact student model has demonstrated performance comparable to the original model~\cite{yang2024clip,dai2022enabling}. Inspired by this strategy, we propose training a student biomedical CLIP model by distilling knowledge from multiple domain-specific/general CLIP models, each pretrained individually on millions of biomedical image-text pairs. This approach aims to effectively address the generalization challenges inherent in biomedical CLIP, thereby enabling the student model to robustly integrate multimodal information across diverse medical specialties and imaging modalities.

Here, we present a generalist biomedical foundation model based on \textbf{M}ultiple \textbf{M}edical CLIP \textbf{K}nowledge \textbf{D}istillation and vision-language pre-training, named MMKD-CLIP. As shown in Fig. \ref{fig:overview} a b, and c, MMKD-CLIP is a two-stage training pipeline. In the first stage, MMKD-CLIP undergoes CLIP-style pretraining on 2,911,190 biomedical image-text pairs spanning 26 imaging modalities, collected from the PMC-OA database. In the second stage, we select nine state-of-the-art domain-specific or generalist biomedical CLIP models as teacher models. We use these teachers to extract features offline from the original 2.9 million image-text pairs, resulting in a comprehensive set of 19,229,852 image-text feature pairs. These feature pairs are then utilized for knowledge distillation into a single student model, enabling the student model to effectively leverage the specialized expertise of each teacher model and generalize robustly across diverse biomedical imaging modalities. As shown in Fig. \ref{fig:overview} d and e, we conduct extensive evaluations across a broad range of downstream tasks, encompassing 6 task types and 9 imaging modalities, spanning 58 datasets and a total of 10,808,657 images. Specifically, our evaluation includes 38 datasets for medical image classification, incorporating both zero-shot and linear probe settings; 2 datasets for cross-modal retrieval; 2 datasets for VQA; 12 datasets for survival prediction; and 4 datasets for cancer diagnosis. The results show that our distilled student model consistently outperforms all state-of-the-art teacher models across the full range of evaluated tasks. More importantly, the MMKD-CLIP pipeline is designed to continually distill knowledge from any future state-of-the-art biomedical CLIP models developed by different institutions, enabling the student model to remain up-to-date and at the forefront of performance.

Our main contributions are as follows:
\begin{itemize}
	\item We introduce a large-scale multimodal distillation corpus comprising 19.2 million image-text feature pairs. This corpus is generated by applying teacher CLIP models to the PMC-OA dataset and enables efficient knowledge transfer without the need for direct access to large-scale raw data or labels.
	\item We propose MMKD-CLIP, the first biomedical vision-language foundation model that combines large-scale pretraining with multi-teacher knowledge distillation across domains. This design enables the student model to integrate specialized knowledge from diverse medical fields and achieve robust cross-modal generalization.
	\item We conduct extensive evaluations of MMKD-CLIP across 58 benchmark datasets covering 6 task types and 9 imaging modalities, totaling 10.8 million images. These tasks include classification, cross-modal retrieval, VQA, survival prediction, and cancer diagnosis. MMKD-CLIP consistently outperforms all individual teacher models across the board.
	\item We design MMKD-CLIP as an open and extensible distillation framework that supports continual integration of future biomedical CLIP models. This ensures that the student model can remain up-to-date with the latest advancements from diverse institutions, promoting long-term scalability, adaptability, and generalization in real-world clinical applications.
\end{itemize}

\section{Results}
\subsection{Curating the first largest-scale biomedical offline distillation dataset}

The training pipeline of MMKD-CLIP consists of two stages: pretraining and offline knowledge distillation. In the pretraining stage, we directly used the ``Concept-Filtering'' subset of the BIOMEDICA dataset~\cite{lozano2025biomedica}, which contains about 6 million image-text pairs. To further clean the dataset, we apply an additional filtering step that retains only those pairs annotated with an ``image\_primary\_label'' of ``Microscopy'' or ``Clinical Imaging.'' In the end, we obtained a total of 2,911,190 image-text pairs, including 1,416,257 single-panel images and 1,494,933 multi-panel images. In Table \ref{tab:modality_distribution}, we summarize the distribution of image-text pairs across 26 imaging modalities represented in the filtered subset.

In the offline knowledge distillation stage, we constructed the largest biomedical offline knowledge distillation dataset to date, which contains 19,229,852 (``image''–``text''–``teacher image feature''–``teacher text feature'') quadruplets. The extraction pipeline and rules for these quadruplets are shown in Fig. \ref{fig:choosteacher} b and section \ref{mdp} and \ref{mtda}. This dataset covers 26 imaging modalities and contains the distribution of 9 state-of-the-art biomedical CLIPs. By implementing offline knowledge distillation on this dataset, we can learn the joint distribution across modalities and CLIPs, thereby achieving robust generalization effects.

\subsection{Extensive evaluation and state-of-the-art model comparison}

We present results from 58 downstream datasets across 6 types of evaluation tasks and 9 modality settings. Fig. \ref{fig:overview} d shows the distribution of downstream dataset and the overall performance of each evaluation task. Specifically, we present quantitative results on zero-shot biomedical image classification (Fig. \ref{fig:f1}), linear probing image classification (Fig. \ref{fig:f3} a), cross-modal retrieval (Fig. \ref{fig:f3} b), VQA (Fig. \ref{fig:f4}), survival prediction (Fig. \ref{fig:f5}), and supervised cancer diagnosis tasks (Fig. \ref{fig:f6} a). Finally, we conduct ablation studies to demonstrate the necessity of pipeline combinations and loss selections (Fig. \ref{fig:f6} b). In addition, since MMKD-CLIP is distilled from 9 renowned and state-of-the-art biomedical CLIP models, we select these 9 models for comparison, which include BMCA~\cite{lozano2025biomedica}, GenMedClip~\cite{ikezogwo2025medicalnarratives}, MedCLIP~\cite{wang2022medclip}, BiomedCLIP~\cite{zhang2025multimodal}, UniMedCLIP~\cite{khattak2024unimed}, PLIP~\cite{huang2023visual}, PubMedCLIP~\cite{eslami2023pubmedclip}, QuiltNet~\cite{ikezogwo2023quilt}, and PMC-CLIP~\cite{lin2023pmc}. These 9 models are described in detail in Section \ref{sota}.

\subsubsection{Zero-shot biomedical image classification}

\begin{figure}[!t] 
	\centering
	\includegraphics[width=0.94\textwidth]{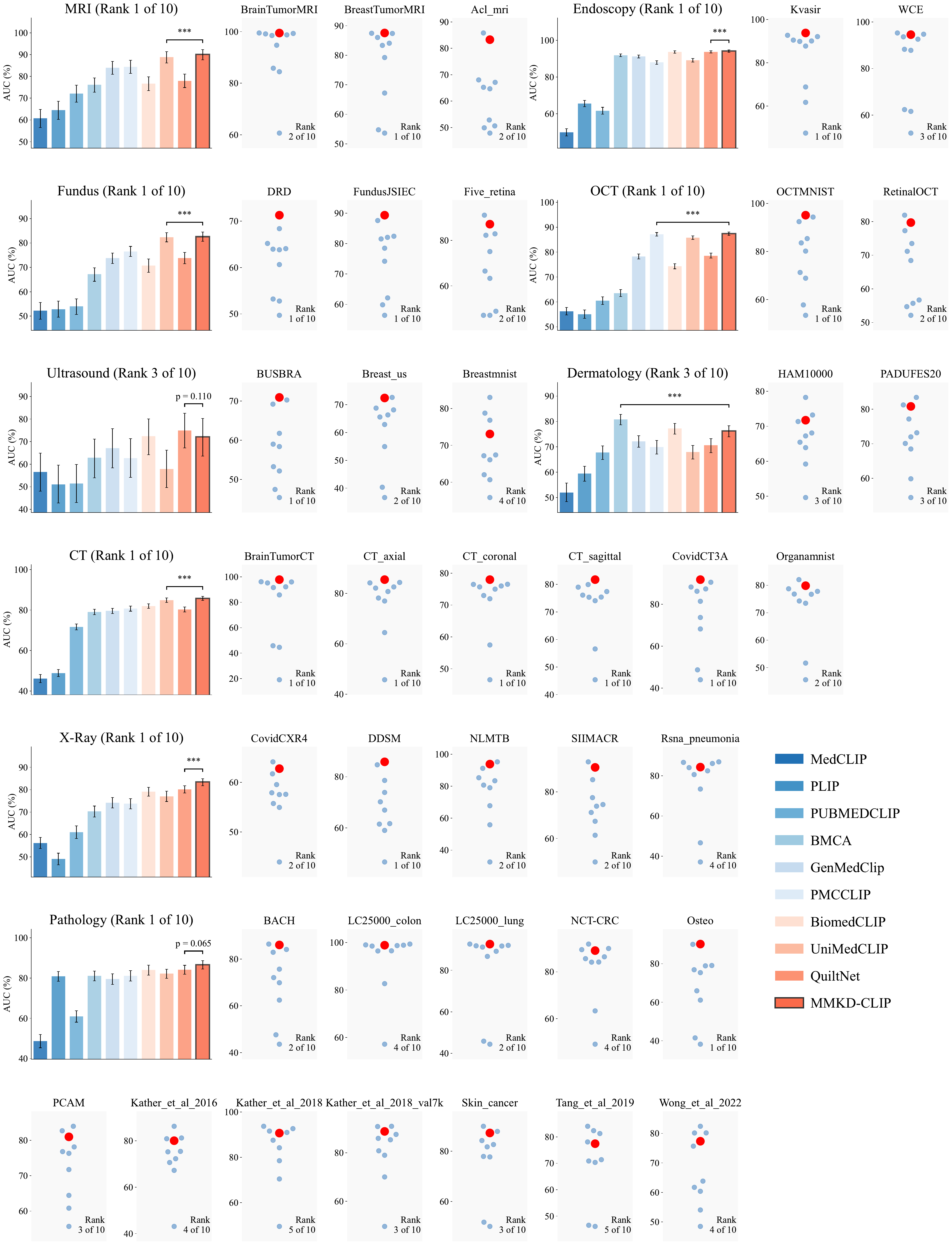}
	\caption{\textbf{Zero-shot classification performance of MMKD-CLIP across nine medical imaging modalities.} Bar charts show the mean AUC and 95\% CI, which estimated using the bootstrap method (n = 1,000 replicates). for each modality (MRI, endoscopy, fundus, OCT, ultrasound, dermatology, CT, X-ray, and pathology). Each scatter plot  visualizes the rank position (1-10) of MMKD-CLIP on individual datasets within that modality. A two-sided Mann-Whitney U test was performed to evaluate whether MMKD-CLIP’s zero-shot AUC differs significantly from the best competing model. Significance is indicated as ***: p < 0.001; **: p < 0.01; *: p < 0.05.}
	\label{fig:f1} 
\end{figure}

We first constructed a zero-shot image classification task to systematically evaluate MMKD-CLIP, specifically involving 9 medical imaging modalities and covering a total of 38 datasets (Fig.~\ref{fig:f1}). The introduction of these 38 datasets is shown in Tables~\ref{zerobeggin} to \ref{zeroend} in the Supplementary Material. The results consistently demonstrate the superior diagnostic performance and generalization capability of MMKD-CLIP across diverse downstream zero-shot image classification tasks. The quantitative results are in Table \ref{tab:Zero-shot classification} in the Supplementary Materials.

MMKD-CLIP achieves the highest area under the receiver operating characteristic curve (AUC) in 7 out of 9 imaging modalities (Magnetic Resonance Imaging (MRI), Fundus, Optical Coherence Tomography (OCT), Computed Tomography (CT), X-ray, Endoscopy, and Pathology), and ranks within the top 3 in the remaining 2 modalities. Specifically, it achieves notable AUC improvements in MRI (1.23\% vs. second-best, p < 0.001), fundus imaging (0.24\% vs. second-best, p < 0.001), OCT (0.24\% vs. second-best, p < 0.001), CT (0.90\% vs. second-best, p < 0.001), X-ray (3.25\% vs. second-best, p < 0.001), Endoscopy (0.45\% vs. second-best, p < 0.001), Pathology (2.48\% vs. second-best, p = 0.065). Even in challenging settings like ultrasound and dermatology where other generic models previously held advantages, MMKD-CLIP ranks within the top 3, showcasing its versatility. Across individual datasets, MMKD-CLIP achieves the best or second-best performance in 24 out of 38 tasks, as shown in the rank distribution plots. These results collectively underscore the efficacy of MMKD-CLIP in capturing fine-grained medical semantics through multimodal knowledge distillation, enabling high performance across modality-specific diagnostic tasks.

Notably, MMKD-CLIP exhibits greater stability and generalization across imaging modalities compared to existing biomedical CLIP variants. As shown in Fig. \ref{fig:f1}, MMKD-CLIP consistently ranks higher when evaluated at the modality level (averaged across datasets) than at the individual dataset level. For example, MMKD-CLIP ranks 1st in the MRI, CT, Fundus, OCT, Endoscopy, and X-ray modalities, yet in some of the constituent datasets (e.g., BreastTumorMRI or SIIMACR), it ranks 2nd or 3rd. This demonstrates that its aggregate performance is robust and not overly dependent on any single dataset. In contrast, models such as BiomedCLIP and UniMedCLIP display more variability. For instance, BiomedCLIP achieves 1st place on specific datasets like NCT-CRA (Pathology) or WCE (Endoscopy), but its average rank drops considerably at the modality level, suggesting it may overfit to certain data characteristics. Similarly, UniMedCLIP achieves a top 2 ranking in OCTMNIST but fails to maintain the high average performance across the broader OCT modality. MMKD-CLIP performs well in most datasets and also maintains consistently high performance across diverse imaging modalities. Such behavior is desirable in clinical applications, where the model must generalize across institutions, devices, and patient populations.

\subsubsection{Linear probing image classification}

\begin{figure}[!t] 
	\centering
	\includegraphics[width=\textwidth]{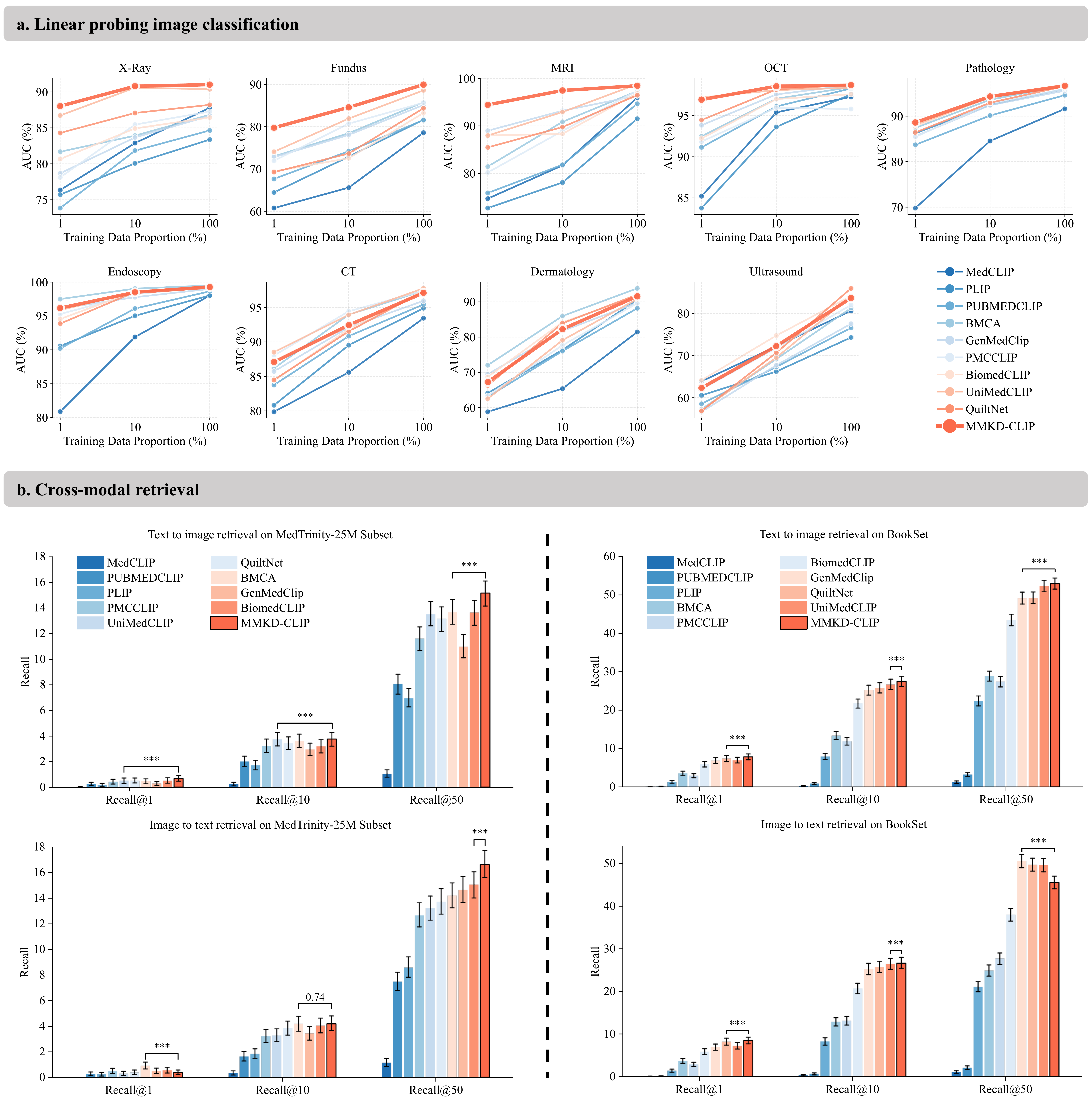}
	\caption{\textbf{Linear probing image classification and cross-modal retrieval performance of MMKD-CLIP.} a. Linear probing image classification. For each of nine imaging modalities (X-ray, fundus, MRI, OCT, pathology, endoscopy, CT, dermatology, ultrasound), we trained a single linear classifier on 1\%, 10\% and 100\% subsets of the respective training data. Lines show mean AUC (\%) across datasets; MMKD-CLIP is highlighted in red, and nine teacher models are shown in other shades. b. Cross-modal retrieval. Left panels report recall@1, @10 and @50 on the MedTrinity-25M subset; right panels show the same on the BookSet benchmark. Bars indicate mean recall (\%) and 95\% CI, which estimated using the bootstrap method (n = 1,000 replicates). Significance of MMKD-CLIP versus the next best model was assessed by a two-sided Mann–Whitney U test: ***: p < 0.001; **: p < 0.01; *: p < 0.05.}
	\label{fig:f3} 
\end{figure}

We benchmark MMKD-CLIP against nine existing biomedical foundation models, including MedCLIP, PLIP, PUBMEDCLIP, BMCA, GenMedClip, PMCCLIP, BiomedCLIP, UniMedCLIP, and QuiltNet, using linear probing classification tasks across 9 medical imaging modalities (Fig. \ref{fig:f3} a). Evaluation was performed under three training data regimes (1\%, 10\%, and 100\%) to assess model generalization under varying supervision. MMKD-CLIP consistently ranks first or ties for the best performance across nearly all modality–proportion combinations. The quantitative results are in Tables \ref{tab:linear_probe_results}--\ref{tab:linear_probe_100_results} in the Supplementary Materials.

In the MRI modality, MMKD-CLIP achieves an AUC of 94.44\% at 1\% training data (95\% CI: 92.55-96.09\%), outperforming the second-best model, GenMedClip, by 5.46\% (p < 0.001). Similar margins are observed in Fundus, where MMKD-CLIP yields an AUC of 79.76\% (95\% CI: 77.69-81.78\%), surpassing UniMedCLIP by 5.69\% (p < 0.001). These statistically significant improvements in high-resolution structural modalities suggest MMKD-CLIP captures fine-grained medical detail with high fidelity. Even in challenging domains like pathology, MMKD-CLIP shows marked robustness, achieving 88.61\% AUC with 1\% data versus 86.25\% from the PLIP specially trained for pathology (p < 0.001), highlighting its adaptability to pathology images.

At higher data availability (10\% and 100\%), all models improve, yet MMKD-CLIP maintains its advantage. For example, in X-ray, AUC improves from 88.03\% (1\%) to 91.02\% (100\%), consistently outpacing other models at each step. In fundus imaging, MMKD-CLIP delivers 89.95\% AUC at 100\% training data, marginally higher than the next-best (88.59\%) but with almost non-overlapping confidence intervals (95\% CI: 88.56-91.16\%, p < 0.001), indicating statistically significant gains even in saturated training scenarios.

On average across all modalities at the 1\% training level, MMKD-CLIP achieves a macro AUC of 84.51\%, representing a 0.37\% improvement over the strongest baseline model (p < 0.001). These advantages persist across both low-data and full-data conditions, pointing to the superior generalization ability and representation quality of MMKD-CLIP. The consistent gap across diverse modalities, from cross-sectional imaging (e.g. MRI and CT) to photographic domains such as dermatology and fundus imaging, demonstrates MMKD-CLIP’s scalability and robustness in real-world diagnostic contexts. Collectively, these results establish MMKD-CLIP as a state-of-the-art vision-language foundation model for medical image understanding.

\subsubsection{Cross-modal retrieval}

To evaluate the alignment capacity between vision and language, we benchmark MMKD-CLIP and nine baseline medical foundation models on large-scale cross-modal retrieval tasks across two diverse datasets: MedTrinity-25M subset~\cite{xie2024medtrinity}, a heterogeneous clinical multimodal corpus, and BookSet~\cite{gamper2021multiple}, a clean academic captioning dataset from ten academic textbooks. The tasks include both text-to-image and image-to-text retrieval, with performance measured at Recall@1, Recall@10, and Recall@50. The quantitative results are in Tables \ref{cross1}--\ref{cross2} in the Supplementary Materials.

\begin{figure}[!t] 
	\centering
	\includegraphics[width=0.9\textwidth]{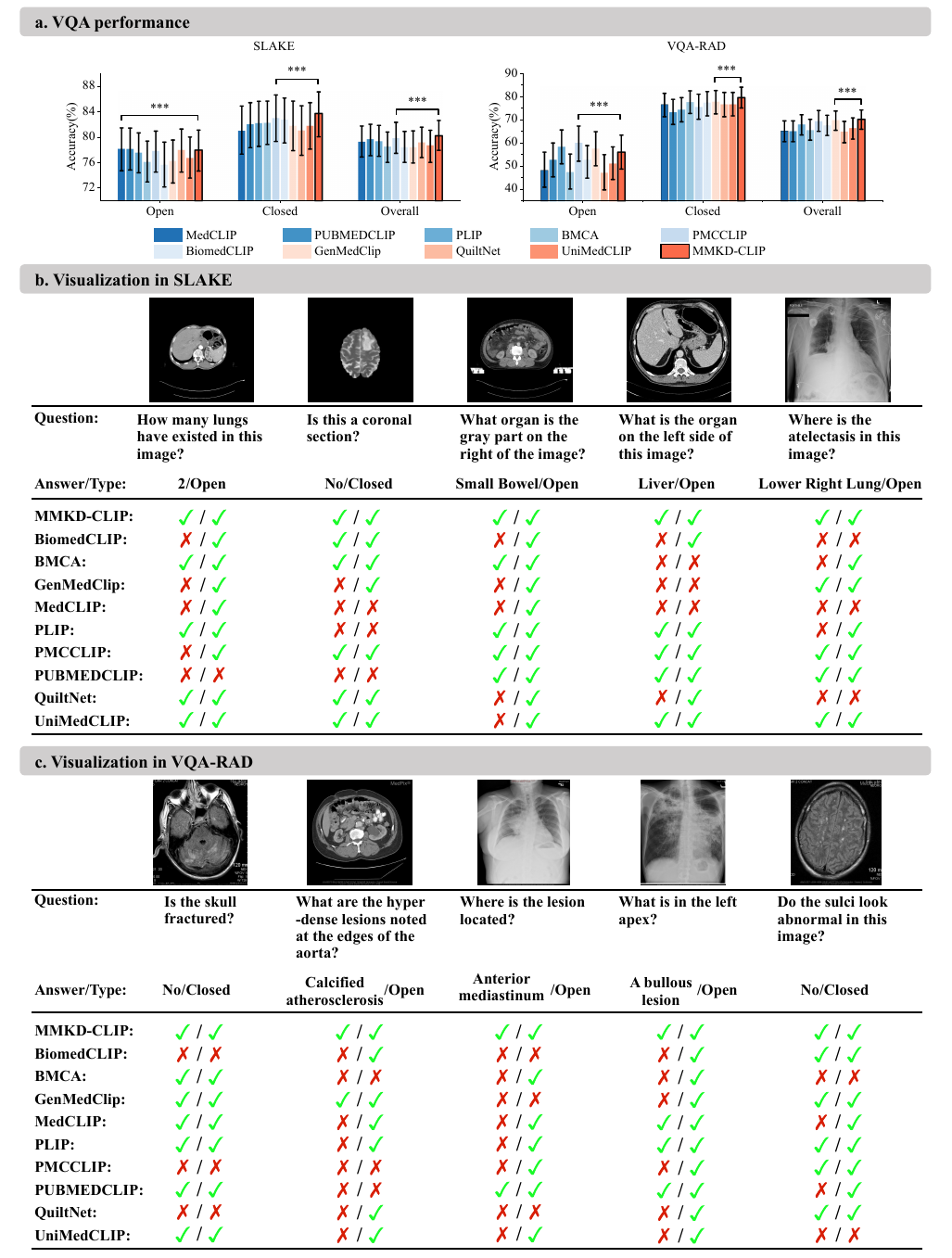}
	\caption{\textbf{Visual question answering performance and qualitative examples.} VQA accuracy (\%) on SLAKE and VQA-RAD benchmarks. Bars show mean accuracy and 95\% CI over all questions in each dataset, separated by ``Open'', ``Closed'' , and ``Overall''. Statistical significance versus the next best model was assessed by a two-sided Mann-Whitney U test: ***: p < 0.001; **: p < 0.01; *: p < 0.05. b and c. Qualitative examples and model responses on SLAKE and VQA-RAD. Five representative prompts (columns) spanning anatomical, pathological and spatial questions are shown with the correct answer and answer type (Open/Closed) beneath each image.  For each model (rows), green checks denote correct classification and content, and red crosses denote errors in either component. MMKD-CLIP achieves the highest consistency across both open- and closed-format queries.}
	\label{fig:f4} 
\end{figure}

\begin{figure}[!t] 
	\centering
	\includegraphics[width=0.82\textwidth]{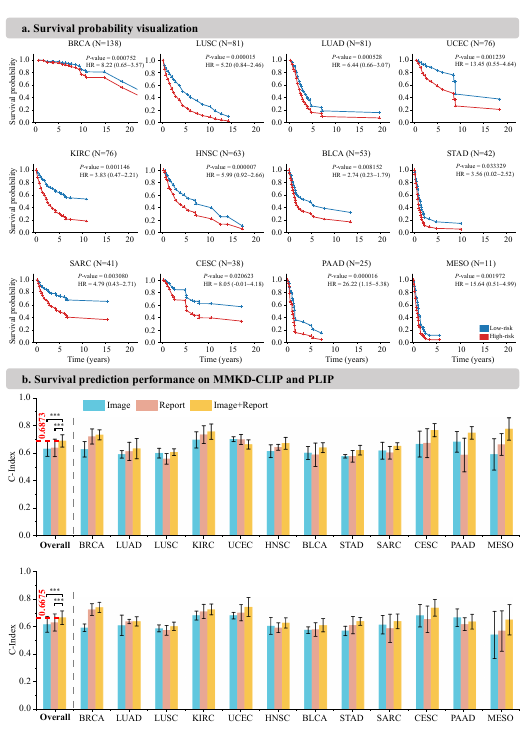}
	\caption{\textbf{Survival prediction with MMKD-CLIP and PLIP.} a. Kaplan-Meier curves for risk stratification across TCGA cohorts. For 12 cancer types (BRCA, LUAD, LUSC, KIRC, HNSC, BLCA, STAD, SARC, CESC, PAAD, UCEC, MESO), patients were split into high-risk (red) and low-risk (blue) groups based on median risk scores derived from whole-slide images and clinical reports. Survival probability is plotted over 20 years; hazard ratios (HR) and two-sided log-rank p-values are indicated on each panel. b. C-index performance for survival prediction. Bars show mean C-index \textpm s.d. for each dataset. Three input modalities were compared: image only (cyan), report only (pink), and combined image+report (apricot). Top plot corresponds to MMKD-CLIP and bottom to the PLIP. In b, data are represented as the mean with standard deviation calculated using five-fold cross-validation experiments. Statistical significance on the C-index was assessed by a two-sided Mann-Whitney U test: ***: p < 0.001; **: p < 0.01;  *: p < 0.05.}
	\label{fig:f5} 
\end{figure}

On the MedTrinity-25M subset, MMKD-CLIP demonstrates substantial improvements in retrieval recall across all metrics. For text-to-image retrieval, MMKD-CLIP achieves a Recall@50 of 15.16\% (95\% CI: 14.15-16.11\%), outperforming the second-best model, BMCA (Recall@50 = 13.67\%, 95\% CI: 12.73-14.66\%), by a margin of 1.49\% (p < 0.001).  Notably, the advantage is also pronounced in low-recall regimes: MMKD-CLIP reaches Recall@10 of 3.76\% vs. UniMedCLIP’s 3.74\% (p < 0.001). MMKD-CLIP's Recall@1 of 0.68\%, whereas most baselines fall below 0.5\% (p < 0.001). This suggests MMKD-CLIP encodes text semantics more precisely for matching diverse medical image types under high-noise supervision. In the image-to-text retrieval task on MedTrinity-25M subset, MMKD-CLIP also ranks first in Recall@50 (16.63\%, 95\% CI: 15.62-17.72\%), outperforming BiomedCLIP by 1.57\% (p < 0.001), and achieves competitive Recall@10 (4.20\%) and Recall@1 (0.38\%, p < 0.001). While some close competitors (e.g., BMCA) show marginal performance on Recall@10 (4.19\%, p < 0.001), MMKD-CLIP exhibits stronger stability across both retrieval directions and across recall thresholds.

BookSet contains better-aligned image–text pairs with fewer domain-specific abbreviations or noise. MMKD-CLIP sustains its lead when tested on this dataset. In the text-to-image setting, Recall@50 rises to 52.95\% (95\% CI: 51.52-54.38\%), compared to UniMedCLIP’s 52.36\% (p < 0.001). For the image-to-text retrieval task, MMKD-CLIP again leads with Recall@10 of 26.61\%, a statistically significant gain of 0.18\% over UniMedCLIP (p < 0.001). The high Recall@1 (8.45\%) further confirms MMKD-CLIP's superior fine-grained semantic grounding.

Collectively, these results highlight MMKD-CLIP’s dual capability in robust retrieval under noisy, large-scale clinical conditions (MedTrinity) and fine-grained matching in curated academic datasets (BookSet). Its performance gains across both retrieval directions, particularly under low-resource conditions (e.g., Recall@1), point to stronger multimodal alignment and superior representation disentanglement compared to existing medical CLIP variants. These findings underscore MMKD-CLIP’s promise as a unified foundation model for downstream multimodal medical tasks.

\subsubsection{Visual question answering}

To evaluate the fine-grained understanding and reasoning capabilities of MMKD-CLIP, we conduct comprehensive evaluations on two established VQA benchmarks. The first is SLAKE~\cite{liu2021slake}, which focuses on multi-organ CT-based anatomical and clinical understanding. The second is VQA-RAD~\cite{lau2018dataset}, a manually curated dataset consisting of 3,515 question–answer pairs, where medical professionals pose natural language inquiries regarding radiological images and provide corresponding reference responses. Tasks are categorized into open-ended, closed-form, and overall questions, with accuracy as the main evaluation metric. The quantitative results are presented in Table \ref{tab:vqa} in the Supplementary Materials.

As shown in Fig. \ref{fig:f4} a, MMKD-CLIP outperforms all baselines across both datasets and question types. On SLAKE, MMKD-CLIP achieves the highest accuracy in overall questions (80.21\%, 95\% CI: 77.95-82.56\%), significantly outperforming the next best model, PMCCLIP (79.83\%, p < 0.001), and far exceeding earlier models like MedCLIP (79.26\%, p < 0.001). The performance gap widens in closed questions, where MMKD-CLIP attains 83.65\% accuracy, a 0.72\% improvement over PMCCLIP (p < 0.001). In VQA-RAD, a dataset known for subtle diagnostic clues and ambiguous visual grounding, MMKD-CLIP continues to lead. On closed questions, MMKD-CLIP achieves 79.34\%, the highest among all models evaluated, resulting in an overall accuracy of 70.07\%, which confirms its robustness across both reasoning types.

To probe the interpretability and alignment of model predictions, we perform qualitative visualizations of representative VQA samples (Fig. \ref{fig:f4} b \& c). In SLAKE (Fig. b), MMKD-CLIP correctly identifies both anatomical structures (e.g., ``liver'' on the left side) and spatial relations (e.g., ``small bowel on the right''), outperforming other models on all samples. In contrast, models like MedCLIP and GenMedClip exhibit inconsistent reasoning, especially under open-ended formulations. In VQA-RAD (Fig. c), MMKD-CLIP demonstrates superior localization (e.g., “anterior mediastinum”) and diagnosis (e.g., “calcified atherosclerosis”), capturing both clinical semantics and visual context. While BMCA and UniMedCLIP succeed on many samples, they occasionally fail under ambiguous syntax or visual occlusion (e.g., sulcal abnormalities). 

Collectively, these results suggest that MMKD-CLIP offers the most balanced and robust VQA capability, excelling in both fine-grained recognition and clinical reasoning, likely attributed to its multi-teacher alignment and extensive modality-spanning pretraining. Its consistent performance across both metrics and examples underscores its potential for real-world decision support in multimodal medical contexts.

\subsubsection{Survival prediction task}

To further assess the clinical utility of MMKD-CLIP, we investigate its capacity for multimodal survival prediction across 12 distinct cancer types from the TCGA dataset~\cite{komura2022universal}, incorporating pathology patch images, reports, and their fusion. Kaplan-Meier (KM) curves and the concordance indices (C-Index) are used to evaluate stratification and prediction quality, respectively. The quantitative results are in Tables \ref{sp1}--\ref{sp2} in the Supplementary Materials.

As visualized in Fig. \ref{fig:f5} a, patients were classified into high-risk and low-risk groups based on model-predicted hazard. Across most cohorts, the stratification by MMKD-CLIP yields statistically significant separation. For instance, in BRCA (N=138), the survival curves diverge significantly (p = 0.000752), with a hazard ratio (HR) of 8.22, indicating MMKD-CLIP's robust ability to identify high-risk individuals. Similar trends are observed in LUAD (HR = 6.44, p = 0.000528), HNSC (HR = 5.99, p = 0.000007), and PAAD (HR = 26.22, p = 0.000016), demonstrating its consistent effectiveness across both large and small cohorts. While small datasets like MESO (N=11) exhibit greater variance, the model still captures meaningful risk trends (HR = 15.64, p = 0.001972), suggesting potential under limited data.

We further benchmark survival prediction performance using the C-Index metric, comparing MMKD-CLIP and PLIP across unimodal and multimodal inputs (Fig. \ref{fig:f5} b). Overall, MMKD-CLIP achieves the highest average C-index of 0.6873 using both image and report inputs, significantly outperforming PLIP (0.6675, p < 0.0001). Notably, the fusion of image and report consistently improves prediction compared to either modality alone across multiple cancers, such as STAD, KIRC, and BLCA, confirming the complementary value of multimodal representations. For example, in CESC, MMKD-CLIP reaches a C-index of 0.7652 with multimodal input, compared to 0.6643 (image-only) and 0.6712 (report-only), indicating improved discrimination of survival outcomes.

Importantly, MMKD-CLIP consistently outperforms PLIP across common cancers (e.g., BRCA, LUAD) and rare, low-sample-size cancers (e.g., SARC, MESO), indicating enhanced generalization capacity. The standard deviation bars also reveal that MMKD-CLIP yields more stable predictions, particularly under multimodal fusion. These results affirm that the joint modeling of visual and textual clinical signals enables MMKD-CLIP to deliver precise and generalizable survival risk prediction, a capability critical for downstream translational applications.

\subsubsection{Supervised cancer diagnosis task}

\begin{figure}[!t] 
	\centering
	\includegraphics[width=\textwidth]{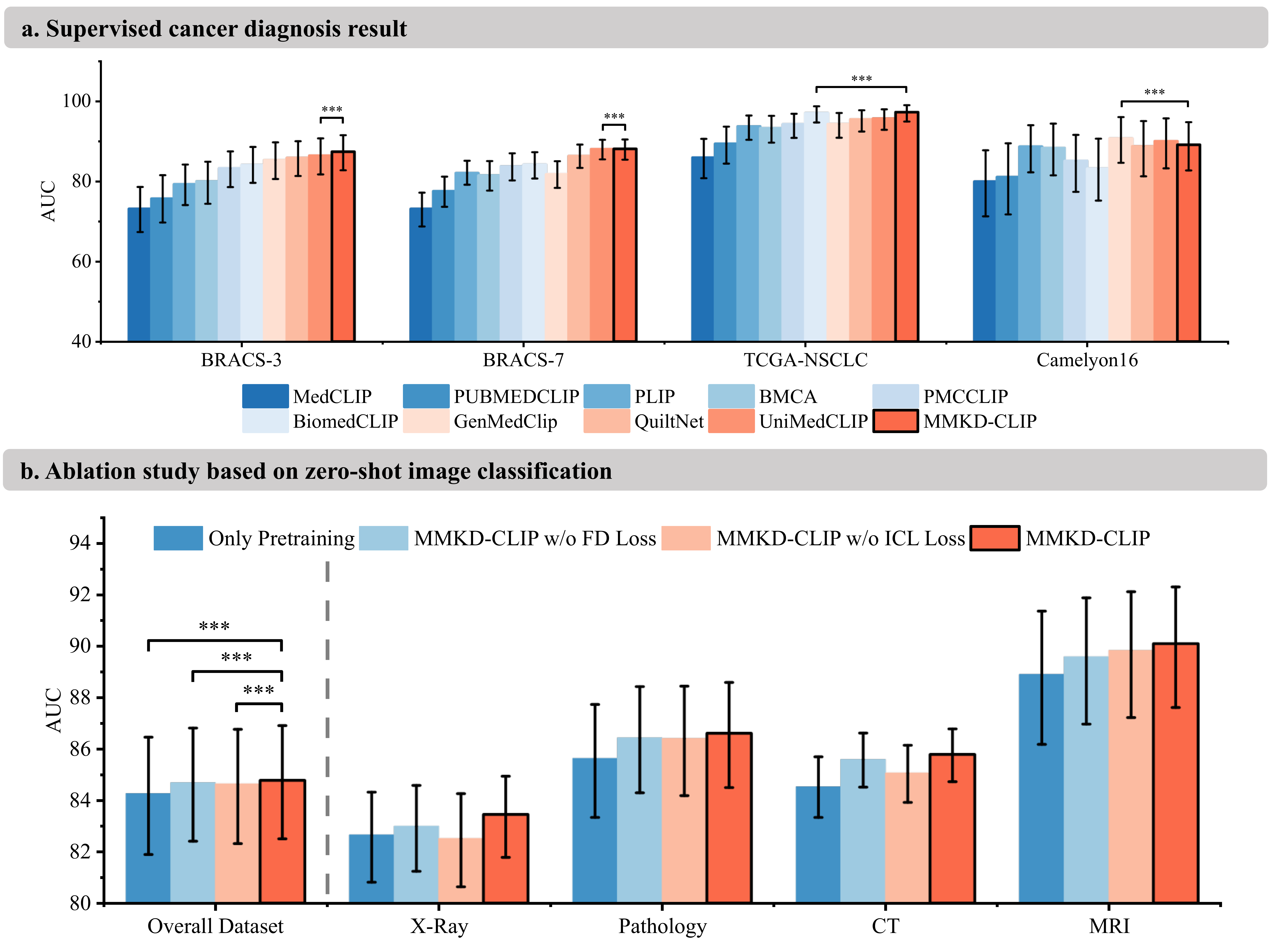}
	\caption{\textbf{Supervised cancer diagnosis performance and ablation study of MMKD-CLIP.} a. Supervised fine-tuning results on cancer diagnosis. Bar plots show mean AUC and 95\%CI on four histopathology cohorts (BRACS-3, BRACS-7, TCGA-NSCLC and Camelyon16) after fine-tuning. Statistical significance versus the next best model was evaluated by a two-sided Mann–Whitney U test: ***: p < 0.001. b. Ablation study on zero-shot image classification. We compare (1) only the contrastive pretraining step, (2) MMKD-CLIP without FD loss, (3) MMKD-CLIP without ICL loss, and (4) full MMKD-CLIP. Bars indicate mean AUC and 95\% CI on the overall test set and on modality-specific subsets (X-ray,pathology, CT, MRI). Significant differences are marked with ***: p < 0.001 by two-sided Mann-Whitney U tests.}
	\label{fig:f6} 
\end{figure}

To evaluate the fine-tuning performance of MMKD-CLIP on supervised cancer classification tasks, we conduct experiments across four benchmark datasets encompassing multiple imaging modalities and anatomical domains: BRACS-3~\cite{brancati2022bracs}, BRACS-7~\cite{brancati2022bracs}, TCGA-NSCLC~\cite{weinstein2013cancer}, and Camelyon16~\cite{bejnordi2017diagnostic}. As shown in Fig. \ref{fig:f6} a, the AUC is used to assess diagnostic accuracy. The quantitative results are presented in Table \ref{cd} in the Supplementary Materials.

On BRACS and TCGA-NSCLC, MMKD-CLIP achieves consistently superior AUC values, outperforming both generalist (e.g., BMCA, GenMedClip) and specialist (e.g., QuiltNet, PLIP) baselines. On BRACS-3, MMKD-CLIP yields an AUC of 87.42\%, significantly higher than the best baseline (PLIP, AUC = 86.44\%, p < 0.001). This margin is further widened in BRACS-7, a more granular 7-class variant, where MMKD-CLIP reaches an AUC of 88.13\%, demonstrating robust class separability in multi-class settings (p < 0.001). For TCGA-NSCLC, MMKD-CLIP maintains top-tier performance with an AUC of 97.28\%, exceeding all multimodal or unimodal-pretrained counterparts (p < 0.001). Meanwhile, on Camelyon16, a challenging histopathology dataset, MMKD-CLIP achieves 89.17\% AUC, surpassing most competitors including expert models like QuiltNet and PLIP (p < 0.001), highlighting its capacity to adapt across modality shifts.

These results underscore MMKD-CLIP’s ability to generalize effectively across distinct cancer diagnosis tasks, ranging from radiology to pathology, binary to multi-class classification, and coarse to fine-grained labels. The consistent outperformance across datasets reaffirms the benefits of multi-teacher distillation and unified vision-language pretraining in improving downstream diagnostic accuracy.

\subsection{Ablation study}

we perform a comprehensive ablation study on the zero-shot image classification task across four major imaging modalities: X-ray, pathology, CT, and MRI. Furthermore, we compare the overall performance of each component on all 38 datasets (``Overall Dataset''). As shown in Fig. \ref{fig:f6} b, we compare four settings: (i) pretraining only without distillation, (ii) full model without the FD loss, (iii) full model without the ICL loss, and (iv) the full MMKD-CLIP with all components. The quantitative results are presented in Table \ref{ab} in the Supplementary Materials.

Across all domains, removing either FD or ICL leads to consistent performance drops, underscoring their complementary effects. Specifically, removing FD loss results in a 0.09-1.26\% decrease across modalities, while removing ICL results in a 0.14-0.72\% drop. The full MMKD-CLIP consistently achieves the best performance in all settings, with an overall accuracy of 84.78\% (95\% CI: 82.51-86.91\%), outperforming the second-best configuration (w/o FD loss, 84.69\%) and the pretraining-only baseline (84.26\%). Notably, the X-ray domain benefits most from ICL, where the model without ICL underperforms compared to the version without FD loss (82.52\% vs. 82.99\%), suggesting strong inter-modal teacher consistency helps address modality sparsity in radiography. In contrast, the pathology and CT domains exhibit larger gains from the FD loss, possibly due to the dense, high-resolution visual information that benefits more from precise feature alignment. The MRI domain demonstrates the highest classification scores overall, with MMKD-CLIP reaching 90.09\% (CI: 87.61-92.30\%), confirming the model's strong generalization to complex anatomical patterns. The gap between pretraining-only (88.90\%) and full MMKD-CLIP (90.09\%) further emphasizes the added value of multi-teacher supervision in high-dimensional medical modalities. 

These results validate that each loss component contributes uniquely to model performance, and the synergy of ICL loss and FD loss is essential for achieving state-of-the-art generalization in zero-shot biomedical classification across diverse modalities.

\section{Discussion}
\label{sec:discussion}

Foundation models in the general vision-language domain have profoundly reshaped how we tackle open-ended reasoning, zero-shot classification, and cross-modal retrieval. However, their direct application in medicine is hampered by the heterogeneous nature of biomedical imaging and the fragmented availability of large-scale image-text corpora. In this study, we introduced MMKD-CLIP, a generalist biomedical foundation model trained via multi-teacher medical CLIP knowledge distillation, and demonstrated its ability to bridge the gap between narrow, domain-specific CLIP variants and the broad interdisciplinary demands of real-world clinical practice. Our extensive evaluation on 58 datasets spanning zero-shot classification, linear probing, cross-modal retrieval, VQA, survival prediction, and supervised cancer diagnosis validates that MMKD-CLIP not only achieves state-of-the-art performance but also exhibits remarkable stability and generalizability across heterogeneous imaging domains.

By leveraging a two-stage pipeline (pretraining on 2.9 million image-text pairs followed by offline distillation from 9 specialist and generalist biomedical CLIP teachers), MMKD-CLIP acquires a unified representation space. The zero-shot classification results (Fig. \ref{fig:f1}) confirm that MMKD-CLIP not only outperforms each individual teacher across seven of nine modalities but also maintains stability when aggregated at the modality level. This finding directly addresses the shortcoming that existing biomedical CLIP models often overfit to their narrowly curated domains and fail to generalize when confronted with unfamiliar modalities or mixed clinical settings. MMKD-CLIP’s strong performance across MRI, CT, X-Ray, Endoscopy, Pathology, OCT, and Fundus demonstrates its ability to internalize diverse, high-fidelity visual semantics without additional fine-tuning.

In the linear probing experiments (Fig. \ref{fig:f3} a), particularly under low-data regimes (1\% and 10\% of training samples), MMKD-CLIP consistently maintains significant AUC advantages over all baselines (e.g., 5.46\% in MRI at 1\% data, p < 0.001). These results underscore how the multi-teacher distillation strategy allows MMKD-CLIP to ``borrow'' complementary strengths from each teacher, such as PubMedCLIP's fine-grained subfigure-subcaption alignments, and QuiltNet and PLIP’s histopathology expertise, which ultimately produces a student that requires far less task-specific supervision. In practical terms, this means that clinics or resource-constrained settings could deploy MMKD-CLIP for novel classification tasks with very limited labeling effort, a critical advantage in low- and middle-income regions where medical annotations are costly and scarce.

Cross-modal retrieval (Fig. \ref{fig:f3} b) further illustrates MMKD-CLIP’s robust alignment of vision and language representations. On the MedTrinity-25M subset, MMKD-CLIP achieves a text-to-image Recall@50 of 15.16\%, representing a 1.49\% improvement over the strongest baseline (BMCA; p < 0.001), and an image-to-text Recall@50 of 16.63\%, a 1.57\% gain (BiomedCLIP; p < 0.001). Meanwhile, on the cleaner BookSet, MMKD-CLIP reaches a Recall@50 of 52.95\%, outperforming UniMedCLIP by 0.59\% (p < 0.001). These dual gains demonstrate the utility of MMKD-CLIP as a flexible backbone for both chart searches and real-time case‐query systems, which is an essential capability for a generalist biomedical model operating across varied clinical repositories and textual conventions.

In VQA (Fig. \ref{fig:f4}) task, MMKD-CLIP achieves the highest accuracy across SLAKE and VQA-RAD benchmarks. Its ability to correctly label both anatomic structures and pathologic features, such as ``small bowel on the right'' or ``calcified atherosclerosis'', demonstrates that the joint embedding space learned during distillation effectively captures both local visual details and contextual clinical semantics. This balanced reasoning ability integrates multiple disciplines, simulating the physician's need to combine radiology, pathology, and narrative reports when forming a diagnosis. Importantly, MMKD-CLIP’s qualitative visualizations reveal consistently correct reasoning even under ambiguous phrasing or occluded imagery, which is an essential trait for safe deployment in diagnostic support tools.

Survival prediction on TCGA (Fig. \ref{fig:f5}) further extends MMKD-CLIP's utility beyond classification and retrieval tasks into prognostic modeling. By fusing pathology patches with clinical reports, MMKD-CLIP achieves significantly higher C-Index values (e.g., 0.6873 versus PLIP’s 0.6675, p < 0.0001). This improvement underscores that a unified vision-language representation can capture complementary prognostic signals (morphologic patterns in histology and nuanced clinical notes) that neither modality can fully convey alone. In effect, MMKD-CLIP operationalizes this ideal by learning a single model capable of mapping all relevant data sources into a risk-score prediction, thereby advancing the goal of comprehensive, data-driven clinical decision support.

In supervised cancer diagnosis tasks (Fig. \ref{fig:f6} a), MMKD-CLIP’s fine-tuned AUCs, ranging from 87.42\% on BRACS-3 to 97.28\% on TCGA-NSCLC, affirm that a distilled model can match or exceed specialist teacher performance when provided with a modest amount of labeled data. Notably, MMKD-CLIP’s low variance across folds suggests that its pretrained weights are less sensitive to sampling noise, which is a recurring challenge in medical imaging where dataset sizes can vary widely. By outperforming domain-specific models (e.g., QuiltNet for histopathology, PLIP for pathology) and generalist variants (e.g., GenMedCLIP, BMCA), MMKD-CLIP fulfills the central objective of this study: providing a unified foundation that can be reliably fine-tuned for diverse downstream tasks without requiring multiple separate models.

Ablation studies (Fig. \ref{fig:f6} b) confirm that both the FD loss and the ICL loss are essential for MMKD-CLIP’s performance. Removing FD leads to a 0.09-1.26\% drop across modalities, and removing ICL causes a 0.14-0.72\% decrease. These complementary effects suggest that FD fosters precise alignment with each teacher’s modality-specific expertise, while ICL facilitates cross-teacher semantic harmonization.

In summary, MMKD-CLIP is a generalist biomedical foundation model that unifies diverse imaging modalities and narrative contexts. Through multi-teacher distillation, we demonstrate that integrating expert knowledge from multiple domain-specific CLIP models yields a student that is both more accurate and more robust than any individual teacher. Our extensive evaluations across 58 datasets and six task categories confirm that MMKD-CLIP sets a new benchmark for biomedical vision-language AI, offering a scalable, extensible platform for future clinical and research applications.

\section{Methods}

\subsection{Pretraining data curation}
For pretraining, we directly utilize the PMC-OA dataset curated by the BMCA~\cite{lozano2025biomedica} project. Specifically, we adopt the ``Concept-Filtering'' subset, which comprises approximately 6 million image–text pairs. To further refine the dataset, we apply an additional filtering step that retains only those pairs annotated with an ``image\_primary\_label'' of ``Microscopy'' or ``Clinical Imaging.'' This results in a total of 2,911,190 pairs, including 1,416,257 single-panel images and 1,494,933 multi-panel images. In Table \ref{tab:modality_distribution}, we summarize the distribution of image-text pairs across 26 imaging modalities represented in the filtered subset.

\subsection{Model design and pretraining}
\label{mdp}
The pre-training and distillation of MMKD-CLIP is inspired by CLIP-KD~\cite{yang2024clip}, consists of two main steps. The first step utilizes approximately 2.9 million image-text pairs collected from the PMC-OA database and contrastive learning to align the semantics of the two modalities. The second step, as shown in Fig. \ref{fig:choosteacher}, for each image–text pair, we introduce four additional unpaired text samples and perform zero-shot classification using nine biomedical CLIP models. A CLIP model is deemed a trustworthy teacher if it achieves over 90\% accuracy on the correct class. This pipeline yields a distillation-ready quadruplet comprising the original image, original text, and the corresponding visual and textual features extracted by the teacher model. Ultimately, this pipeline yielded a total of 19,229,852 quadruplet samples, with the distribution of samples extracted by each CLIP model illustrated in Fig. \ref{fig:choosteacher} c. As illustrated in Fig. \ref{fig:overview} c, knowledge distillation is then performed using both FD loss and ICL loss. The network backbone includes a visual encoder, using the MetaCLIP ViT-B/16 model~\cite{xu2023demystifying}, and a text encoder, using the BioMed-BERT text encoder~\cite{chakraborty2020biomedbert}. Model configurations are specified in Supplementary Table \ref{tab:multimodal_config}.

\subsection{Multi-teacher dimension alignment}
\label{mtda}
BMCA~\cite{lozano2025biomedica} and PMCCLIP~\cite{lin2023pmc} produce outputs of 768 dimensions, whereas other CLIP models output 512 dimensions, which hinders the implementation of distillation. Therefore, as illustrated in Fig. \ref{tab:multimodal_config} b, we constructed a dual-stream autoencoder for dimensional alignment. Specifically, we prepared a corresponding projection encoder for each CLIP to map them to the same dimension, where we chose 512. Then, the projection vectors of all CLIPs are fed into a set of shared dual-stream autoencoders to learn a joint feature space for all CLIPs. Finally, we designed a projection decoder for each CLIP to reconstruct the features. It is noteworthy that we do not seek to implement contrastive learning loss in the latent space, as this would disrupt the distribution of the original feature space of each CLIP.

\subsection{The formalized description of the MMKD-CLIP pipeline}
\label{44}
\textbf{Pretraining phase}, given a multi-modal dataset containing $N$ image-text pairs, denoted as $\mathcal{D}=\{(I_1,T_1),...,(I_i,T_i),...,(I_N,T_N)\}$, where $I_i$ represents the $i$-th image, $T_i$ represents the $i$-th caption. Our goal is to develop a biomedical VLM utilizing an visual encoder $f(\cdot)$ and a text encoder $g(\cdot)$ that performs an image-text alignment task to push the paired image-text close and unpaired ones apart in the feature embedding space. Specifically, given an image-text pair $(I_i,T_i)$, the visual encoder $f(\cdot)$ and text encoder $g(\cdot)$ are employed to encode the pair as follows:

\begin{equation}
	v_i = f(I_i), v_i \in \mathcal{R}^d
	\label{e1}
\end{equation}

\begin{equation}
	t_i = g(T_i), t_i \in \mathcal{R}^d
	\label{e2}
\end{equation}

Where $v_i$ represents the global features of the image, $t_i$ represents the global features of the caption. $d$ represents the dimension of feature embedding space. Here, all features are post-processed by L2 normalization.

CLIP-style pretraining employs a contrastive learning framework inspired by InfoNCE~\cite{oord2018representation} to align visual and textual representations. Specifically, given an image embedding vector \( v_i \) and its corresponding text embedding \( t_i \), the objective is to maximize their similarity while minimizing similarity to mismatched pairs within a mini-batch. The image-to-text contrastive loss, where the image serves as the anchor, is defined as:

\begin{equation}
	\mathcal{L}_{I \rightarrow T} = - \sum_{i=1}^{|\mathcal{B}|} \log \frac{\exp(v_i \cdot t_i / \tau)}{\sum_{b=1}^{|\mathcal{B}|} \exp(v_i \cdot t_b / \tau)}
	\label{e3}
\end{equation}

Conversely, the model also optimizes a symmetric contrastive loss in the reverse direction, with the text embedding \( t_i \) as the anchor. The corresponding text-to-image loss is formulated as:

\begin{equation}
	\mathcal{L}_{T \rightarrow I} = - \sum_{i=1}^{|\mathcal{B}|} \log \frac{\exp(t_i \cdot v_i / \tau)}{\sum_{b=1}^{|\mathcal{B}|} \exp(t_i \cdot v_b / \tau)}
	\label{e4}
\end{equation}

Where the operator ``\(\cdot\)'' denotes the dot product measuring the cosine similarity between embedding vectors, and \(\tau\) is a learnable temperature parameter that scales the logits. The summations are computed over all samples within a mini-batch \(\mathcal{B}\). The final training objective of CLIP integrates both directions by averaging the image-to-text and text-to-image losses:

\begin{equation}
	\mathcal{L}_{\mathrm{CLIP}} = \frac{1}{2} \left( \mathcal{L}_{I \rightarrow T} + \mathcal{L}_{T \rightarrow I} \right)
	\label{e5}
\end{equation}

\textbf{Knowledge distillation phase}, to narrow the performance discrepancy between teacher and student models, a straightforward yet effective method involves aligning their intermediate feature representations, which is called feature distillation (FD). The core idea is that if the student's features can approximate those of the teacher, the difference in performance may be minimized. Therefore, we encourage the student to replicate both the image and text embeddings of the teacher by minimizing the Mean Squared Error (MSE) loss, defined as:

\begin{equation}
	\mathcal{L}_{\mathrm{FD}} = \frac{1}{|\mathcal{B}|} \sum_{i=1}^{|\mathcal{B}|} \left( \left\| v_i^{\mathrm{T}} - v_i^{\mathrm{S}} \right\|_2^2 + \left\| t_i^{\mathrm{T}} - t_i^{\mathrm{S}} \right\|_2^2 \right)
	\label{e6}
\end{equation}

Where \(v_i^{\mathrm{T}}\) and \(t_i^{\mathrm{T}}\) represent the teacher's visual and textual feature embeddings for the \(i\)-th instance, while \(v_i^{\mathrm{S}}\) and \(t_i^{\mathrm{S}}\) denote those from the student. The loss aggregates the squared \(\ell_2\) distance across all samples in the mini-batch \(\mathcal{B}\).

To enhance knowledge transfer between the teacher and student models, we introduce an approach termed \textit{Interactive Contrastive Learning} (ICL), which builds contrastive relationships between cross-model embeddings. Specifically, the student’s encoder outputs are contrasted against teacher embeddings rather than within the same network. When using the student's visual representation \(v_i^{\mathrm{S}}\) as the query, and teacher text embeddings \(\{t_b^{\mathrm{T}}\}_{b=1}^{|\mathcal{B}|}\) as targets, the image-to-text contrastive objective under the ICL framework is defined as:

\begin{equation}
	\mathcal{L}_{\mathrm{ICL}, I \rightarrow T} = - \sum_{i=1}^{|\mathcal{B}|}\log \frac{\exp(v_i^{\mathrm{S}} \cdot t_i^{\mathrm{T}} / \tau)}{\sum_{b=1}^{|\mathcal{B}|} \exp(v_i^{\mathrm{S}} \cdot t_b^{\mathrm{T}} / \tau)}
	\label{e7}
\end{equation}

In a reciprocal manner, for student-generated text features \(t_i^{\mathrm{S}}\), the corresponding teacher visual embeddings \(\{v_b^{\mathrm{T}}\}_{b=1}^{|\mathcal{B}|}\) are used to compute the text-to-image contrastive loss:

\begin{equation}
	\mathcal{L}_{\mathrm{ICL}, T \rightarrow I} = - \sum_{i=1}^{|\mathcal{B}|}\log \frac{\exp(t_i^{\mathrm{S}} \cdot v_i^{\mathrm{T}} / \tau)}{\sum_{b=1}^{|\mathcal{B}|} \exp(t_i^{\mathrm{S}} \cdot v_b^{\mathrm{T}} / \tau)}
	\label{e8}
\end{equation}

The overall ICL objective integrates both directional losses by averaging:

\begin{equation}
	\mathcal{L}_{\mathrm{ICL}} = \frac{1}{2} \left( \mathcal{L}_{\mathrm{ICL}, I \rightarrow T} + \mathcal{L}_{\mathrm{ICL}, T \rightarrow I} \right)
	\label{e9}
\end{equation}

Optimizing the ICL loss can be interpreted as increasing the lower bound on mutual information between student and teacher representations. CLIP-KD~\cite{yang2024clip} has proved this theory. This mutual information perspective suggests that as the student learns to align with the teacher’s embeddings, particularly when contrasting against teacher outputs, it gains reduced uncertainty and better representation quality. This dynamic encourages deeper knowledge sharing from teacher to student networks. 

To unify the objectives in the distillation phase, we adopt a weighted combination of the aforementioned losses. Specifically, we retain the original CLIP loss with a smaller weight to maintain modality alignment, and integrate both the feature distillation loss and the interactive contrastive loss to fully facilitate knowledge transfer. The overall distillation objective is expressed as:

\begin{equation}
	\mathcal{L}_{\mathrm{KD}} = \alpha_1 \cdot \mathcal{L}_{\mathrm{CLIP}} + \alpha_2 \cdot \mathcal{L}_{\mathrm{FD}} + \alpha_3 \cdot \mathcal{L}_{\mathrm{ICL}}
	\label{e10}
\end{equation}
Where $\alpha_1$, $\alpha_2$, $\alpha_3$ are weight coefficients, which are set to 0.1, 50, and 1 respectively.

\subsection{Zero-shot and linear probing image classification}

We evaluate the zero-shot and linear probing image classification abilities of MMKD-CLIP alongside previous leading medical CLIP models across 38 datasets (Tables \ref{zerobeggin} to \ref{zeroend}) covering 9 imaging modalities. In the zero-shot scenario, no ground truth labels are available. We adhere to the zero-shot protocol of the CLIP model~\cite{radford2021learning}, that is, to combine the label and the prompt word into a sentence, and calculate the similarity between the image, and the positive and negative sentence samples. Notably, for both positive and negative samples, we generate multiple sentences for each label using various prompts and aggregate these by averaging their embeddings to accommodate descriptive variations. Ultimately, the class corresponding to the highest image-text similarity is designated as the final classification result.
For linear probing, we utilize the frozen MMKD-CLIP to extract embedded features for each image and employ a lightweight linear classifier for the classification process. The classifier is optimized using the training set of downstream datasets and then evaluated on their test sets. Notably, we keep the test set unchanged while varying the proportion of the training set used during optimization to quantify the performance of low-cost generalization.
We use the AUC score for evaluation to ensure that the performance is not influenced by thresholds.

\subsection{Cross-modal retrieval}

To evaluate whether MMKD-CLIP aligns image concepts and text concepts in the latent space, we perform image-to-text and text-to-image retrieval. In this task, we freeze MMKD-CLIP to compute embedding features for all images and texts, and then use cosine similarity to determine the similarity between each possible pair. For text-to-image retrieval, we retrieve the top-K images closest to a given text query in the aligned  latent space. Similarly, image-to-text retrieval follows an analogous procedure. We utilize the Recall@K metric to evaluate the retrieval performance of MMKD-CLIP. This metric calculates the proportion of test set candidates where the true result is found within the top-K retrieved samples. We chose K values from \{1, 10, 50\}.

\subsection{Visual question answering}

The VQA task is the simplest way to achieve human-computer interaction. Given an image and a natural language question, the model generates a reasonable answer to the question. We use the framework provided by PubMedCLIP~\cite{eslami2023pubmedclip} to facilitate our experiments on the VQA task, which treats VQA as a classification task. We replace the image and text encoders in this framework with frozen MMKD-CLIP or various other state-of-the-art biomedical CLIP models, while keeping the rest unchanged. All models were fine-tuned using the QCR~\cite{eslami2023pubmedclip} (question answering via conditional reasoning) framework. Finally, we use closed accuracy, open accuracy, and overall accuracy to evaluate the performance of all models.

\subsection{Supervised cancer diagnosis}

We used frozen MMKD-CLIP to extract visual features of pathology slides offline, which are input into the ABMIL model~\cite{ilse2018attention} for multiple instance learning (MIL). As shown in Fig. \ref{fig:overview} d, the output of the model is the probability distribution of cancer categories. We use AUC as the evaluation metric to eliminate the impact of the threshold.

\subsection{Survival prediction}

We evaluated MMKD-CLIP on a survival prediction task using a combination of pathology images and clinical reports. To this end, we used the slice dataset (0.5 $\mu$m/pixel) provided by the TCGA-UT~\cite{komura2022universal}, which involves 32 types of cancer. In addition, according to MUSK's practice~\cite{xiang2025vision}, we matched the TCGA-UT dataset with TCGA-reports~\cite{kefeli2024tcga} and eliminated cancer types with a small number of patients, and finally selected 12 cancer types: BRCA, LUAD, LUSC, KIRC, UCEC, HNSC, BLCA, STAD, SARC, CESE, PAAD, MESO. We used ABMIL~\cite{ilse2018attention} to aggregate image features and trained corresponding prognostic models for each cancer type based on multiple instance learning (MIL). Since this task predicts the probability of survival time falling into each interval, it can be regarded as a classification task. We evaluated the model through five-fold cross validation and chose concordance index (C-Index) as the evaluation metric.

\subsection{State-of-the-art models for comparison}
\label{sota}
MMKD-CLIP is distilled from 9 renowned and state-of-the-art biomedical CLIP models. Therefore, we selected these 9 models for comparison, which include BMCA~\cite{lozano2025biomedica}, GenMedClip~\cite{ikezogwo2025medicalnarratives}, MedCLIP~\cite{wang2022medclip}, BiomedCLIP~\cite{zhang2025multimodal}, UniMedCLIP~\cite{khattak2024unimed}, PLIP~\cite{huang2023visual}, PubMedCLIP~\cite{eslami2023pubmedclip}, QuiltNet~\cite{ikezogwo2023quilt}, and PMC-CLIP~\cite{lin2023pmc}. 
\begin{itemize}
	\item BMCA is trained on the large-scale BIOMEDICA dataset, which includes over 24M image-text pairs from over 6M open-access scientific articles. 
	\item GenMedClip is a vision-language model built on the CLIP architecture, pretrained on 4.7M medical image-text pairs spanning 12 imaging modalities. It is trained using the MedicalNarratives dataset, which captures synchronized instructor speech and mouse-tracking data from pedagogical videos, providing both semantic and dense annotations (e.g., traces and bounding boxes) to support unified learning across diverse medical imaging tasks. 
	\item MedCLIP addresses the scarcity of large-scale paired datasets and the prevalent issue of false negatives in traditional contrastive learning. By decoupling images and text, MedCLIP can combine and expand training pairs, leveraging a semantic matching loss function based on medical knowledge. Remarkably, even with only 10\% of the training data, MedCLIP outperforms state-of-the-art models.
	\item BiomedCLIP is a multimodal foundation model pretrained on PMC-15M, a large-scale dataset of 15 million image-text pairs sourced from 4.4 million scientific articles. 
	\item UniMedCLIP is a unified vision-language model trained on UniMed, a large-scale open-source dataset of 5.3 million image-text pairs spanning six imaging modalities (X-ray, CT, MRI, Ultrasound, Pathology, Fundus).
	\item PLIP is a multimodal vision-language model trained on OpenPath, a large-scale dataset of 208,414 de-identified pathology images with natural language descriptions curated from public platforms like medical Twitter.
	\item PubMedCLIP is a domain-adapted version of CLIP fine-tuned on PubMed articles to tackle Medical Visual Question Answering (MedVQA). By leveraging medical image-text pairs, PubMedCLIP outperforms prior methods such as MAML, achieving up to 3\% higher accuracy on MedVQA benchmarks and demonstrating the promise of contrastive pretraining in specialized biomedical tasks.
	\item QuiltNet is trained on QUILT-1M, which is a vision-language dataset in histopathology, built from 1,087 hours of expert YouTube videos and diverse web sources. By fine-tuning a pre-trained CLIP model on this 1M image-text pair dataset, QuiltNet enables superior zero-shot classification and cross-modal retrieval across 13 datasets spanning 8 histopathological subtypes.
	\item PMCCLIP is trained on PMC-OA, a 1.6M biomedical image-caption dataset curated from PubMed OA database. Leveraging fine-grained subfigure-subcaption alignments, PMC-CLIP achieves state-of-the-art performance across retrieval, classification, and VQA tasks in the biomedical domain.
\end{itemize}

\subsection{Benchmark datasets}

We evaluated the MMKD-CLIP model across multiple medical AI tasks using 58 publicly available benchmark datasets spanning zero-shot classification, linear probing image classification, cross-modal retrieval, visual question answering, and outcome prediction. For zero-shot and linear probing image classification, we used 38 different medical imaging datasets across 9 modalities:
\begin{itemize}
	\item \textbf{X-Ray}: X-Ray datasets include CovidCXR4~\cite{wang2020covid}, which comprises 84,818 chest X-ray images from 45,342 patients, labeled for binary classification to detect COVID-19 infection versus normal findings. DDSM~\cite{heath2001ddsm} contains 71,249 pre-processed breast X-ray images, combining data from the DDSM and CBIS-DDSM collections. These images are labeled for binary classification to detect breast cancer, distinguishing between positive findings (malignant/suspicious lesions) and negative findings (normal breast tissue). NLMTB~\cite{jaeger2014two} includes 4,200 chest X-ray radiographs from the U.S. National Library of Medicine, labeled for binary classification to detect tuberculosis versus normal chest findings. SIIMACR~\cite{zawacki2019siim} is a chest X-ray dataset with 12,047 DICOM images labeled for binary classification and segmentation to detect pneumothorax, with pixel-level annotations for positive cases. Rsna\_pneumonia~\cite{shih2019augmenting} comprises 30,227 frontal-view chest radiographs labeled for binary classification to detect pneumonia, derived from the RSNA Pneumonia Detection Challenge 2018.
	
	\item \textbf{CT}: CT datasets include BrainTumorCT~\cite{likhon2023brain}, which comprises 4,618 high-resolution brain CT scans labeled for binary classification to distinguish between healthy brain tissue and brain tumors.  CT\_axial, CT\_coronal, CT\_sagittal~\cite{woerner2024comprehensive} are three subsets from the MedIMeta collection, comprising 1,645 standardized abdominal CT image slices in axial, coronal, and sagittal planes. These images are labeled for 11-class organ classification, identifying anatomical structures such as the heart, lungs, liver, kidneys, spleen, pancreas, bladder, and femoral heads. CovidCT3A~\cite{gunraj2022covid} is a large-scale dataset with 425,024 chest CT slices from 5,312 patients, labeled for 3-class classification to distinguish between normal lungs, pneumonia, and COVID-19 infection. Organamnist~\cite{yang2023medmnist} includes 58,830 abdominal CT images pre-processed to 28×28 pixels from the MedMNIST collection, labeled for 11-class classification to identify different organs, including the bladder, femurs, heart, kidneys, liver, lungs, pancreas, and spleen.

	\item \textbf{MRI}: MRI datasets include BrainTumorMRI~\cite{likhon2023brain}, which comprises 5,000 high-resolution brain MRI scans from multiple patients. This dataset is labeled for binary classification to distinguish between healthy brain tissue and the presence of brain tumors, leveraging MRI's superior soft tissue contrast for detailed analysis. BrainTumorMRI2~\cite{msoud_nickparvar_2021} contains 7,023 brain MRI images labeled for 4-class classification, distinguishing between glioma, meningioma, pituitary tumors, and healthy brain tissue. Acl\_mri~\cite{bien2018deep} includes 1,021 knee MRI examinations for binary classification of ACL (anterior cruciate ligament) tears.

	\item \textbf{Ultrasound}: Ultrasound datasets include BUSBRA~\cite{nastase2024role}, which comprises 1,875 breast ultrasound images from 1,064 patients with biopsy-proven cases. The dataset is labeled for binary classification to distinguish between malignant and benign breast lesions. Breast\_us~\cite{al2020dataset} contains 780 images from 600 female patients, also labeled for binary classification between benign and malignant breast lesions.  Breastmnist~\cite{yang2023medmnist} includes 780 breast ultrasound images pre-processed to 28×28 pixels from the MedMNIST collection. It is labeled for binary classification to differentiate between malignant and normal/benign breast conditions.

	\item \textbf{Fundus}: Fundus datasets include DRD~\cite{dugas2015drd}, which comprises 35,125 retinal photographs labeled for 5-class diabetic retinopathy severity grading, ranging from no diabetic retinopathy to proliferative diabetic retinopathy. FundusJSIEC~\cite{cen2021automatic} contains 1,000 retinal images from the Joint Shantou International Eye Centre, labeled for 39-class classification. This dataset covers a wide range of retinal diseases and conditions, including diabetic retinopathy, glaucoma, macular disorders, and vascular occlusions. Five\_retina~\cite{jin2022fives} includes 800 retinal photographs labeled for 4-class classification to detect age-related macular degeneration, diabetic retinopathy, glaucoma, and normal conditions, derived from the FIVES dataset.
	
	\item \textbf{OCT}: OCT datasets include OCTMNIST~\cite{yang2023medmnist}, which comprises 109,309 optical coherence tomography images from the MedMNIST v2 collection. These images are pre-processed to a standardized 28×28 pixel grayscale format, facilitating lightweight classification tasks. The dataset is labeled for 4-class classification to identify choroidal neovascularization (CNV), diabetic macular edema (DME), drusen, and normal retinal conditions. RetinalOCT~\cite{naren2021oct} includes 24,000 high-quality OCT images, labeled for 8-class classification. This dataset is used to detect a range of retinal conditions, including age-related macular degeneration (AMD), CNV, central serous retinopathy (CSR), DME, diabetic retinopathy (DR), drusen, macular hole (MH), and normal retinal conditions.
	
	\item \textbf{Pathology}: Pathology datasets include BACH~\cite{aresta2019bach} with 400 breast histopathology images for 4-class classification, distinguishing between normal tissue, benign lesions, in-situ carcinoma, and invasive carcinoma. LC25000\_colon~\cite{borkowski2019lung} comprises 10,000 colon tissue images for binary classification of adenocarcinoma and benign tissue. LC25000\_lung~\cite{borkowski2019lung} includes 15,000 lung tissue images for 3-class classification, identifying adenocarcinoma, benign tissue, and squamous cell carcinoma. NCT\_CRC~\cite{kather2018100} offers 107,180 colorectal cancer tissue patches for 9-class classification, covering various tissue types. Osteo~\cite{arunachalam2019viable} contains 1,091 osteosarcoma tissue tiles for 3-class classification, distinguishing non-tumor, necrotic, and viable tumor regions. PCAM~\cite{kawai2023large} provides 327,680 lymph node images for binary classification of metastatic breast cancer presence. Kather\_et\_al\_2016~\cite{kather2016multi} features 5,000 colorectal cancer histology images for 8-class classification of tissue types. Kather\_et\_al\_2018~\cite{kather2019predicting} includes 1,800 images for 9-class classification, derived from the NCT-CRC-HE-100K dataset. Kather\_et\_al\_2018\_val7k~\cite{kather2019predicting} is a validation set with 1,314 images for 9-class classification. Skin\_cancer~\cite{kriegsmann2022deep} comprises 6,107 skin histopathology tiles for 16-class classification, including 12 non-tumor and 4 tumor categories. Tang\_et\_al\_2019~\cite{tang2019interpretable} offers 491 brain tissue images for 4-class classification of amyloid pathologies. Wong\_et\_al\_2022~\cite{wong2022deep} provides 800 brain tissue images for 4-class classification of amyloid pathologies, based on multi-expert annotations.
	
	\item \textbf{Endoscopy}: Endoscopy datasets include Kvasir~\cite{Pogorelov:2017:KMI:3083187.3083212} with 8,000 gastrointestinal images for 8-class classification, covering anatomical landmarks and pathological findings, and WCE~\cite{montalbo2022colon} containing 6,000 wireless capsule endoscopy images across four categories, aimed at detecting normal tissue, ulcerative colitis, polyps, and esophagitis conditions.
	
	\item \textbf{Dermatology}: Dermatology datasets comprise HAM10000~\cite{tschandl2018ham10000} with 10,015 skin lesion images for 7-class classification and PADUFES20~\cite{pacheco2020pad} containing 2,298 smartphone-captured lesions across six categories.
	
\end{itemize}

For cross-modal retrieval, BookSet~\cite{gamper2021multiple} provides 4,265 image-text pairs from academic textbooks, while MedTrinity-25M subset~\cite{xie2024medtrinity} contains 3,771 multimodal medical samples. Visual question answering evaluation used VQA-RAD~\cite{lau2018dataset} with 3,515 radiological question-answer pairs and SLAKE~\cite{liu2021slake} containing 14,000 questions across 642 medical images (Only the English portion was utilized in this study). Supervised cancer diagnosis employed Camelyon16~\cite{bejnordi2017diagnostic} with 4,809,940 patches from 399 whole-slide lymph node images, TCGA-NSCLC~\cite{weinstein2013cancer} containing 4,377,051 patches from 1,046 lung cancer slides, and BRACS~\cite{brancati2022bracs} with 45,390 breast histopathology patches across seven lesion categories. Survival prediction was evaluated on TCGA-UT~\cite{komura2022universal} containing 138,220 patches from 12 cancer types using 5-fold cross-validation. Detailed descriptions of each dataset are provided in Supplementary Methods.

\subsection{Implementation details}

We initialize image encoder from MetaCLIP Vit-B/16~\cite{xu2023demystifying}, and initialize text encoder from BioMed-BERT~\cite{chakraborty2020biomedbert}. The size of each input image is 224$\times$224 pixels. During the pretraining phase, we use the ADAMW~\cite{loshchilov2017decoupled} optimizer with a learning rate of 5e-5. Training is conducted on two A100 GPUs with a batch size of 512 for 20 epochs, with the first 2000 steps serving as a warm-up phase. In the distillation stage, we ensure that each original image and text are traversed for 20 epochs, and each image-text pair extracts a teacher feature pair from its corresponding trustworthy teacher group for distillation. The batch size at this stage is 384. 

For linear probing image classification task, we used a single-layer fully connected neural network with the input dimension being the CLIP feature dimension and the output dimension being the number of categories in the downstream task dataset. We use the ADAMW optimizer with a learning rate of 5e-5. Training is conducted on one A6000 ADA GPU with a batch size of 128 for 20 epochs.
For VQA task, we kept the same settings as PubMedCLIP\footnote{\url{https://github.com/sarahESL/PubMedCLIP}}. The number of epoch is set to 150.
For survival prediction task, we estimate the survival function using the Kaplan-Meier method, dividing event times into equal probability intervals to obtain time interval cut points that reflect event density. Then, we used ADAM as the optimizer with a learning rate of 1e-4 and performed 50 epochs of training. When visualizing, we grouped individuals into high-risk and low-risk categories based on the median of the predicted risk score.
\subsection{Metrics}

We employ 7 widely-recognized evaluation metrics to assess MMKD-CLIP’s performance across classification, retrieval, VQA, and survival prediction tasks: AUC, Accuracy, Recall, C-Index, C-Index$_{td}$, IBS, and INBLL.

\paragraph{AUC} 
The AUC metric quantifies a classifier’s capability to distinguish between positive and negative classes across various decision thresholds. It is particularly important in medical classification tasks for its threshold-independent nature and ability to summarize diagnostic performance.

\paragraph{Accuracy} 
Accuracy (ACC) reflects the proportion of correct predictions over the total number of predictions made. It is especially relevant in visual question answering, where it captures the model's ability to choose the right answer from multiple choices. It is defined as:

\begin{equation}
\text{ACC} = \frac{\text{TP} + \text{TN}}{\text{TP} + \text{TN} + \text{FP} + \text{FN}}
\end{equation}

where TP, TN, FP, and FN denote true positives, true negatives, false positives, and false negatives respectively.

\paragraph{Recall}
Recall quantifies the model’s ability to retrieve all relevant items from the total pool of ground-truth positives. In our retrieval setting, we report Recall@K, where \( K \in \{1, 10, 50\} \), indicating the proportion of times the correct match appears within the top-\( K \) retrieved results. This helps assess performance at varying tolerance levels for top-ranked predictions. Formally, recall is defined as:

\begin{equation}
\text{Recall} = \frac{\text{TP}}{\text{TP} + \text{FN}}
\end{equation}

Higher Recall@1 suggests strong precision at top-1 prediction, while Recall@10 and Recall@50 measure broader retrieval robustness.

\paragraph{C-Index} 
C-Index assesses how well a model predicts the relative order of event times. It is commonly used in survival analysis to evaluate the model’s ranking capability. The C-Index is computed as the proportion of all comparable subject pairs for which the predicted risk scores and the actual outcomes are in agreement. The formal definition is:

\begin{equation}
\text{C-Index} = \frac{1}{|\mathcal{P}|} \sum_{(i,j) \in \mathcal{P}} \mathbb{1} \left( \hat{t}_i < \hat{t}_j \right)
\end{equation}

where \( \mathcal{P} \) is the set of comparable pairs such that subject \( i \) experienced the event before subject \( j \), \( \hat{t}_i \) and \( \hat{t}_j \) are the predicted risk scores or survival times, and \( \mathbb{1}(\cdot) \) is the indicator function that equals 1 if the predicted order matches the observed order. A C-Index of 1 indicates perfect concordance, whereas 0.5 implies random prediction.

\paragraph{C-Index$_{td}$}
The time-dependent concordance index (\( \text{C-Index}_{td} \)) generalizes the traditional C-Index by evaluating the model’s ability to rank event risks at multiple time horizons. Unlike the standard version, it incorporates time-specific risk estimates, making it suitable for dynamic prediction settings. It is typically defined as:

\begin{equation}
\text{C-Index}_{td} = \frac{1}{|\mathcal{P}_t|} \sum_{t \in \mathcal{T}} \sum_{(i,j) \in \mathcal{P}_t} \mathbb{1} \left( \hat{S}_i(t) < \hat{S}_j(t) \right)
\end{equation}

where \( \mathcal{P}_t \) denotes the set of comparable pairs at time \( t \), and \( \hat{S}_i(t) \) is the predicted survival probability for individual \( i \) at time \( t \).

\paragraph{IBS}
The Integrated Brier Score (IBS) measures the accuracy of survival probability estimates over a continuous time range. It computes the average of the squared differences between predicted survival probabilities and observed survival status, integrated across time. Lower values indicate better performance. The formula is:

\begin{equation}
\text{IBS} = \frac{1}{T} \int_0^T \mathbb{E} \left[ \left( \hat{S}(t \mid x) - \mathbb{1}(T_i > t) \right)^2 \right] dt
\end{equation}

where \( \hat{S}(t \mid x) \) is the predicted survival probability at time \( t \) for feature vector \( x \), and \( \mathbb{1}(T_i > t) \) is the event indicator.

\paragraph{INBLL}
The Integrated Negative Binomial Log-Likelihood (INBLL) quantifies the model’s probabilistic calibration over time by evaluating the negative log-likelihood of observed event times under the predicted discrete-time risk distribution. It integrates the performance across all time intervals:

\begin{equation}
\text{INBLL} = - \frac{1}{N} \sum_{i=1}^N \sum_{t=1}^T \left[ y_{i,t} \log p_{i,t} + (1 - y_{i,t}) \log(1 - p_{i,t}) \right]
\end{equation}

where \( y_{i,t} \) indicates whether patient \( i \) experienced the event at time \( t \), and \( p_{i,t} \) is the predicted probability of the event occurring at that time.

\section{Statistical analysis}
To evaluate model performance across different experimental settings, we adopted distinct statistical approaches based on the nature of the task. For scenarios involving independent test sets, either in zero-shot or fine-tuned evaluations, we applied a resampling-based approach using 1,000 non-parametric bootstrap replicates to construct 95\% confidence intervals. For experiments employing five-fold cross-validation, the 95\% confidence intervals were computed directly from the fold-wise outcomes. Statistical significance between groups was examined using non-parametric tests, namely the two-tailed Mann-Whitney U test or the two-tailed Wilcoxon signed-rank test, depending on the specific analysis (detailed in corresponding figure captions). In evaluating prognostic models for time-to-event outcomes, we used the C-Index as the primary performance metric. Patient stratification was visualized through KM survival plots, where individuals were split into high- and low-risk categories based on the median of the predicted risk scores. The separation between risk groups was tested for statistical significance using the log-rank test.

\section{Data Availability}
The data used for pre-training is available from the Huggingface project of BIOMEDICA\footnote{\url{https://huggingface.co/BIOMEDICA}}. The data used for distillation can be obtained from MMKD-CLIP\footnote{\url{https://github.com/wangshansong1/MMKD-CLIP}}. Links to downstream task datasets are provided in section \ref{82} of the Supporting Materials.

\section{Code Availability}
The pretrained models, as well as source code for training, inference, can be accessed at \url{https://github.com/wangshansong1/MMKD-CLIP}.

\subsection*{Acknowledgements} 

This research is supported in part by the National Institutes of Health under Award Number R56EB033332, R01EB032680, R01DE033512 and R01CA272991.

\subsection*{Contributions}

\textbf{Shansong Wang:} Writing-original draft, Methodology, Investigation,
Formal analysis, Data curation, Conceptualization.
\textbf{Zhecheng Jin:} Writing – Review \& Editing; Software; Visualization.
\textbf{Mingzhe Hu:} Resources, Writing-review \& editing.
\textbf{Mojtaba Safari:} Data curation, Writing-review \& editing. 
\textbf{Feng Zhao:} Software.
\textbf{Chih-Wei Chang:} Writing-review \& editing. 
\textbf{Richard LJ Qiu:} Writing-review \& editing. 
\textbf{Justin Roper:} Writing-review \& editing. 
\textbf{David S. Yu:} Writing-review \& editing. 
\textbf{Xiaofeng Yang:} Writing-review \& editing, Supervision, Resources, Project administration, Funding acquisition.

\bibliographystyle{unsrt}
\bibliography{newbib}

\clearpage

\newpage
\section{Supplementary materials}

\subsection{Model design, data distribution, and offline feature extraction pipeline}

We built MMKD-CLIP on a dual-stream transformer backbone comprising a ViT-B/16 image encoder (MetaCLIP)~\cite{xu2023demystifying} and a BioMed-BERT~\cite{chakraborty2020biomedbert} text encoder (Table~\ref{tab:multimodal_config}).  During contrastive pretraining, image-text pairs from PMC-OA database are aligned using standard InfoNCE loss; subsequent knowledge distillation further refines the joint space by matching student features to those of multiple high-accuracy teacher CLIP models.

Our filtered dataset spans over 2.91 million image-text pairs covering 26 clinical and research imaging modalities (Table~\ref{tab:modality_distribution}).  Modalities range from radiographic methods (X-ray, CT, MRI, ultrasound) to microscopic modalities (light, confocal, electron microscopy) and endoscopic procedures. The broad coverage ensures that MMKD-CLIP learns robust representations transferable across diverse diagnostic contexts.

\begin{figure}[!h] 
	\centering
	\includegraphics[width=\textwidth]{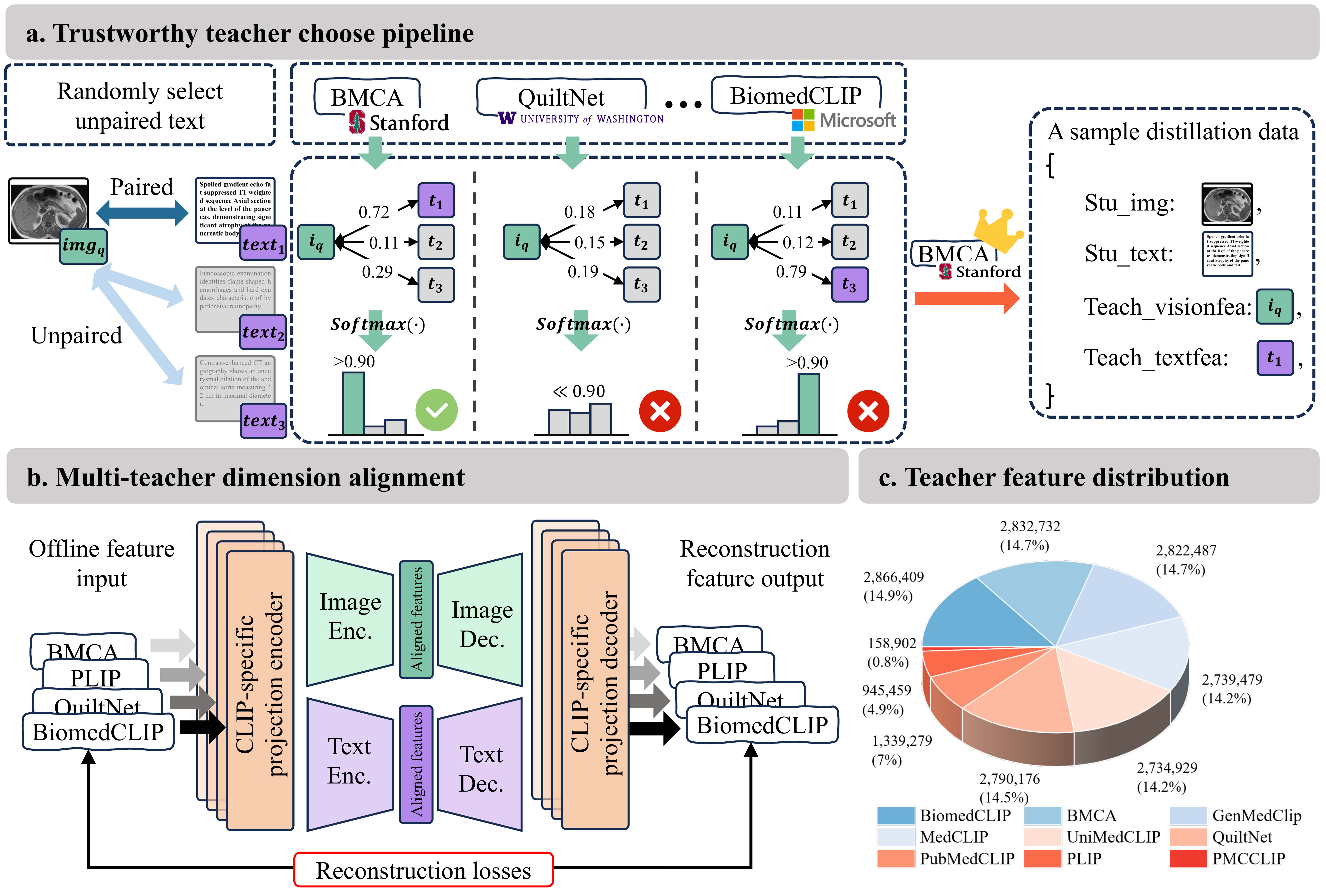}
	\caption{\textbf{Trustworthy teacher selection, multi-teacher feature alignment, and teacher feature distribution.} a. Trustworthy teacher selection pipeline. For each image-text pair, four additional unpaired text samples are randomly drawn and presented to each of nine biomedical CLIP teacher models via zero-shot classification. A teacher is deemed trustworthy for that pair only if its softmax score for the correct class exceeds 0.90 (green check). Trustworthy teachers then yield a distillation-ready quadruplet comprising the original image, the original text, and the teacher's visual and textual feature vectors. b. Multi-teacher dimension alignment. Offline feature vectors from all selected teachers pass through CLIP-specific projection encoders into a unified latent space. Shared dual-stream autoencoders (image and text streams) is used to learn a joint representation. Projected features are then decoded by CLIP-specific projection decoders to reconstruct the original teacher features, ensuring each teacher's native feature distribution. c. Teacher feature distribution. Pie chart showing the total number and proportion of quadruplet samples contributed by each teacher model (e.g., BMCA, PLIP, QuiltNet, BiomedCLIP, etc.), summing to 19,229,852 samples.}
	\label{fig:choosteacher} 
\end{figure}

For offline feature extraction, we first select trustworthy teachers per sample via zero-shot classification (Fig.~\ref{fig:choosteacher} a).  Each selected teacher’s visual and textual embeddings are then passed through CLIP-specific projection encoders into a unified latent space, encoded and decoded by shared dual-stream autoencoders to preserve native feature distributions (Fig.~\ref{fig:choosteacher} b).  Finally, reconstructed teacher features form the distillation quadruplets used to train the student model (Fig.~\ref{fig:choosteacher} c).

\begin{table}[h]
	\centering
	\small
	\caption{Configuration comparison network backbone.}
	\begin{tabular}{ccc}
		\hline
		Configuration Item & Vision Encoder (MetaCLIP ViT-B/16)~\cite{xu2023demystifying} & Text Encoder (BioMed-BERT)~\cite{chakraborty2020biomedbert} \\
		\hline
		Architecture & ViTModel & BertForMaskedLM \\
		Model Type & vit & bert \\
		Input Shape & 224 × 224 image & up to 512 tokens \\
		Patch Size & 16 × 16 & N/A \\
		Input Channels & 3 & N/A \\
		Vocabulary Size & N/A & 30,522 \\
		Hidden Size (Pre-Projection) & 768 & 768 \\
		Number of Transformer Layers & 12 & 12 \\
		Number of Attention Heads & 12 & 12 \\
		Intermediate Feedforward Size & 3072 & 3072 \\
		Activation Function & GELU & GELU \\
		Dropout (Hidden Layers) & 0.0 & 0.1 \\
		Dropout (Attention) & 0.0 & 0.1 \\
		Initialization Range & 0.02 & 0.02 \\
		Projection Features Size & 512 & 512 \\
		\hline
	\end{tabular}
	\label{tab:multimodal_config}
\end{table}

\begin{table}[h]
	\centering
	\caption{Distribution of image–text pairs across 26 imaging modalities in the filtered subset.}
	\begin{tabular}{llr|llr}
		\hline
		Modality & & Count & Modality & & Count \\
		\hline
		Angiography & & 43,544 & Intraoral imaging & & 40,753 \\
		Clinical imaging & & 129,373 & Laryngoscopy & & 12,622 \\
		Computerized tomography & & 390,182 & Light microscopy & & 490,939 \\
		Confocal microscopy & & 104,257 & Magnetic resonance & & 49,518 \\
		Electrocardiography & & 51,085 & Microscopy (unspecified) & & 504,537 \\
		Electron microscopy & & 137,366 & Optical coherence tomography & & 20,738 \\
		Endoscopy & & 79,586 & Phase contrast microscopy & & 9,519 \\
		Epifluorescence microscopy & & 25,467 & Scanning electron microscopy & & 52,879 \\
		Eye & & 14,106 & Skin lesion & & 127,202 \\
		Fluorescence microscopy & & 66,302 & Specimen & & 80,582 \\
		Functional magnetic resonance & & 49,461 & Surgical procedure & & 96,966 \\
		Intraoperative image & & 20,314 & Teeth & & 19,479 \\
		Ultrasound & & 94,862 & X-ray radiography & & 199,551 \\
		\hline
		\multicolumn{3}{l}{\textbf{Total: }} & \multicolumn{3}{r}{\textbf{2,911,190}}\\
		\hline
	\end{tabular}
	\label{tab:modality_distribution}
\end{table}

\clearpage
\newpage

\subsection{Benchmark datasets}
\label{82}
We conducted downstream tasks using 58 benchmark datasets on zero-shot classification, linear probing image classification, cross-modal retrieval,  visual question answering, and outcome prediction. We provide detailed data descriptions here.

\subsubsection{Zero-shot and linear probing image classification}
\begin{table}[h]
	\centering
	
	\caption{Statistics of BACH.}
	\begin{tabular}{|>{\raggedright\arraybackslash}p{3cm}|>{\raggedright\arraybackslash}p{11cm}|}
		\hline
		\multicolumn{2}{|c|}{\textbf{BACH \, (Pathology)}~\cite{aresta2019bach}} \\ \hline
		\textbf{Classes} & \{"Benign": 0, "InSitu": 1, "Invasive": 2, "Normal": 3\} \\ \hline
		\textbf{Splits} & \{"train": 200, "test": 100 , "val": 100\} \\ \hline
		\textbf{Description} & A dataset of 400 breast histopathology microscopy images in TIF format, labeled for 4-class classification to distinguish between normal breast tissue, benign lesions, in-situ carcinoma, and invasive carcinoma. \\ \hline
		\textbf{Prompts} & \{"Breast non-malignant benign tissue": ["a histopathology slide showing Breast non-malignant benign tissue", "histopathology image of Breast non-malignant benign tissue", "pathology tissue showing Breast non-malignant benign tissue", "presence of Breast non-malignant benign tissue on image"], "Breast malignant in-situ carcinoma": ["a histopathology slide showing Breast malignant in-situ carcinoma", "histopathology image of Breast malignant in-situ carcinoma", "pathology tissue showing Breast malignant in-situ carcinoma", "presence of Breast malignant in-situ carcinoma tissue on image"], "Breast malignant invasive carcinoma": ["a histopathology slide showing Breast malignant invasive carcinoma", "histopathology image of Breast malignant invasive carcinoma", "pathology tissue showing Breast malignant invasive carcinoma", "presence of Breast malignant invasive carcinoma tissue on image"], "Breast normal breast tissue": ["a histopathology slide showing Breast normal breast tissue", "histopathology image of Breast normal breast tissue", "pathology tissue showing Breast normal breast tissue", "presence of Breast normal breast tissue on image"]\} \\ \hline
		\textbf{Link} & \url{https://zenodo.org/records/3632035} \\ \hline
	\end{tabular}
	\label{zerobeggin}
\end{table}

\begin{table}[h]
	\centering
	\caption{Statistics of BUSBRA.}
	\begin{tabular}{|>{\raggedright\arraybackslash}p{3cm}|>{\raggedright\arraybackslash}p{11cm}|}
		\hline
		\multicolumn{2}{|c|}{\textbf{BUSBRA \, (Ultrasound)}~\cite{nastase2024role}} \\ \hline
		\textbf{Classes} & \{"malignant": 0, "benign": 1\} \\ \hline
		\textbf{Splits} & \{"train":1124, "test":369,"val":382\} \\ \hline
		\textbf{Description} & A dataset of 1,875 breast ultrasound images from 1,064 patients with biopsy-proven cases, labeled for binary classification to distinguish between malignant and benign breast lesions. \\ \hline
		\textbf{Prompts} & \{"malignant": ["Ultrasound image of the breast showing irregular margins and malignant features.","Malignant breast tumor with heterogeneous echotexture in ultrasound imaging.","Ultrasound evidence of invasive breast carcinoma.","Suspicious hypoechoic lesion suggestive of malignancy in breast scan.","Breast ultrasound reveals characteristics typical of a malignant mass."],"benign": ["Well-circumscribed lesion on ultrasound suggesting benign breast mass.","Breast ultrasound showing fibroadenoma with homogenous texture.","Benign-appearing breast lesion with posterior acoustic enhancement.","Ultrasound features consistent with a non-malignant breast condition.","Hypoechoic oval mass with smooth borders indicating benign lesion."]\} \\ \hline
		\textbf{Link} & \url{https://zenodo.org/records/8231412} \\ \hline
	\end{tabular}
\end{table}

\begin{table}[h]
	\centering
	\caption{Statistics of BrainTumorCT.}
	\begin{tabular}{|>{\raggedright\arraybackslash}p{3cm}|>{\raggedright\arraybackslash}p{11cm}|}
		\hline
		\multicolumn{2}{|c|}{\textbf{BrainTumorCT \, (CT)}~\cite{likhon2023brain}} \\ \hline
		\textbf{Classes} & \{"Healthy": 0, "Tumor": 1\} \\ \hline
		\textbf{Splits} & \{"train":2770, "test":925,"val":923\} \\ \hline
		\textbf{Description} & A dataset of 4,618 high-resolution brain CT scans from multiple patients, labeled for binary classification to distinguish between healthy brain tissue and the presence of brain tumors. The dataset is part of a larger multimodal collection that leverages CT imaging for clear visualization of brain structures and abnormal masses, designed to support the development of AI models for automatic brain tumor detection and analysis. \\ \hline
		\textbf{Prompts} & \{"Healthy": [ "Brain CT showing normal parenchymal structures and ventricles.", "No abnormal densities or masses noted in the brain CT image.", "CT scan of brain reveals normal anatomy without pathology.", "Normal appearance of cerebral hemispheres in CT scan.", "No signs of intracranial abnormality in this brain CT." ], "Tumor": [ "CT scan showing space-occupying lesion within the brain parenchyma.", "Radiographic evidence of brain tumor with mass effect.", "Irregular enhancing lesion visible in cerebral CT image.", "Brain CT displaying abnormal density consistent with tumor growth.", "CT image reveals intracranial mass with surrounding edema."]\} \\ \hline
		\textbf{Link} & \url{https://www.kaggle.com/datasets/murtozalikhon/brain-tumor-multimodal-image-ct-and-mri} \\ \hline
	\end{tabular}
\end{table}

\begin{table}[h]
	\centering
	\caption{Statistics of BrainTumorMRI.}
	\begin{tabular}{|>{\raggedright\arraybackslash}p{3cm}|>{\raggedright\arraybackslash}p{11cm}|}
		\hline
		\multicolumn{2}{|c|}{\textbf{BrainTumorMRI \, (MRI)}~\cite{likhon2023brain}} \\ \hline
		\textbf{Classes} & \{"Healthy": 0, "Tumor": 1\} \\ \hline
		\textbf{Splits} & \{"train":3000, "test":1000,"val":1000\} \\ \hline
		\textbf{Description} & A dataset of 5,000 high-resolution brain MRI scans from multiple patients, labeled for binary classification to distinguish between healthy brain tissue and the presence of brain tumors. This dataset represents the MRI portion of a larger multimodal collection, leveraging MRI's superior soft tissue contrast and detail to provide comprehensive insight into brain tumor characteristics and enabling more accurate analysis for AI model development in automatic brain tumor detection. \\ \hline
		\textbf{Prompts} & \{"Healthy": [ "Brain MRI shows normal cortical and subcortical anatomy.", "No evidence of lesions or abnormal signal on MRI.", "MRI image of brain presenting healthy cerebral structures.", "No abnormal enhancement seen in this brain MRI.", "MRI scan confirms normal intracranial appearance.", ], "Tumor": [ "MRI reveals contrast-enhancing brain tumor with mass effect.", "Abnormal hyperintense lesion on T2-weighted MRI suggestive of tumor.", "Brain MRI image showing space-occupying tumor with edema.", "Malignant lesion detected in cerebral region on MRI.", "MRI shows a lobulated mass consistent with brain tumor pathology."]\} \\ \hline
		\textbf{Link} & \url{https://www.kaggle.com/datasets/murtozalikhon/brain-tumor-multimodal-image-ct-and-mri} \\ \hline
	\end{tabular}
\end{table}

\begin{table}[h]
	\centering
	\caption{Statistics of BrainTumorMRI2.}
	\begin{tabular}{|>{\raggedright\arraybackslash}p{3cm}|>{\raggedright\arraybackslash}p{11cm}|}
		\hline
		\multicolumn{2}{|c|}{\textbf{BrainTumorMRI2 \, (MRI)}~\cite{msoud_nickparvar_2021}} \\ \hline
		\textbf{Classes} & \{"glioma": 0, "meningioma": 1, "notumor": 2, "pituitary": 3\} \\ \hline
		\textbf{Splits} & \{"train":4285, "test":1311,"val":1427\} \\ \hline
		\textbf{Description} & A dataset of 7,023 brain MRI images labeled for 4-class classification to distinguish between glioma, meningioma, pituitary tumors, and healthy brain tissue without tumors. \\ \hline
		\textbf{Prompts} & \{"glioma": [ "A brain MRI image showing glioma.", "MRI scan reveals an irregular mass consistent with glioma.", "Hyperintense lesion in cerebral parenchyma indicating glioma.", "Infiltrative brain tumor with contrast enhancement.", "Image showing high-grade glioma with peritumoral edema." ], "meningioma": [ "A brain MRI image showing meningioma.", "Well-defined extra-axial mass suggestive of meningioma.", "MRI showing dural-based lesion with homogeneous enhancement.", "Image of benign brain tumor arising from meninges.", "Tumor compressing adjacent brain structures consistent with meningioma." ], "notumor": [ "MRI image of a healthy brain without tumor.", "No abnormal mass or lesion detected in brain MRI.", "Normal MRI scan with no signs of neoplasm.", "Intact brain parenchyma without pathological findings.", "Clean anatomical structures in a tumor-free MRI image." ], "pituitary": [ "MRI scan showing pituitary tumor in sella region.", "Hyperintense mass in pituitary gland indicating adenoma.", "Enlargement of pituitary with suprasellar extension visible.", "Contrast-enhanced image showing pituitary lesion.", "Image consistent with microadenoma or macroadenoma of pituitary."]\} \\ \hline
		\textbf{Link} & \url{https://www.kaggle.com/datasets/masoudnickparvar/brain-tumor-mri-dataset} \\ \hline
	\end{tabular}
\end{table}

\begin{table}[h]
	\centering
	\caption{Statistics of CovidCXR4.}
	\begin{tabular}{|>{\raggedright\arraybackslash}p{3cm}|>{\raggedright\arraybackslash}p{11cm}|}
		\hline
		\multicolumn{2}{|c|}{\textbf{CovidCXR4 \, (X-Ray)}~\cite{wang2020covid}} \\ \hline
		\textbf{Classes} & \{"No Finding": 0, "Covid-19": 1\} \\ \hline
		\textbf{Splits} & \{"train": 67863, "test": 8482, "val": 8473\} \\ \hline
		\textbf{Description} & A subset of the COVIDx CXR-4 collection comprising 84,818 chest X-ray images from 45,342 patients across multiple institutions, labeled for binary classification to detect COVID-19 infection versus normal findings. \\ \hline
		\textbf{Prompts} & \{"No Finding": [ "Chest radiograph appears normal with no radiographic abnormalities.", "This X-ray shows no signs of disease in the thoracic cavity.", "Radiographic assessment demonstrates no detectable pathology.", "Clear lungs without any abnormalities seen in the chest X-ray." ], "Covid-19": [ "Chest radiograph showing bilateral infiltrates typical of COVID-19 pneumonia.", "X-ray reveals diffuse ground-glass opacities associated with COVID-19.", "Radiographic features suggest COVID-19 infection.", "Evidence of COVID-19 pneumonia seen on chest radiograph.", "Chest X-ray consistent with viral pneumonia due to SARS-CoV-2."]\} \\ \hline
		\textbf{Link} & \url{https://www.kaggle.com/datasets/andyczhao/covidx-cxr2} \\ \hline
	\end{tabular}
\end{table}

\begin{table}[h]
	\centering
	\caption{Statistics of CovidCT3A.}
	\begin{tabular}{|>{\raggedright\arraybackslash}p{3cm}|>{\raggedright\arraybackslash}p{11cm}|}
		\hline
		\multicolumn{2}{|c|}{\textbf{CovidCT3A \, (CT)}~\cite{gunraj2022covid}} \\ \hline
		\textbf{Classes} & \{"Normal": 0, "Pneumonia": 1, "COVID-19": 2\} \\ \hline
		\textbf{Splits} & \{"train": 357518, "test": 33781, "val": 33725\} \\ \hline
		\textbf{Description} & A large-scale dataset comprising 425,024 chest CT slices from 5,312 patients with confirmed diagnoses, labeled for 3-class classification to distinguish between normal lungs, pneumonia, and COVID-19 infection. This dataset represents the "A" variant of the COVIDx CT-3 collection from the COVID-Net Open Source Initiative, compiled from several open datasets and specifically designed for training and validating AI models for COVID-19 detection from chest CT imaging, with cases verified through RT-PCR testing or radiologist confirmation. \\ \hline
		\textbf{Prompts} & \{"Normal": [ "CT image of the chest showing normal lung parenchyma.", "No abnormal opacities seen in the lung fields on CT.", "Normal chest CT scan without signs of infection or pathology.", "Clear lungs and bronchial structures in this CT scan." ], "Pneumonia": [ "CT scan shows patchy consolidation in the lungs indicating pneumonia.", "Lung opacities consistent with bacterial or viral pneumonia.", "Chest CT with ground-glass opacity and consolidation typical of pneumonia.", "CT image showing localized lung inflammation due to pneumonia.", "Pulmonary CT scan reveals segmental consolidation from infection." ], "COVID-19": [ "CT scan shows bilateral ground-glass opacities typical of COVID-19.", "Diffuse patchy opacities in lungs suggestive of SARS-CoV-2 infection.", "Chest CT displaying features consistent with COVID-19 pneumonia.", "CT imaging reveals peripherally distributed opacities linked to COVID-19.", "Radiological signs of COVID-19 pneumonia seen in chest CT."]\} \\ \hline
		\textbf{Link} & \url{https://www.kaggle.com/datasets/hgunraj/covidxct} \\ \hline
	\end{tabular}
\end{table}

\begin{table}[h]
	\centering
	\caption{Statistics of DDSM.}
	\begin{tabular}{|>{\raggedright\arraybackslash}p{3cm}|>{\raggedright\arraybackslash}p{11cm}|}
		\hline
		\multicolumn{2}{|c|}{\textbf{DDSM \, (X-Ray)}~\cite{heath2001ddsm}} \\ \hline
		\textbf{Classes} & \{"negative": 0, "positive": 1\} \\ \hline
		\textbf{Splits} & \{"train": 55885, "test": 7682, "val": 7682\} \\ \hline
		\textbf{Description} & A mammography dataset comprising 71,249 pre-processed breast X-ray images combining data from the DDSM and CBIS-DDSM collections. Images are standardized to 299×299 pixels and labeled for binary classification to detect breast cancer, distinguishing between positive findings (malignant/suspicious lesions) and negative findings (normal breast tissue). The dataset includes 55,885 training images with 14\% positive cases, designed for computer-aided diagnosis and detection systems in mammographic screening. \\ \hline
		\textbf{Prompts} & \{"negative": [ "Mammography image showing no evidence of malignancy.", "Breast tissue appears normal with no suspicious lesions.", "Negative mammogram with no radiographic signs of cancer.", "No mass or calcification noted in the mammographic image.", "Normal breast density with no abnormal findings." ], "positive": [ "Mammographic image showing suspicious mass suggestive of malignancy.", "Presence of clustered microcalcifications in the breast image.", "Breast X-ray reveals irregular opacity, indicating potential cancer.", "Mammogram with architectural distortion consistent with a lesion.", "Radiographic evidence of malignant findings in breast tissue."]\} \\ \hline
		\textbf{Link} & \url{https://www.kaggle.com/datasets/skooch/ddsm-mammography} \\ \hline
	\end{tabular}
\end{table}

\begin{table}[h]
	\centering
	\caption{Statistics of CT\_axial, CT\_coronal, CT\_sagittal.}
	\begin{tabular}{|>{\raggedright\arraybackslash}p{3cm}|>{\raggedright\arraybackslash}p{11cm}|}
		\hline
		\multicolumn{2}{|c|}{\textbf{CT\_axial, CT\_coronal, CT\_sagittal \, (CT)}~\cite{woerner2024comprehensive}} \\ \hline
		\textbf{Classes} & \{"heart": 0, "left lung": 1, "right lung": 2, "liver": 3, "spleen": 4, "pancreas": 5, "left kidney": 6, "right kidney": 7,  "bladder": 8, "left femoral head": 9,  "right femoral head": 10\} \\ \hline
		\textbf{Splits} & \{"train":871, "test":618,"val":156\} \\ \hline
		\textbf{Description} & Three subsets from the MedIMeta collection comprising 1,645 standardized abdominal CT image slices in axial, coronal, and sagittal viewing planes, extracted from the LiTS dataset and labeled for 11-class organ classification to identify anatomical structures including heart, lungs, liver, kidneys, spleen, pancreas, bladder, and femoral heads. \\ \hline
		\textbf{Prompts} & \{"heart": [ "CT scan image showing the anatomical structures of the heart, including the chambers and major blood vessels.", "Detailed CT scan image of the heart displaying the myocardium and pericardium.", "CT scan revealing the coronary arteries and their branches within the heart.", ... ], "left lung": [ "CT scan image displaying the lobes and segments of the left lung.", "Detailed CT scan image of the left lung highlighting the bronchial tree.", "CT scan revealing the parenchymal structures of the left lung.", ... ], "right lung": [ "CT scan image showing the lobes and segments of the right lung.", "Detailed CT scan image of the right lung highlighting the bronchial tree.", "CT scan revealing the parenchymal structures of the right lung.", ... ], "liver": [ "CT scan image displaying the anatomical lobes and segments of the liver.", "Detailed CT scan image of the liver showing the hepatic vasculature.", "CT scan revealing the liver parenchyma and surrounding structures.", ... ], "spleen": [ "CT scan image displaying the anatomical location and size of the spleen.", "Detailed CT scan image of the spleen highlighting its parenchymal structure.", "CT scan revealing the splenic vasculature, including the splenic artery and vein.", ... ], "pancreas": [ "CT scan image displaying the anatomical structure of the pancreas, including the head, body, and tail.", "Detailed CT scan image of the pancreas showing the pancreatic duct.", "CT scan revealing the relationship of the pancreas to the duodenum.", ... ], "left kidney": [ "CT scan image displaying the anatomical structure of the left kidney.", "Detailed CT scan image of the left kidney showing the renal cortex and medulla.", "CT scan revealing the renal vasculature, including the renal artery and vein.", ... ], "right kidney": [ "CT scan image displaying the anatomical structure of the right kidney.", "Detailed CT scan image of the right kidney showing the renal cortex and medulla.", "CT scan revealing the renal vasculature, including the renal artery and vein.", ... ], "bladder": [ "CT scan image displaying the anatomical structure of the bladder.", "Detailed CT scan image of the bladder showing the bladder wall and lumen.", "CT scan revealing the relationship of the bladder to the surrounding pelvic organs.", ... ], "left femoral head": [ "CT scan image displaying the anatomical structure of the left femoral head.", "Detailed CT scan image of the left femoral head showing the articular surface.", "CT scan revealing the relationship of the left femoral head to the acetabulum.", ... ], "right femoral head": [ "CT scan image displaying the anatomical structure of the right femoral head.", "Detailed CT scan image of the right femoral head showing the articular surface.", "CT scan revealing the relationship of the right femoral head to the acetabulum.", ...]\} \\ \hline
		\textbf{Link} & \url{https://arxiv.org/abs/2404.16000} \\ \hline
	\end{tabular}
\end{table}

\begin{table}[h]
	\centering
	\caption{Statistics of LC25000\_colon.}
	\begin{tabular}{|>{\raggedright\arraybackslash}p{3cm}|>{\raggedright\arraybackslash}p{11cm}|}
		\hline
		\multicolumn{2}{|c|}{\textbf{LC25000\_colon \, (Pathology)}~\cite{borkowski2019lung}} \\ \hline
		\textbf{Classes} & \{"colon\_aca": 0, "colon\_n": 1\} \\ \hline
		\textbf{Splits} & \{"train": 6000, "test": 2000, "val": 2000\} \\ \hline
		\textbf{Description} & A histopathological image dataset comprising 10,000 colon tissue images at 768×768 pixels, labeled for binary classification to distinguish between colon adenocarcinoma and benign colonic tissue. \\ \hline
		\textbf{Prompts} & \{"Colon adenocarcinoma": [ "Histopathology slide of Colon adenocarcinoma with glandular structures", "Histopathology image showing Colon adenocarcinoma cells with mucinous differentiation", "Pathology tissue sample of Colon adenocarcinoma with invasion into surrounding tissue" ], "Benign colonic tissue": [ "Histopathology slide of Benign colonic tissue with normal mucosal architecture", "Histopathology image showing Benign colonic tissue with no dysplastic changes", "Pathology tissue sample of Benign colonic tissue without malignant features"]\} \\ \hline
		\textbf{Link} & \url{https://github.com/tampapath/lung_colon_image_set} \\ \hline
	\end{tabular}
\end{table}

\begin{table}[h]
	\centering
	\caption{Statistics of DRD.}
	\begin{tabular}{|>{\raggedright\arraybackslash}p{3cm}|>{\raggedright\arraybackslash}p{11cm}|}
		\hline
		\multicolumn{2}{|c|}{\textbf{DRD \, (Fundus)}~\cite{dugas2015drd}} \\ \hline
		\textbf{Classes} & \{"No diabetic retinopathy": 0, "Mild diabetic retinopathy": 1, "Moderate diabetic retinopathy": 2, "Severe diabetic retinopathy": 3, ‘Proliferative diabetic retinopathy’: 4\} \\ \hline
		\textbf{Splits} & \{"train": 21072, "test": 7023, "val": 7030\} \\ \hline
		\textbf{Description} & A fundus image dataset comprising 35,125 retinal photographs labeled for 5-class diabetic retinopathy severity grading, ranging from no diabetic retinopathy to proliferative diabetic retinopathy. \\ \hline
		\textbf{Prompts} & \{"No diabetic retinopathy": [ "Fundus image showing no signs of diabetic retinopathy.", "Retina appears healthy with intact vasculature and no lesions.", "No microaneurysms or hemorrhages present in retinal image.", "Normal optic disc and macula visualized in the fundus photo.", "Retinal examination free from diabetic changes." ], "Mild diabetic retinopathy": [ "Scattered microaneurysms seen in early diabetic retinopathy.", "Retinal image showing mild non-proliferative changes.", "Fundus photo revealing small hemorrhages characteristic of mild DR.", "Initial signs of vascular leakage seen in retina.", "Early diabetic changes with minimal impact on vision." ], "Moderate diabetic retinopathy": [ "Retina showing cotton wool spots and vascular beading.", "Moderate diabetic retinopathy with visible intraretinal hemorrhages.", "Fundus image indicates increased number of microaneurysms and exudates.", "Retinal pathology consistent with moderate stage of DR.", "Image showing mid-stage diabetic retinal complications." ], "Severe diabetic retinopathy": [ "Severe non-proliferative DR with widespread retinal damage.", "Numerous hemorrhages and venous abnormalities in retina.", "Advanced retinal pathology indicative of severe DR.", "High-risk pre-proliferative diabetic retinopathy seen in image.", "Severe retinal changes with risk of neovascularization." ], "Proliferative diabetic retinopathy": [ "Neovascularization visible in proliferative diabetic retinopathy.", "Fundus photo shows abnormal new vessels and fibrous proliferation.", "Vitreous hemorrhage present due to proliferative DR.", "Advanced diabetic eye disease with traction retinal detachment risk.", "Retinal image reveals features of proliferative diabetic changes."]\} \\ \hline
		\textbf{Link} & \url{https://www.kaggle.com/competitions/diabetic-retinopathy-detection} \\ \hline
	\end{tabular}
\end{table}

\begin{table}[h]
	\centering
	\caption{Statistics of NLMTB.}
	\begin{tabular}{|>{\raggedright\arraybackslash}p{3cm}|>{\raggedright\arraybackslash}p{11cm}|}
		\hline
		\multicolumn{2}{|c|}{\textbf{NLMTB \, (X-Ray)}~\cite{jaeger2014two}} \\ \hline
		\textbf{Classes} & \{"Normal": 0, "Tuberculosis": 1\} \\ \hline
		\textbf{Splits} & \{"train": 2520, "test": 840, "val": 840\} \\ \hline
		\textbf{Description} & A chest X-ray dataset comprising 4,200 postero-anterior radiographs from the U.S. National Library of Medicine, labeled for binary classification to detect tuberculosis versus normal chest findings. \\ \hline
		\textbf{Prompts} & \{"Normal": ["a close-up view of a chest x-ray presenting evidence of normal", "a visual analysis of a chest x-ray with signs of normal", "a radiographic scan highlighting the presence of normal", "an annotated image from a chest x-ray showing signs of normal", ], "Tuberculosis": [ "a close-up view of a chest x-ray presenting evidence of tuberculosis", "a visual analysis of a chest x-ray with signs of tuberculosis", "a radiographic scan highlighting the presence of tuberculosis", "an annotated image from a chest x-ray showing signs of tuberculosis"]\} \\ \hline
		\textbf{Link} & \url{https://pmc.ncbi.nlm.nih.gov/articles/PMC4256233/} \\ \hline
	\end{tabular}
\end{table}

\begin{table}[h]
	\centering
	\caption{Statistics of OCTMNIST.}
	\begin{tabular}{|>{\raggedright\arraybackslash}p{3cm}|>{\raggedright\arraybackslash}p{11cm}|}
		\hline
		\multicolumn{2}{|c|}{\textbf{OCTMNIST \, (OCT)}~\cite{yang2023medmnist}} \\ \hline
		\textbf{Classes} & \{"CNV": 0, "DME": 1, "DRUSEN": 2, "NORMAL": 3\} \\ \hline
		\textbf{Splits} & \{"train": 97477, "test": 1000, "val": 10832\} \\ \hline
		\textbf{Description} & A retinal OCT image dataset comprising 109,309 optical coherence tomography images from the MedMNIST v2 collection, pre-processed to standardized 28×28 pixel grayscale format for lightweight classification. Images are labeled for 4-class classification to detect choroidal neovascularization (CNV), diabetic macular edema (DME), drusen, and normal retinal conditions. \\ \hline
		\textbf{Prompts} & \{"choroidal neovascularization": [ "OCT image showing subretinal fluid and CNV membranes.", "Choroidal neovascularization with hyperreflective lesion beneath retina.", "Image consistent with macular CNV causing retinal elevation.", "OCT reveals abnormal blood vessel growth under RPE.", "Subretinal hyperreflectivity indicating CNV in macular region." ], "diabetic macular edema": [ "OCT scan showing cystoid spaces within the retina.", "Diabetic macular edema with retinal thickening and fluid pockets.", "Image reveals intraretinal cysts and macular swelling.", "OCT showing fluid accumulation due to diabetic retinopathy.", "Macular region appears edematous with hyporeflective cavities." ], "drusen": [ "OCT image showing sub-RPE elevations consistent with drusen.", "Drusenoid deposits forming small bumps beneath retinal layers.", "OCT reveals early AMD signs with RPE detachment.", "Hyperreflective subretinal accumulations representing drusen.", "Image shows undulating RPE contour due to drusen." ], "normal": [ "Normal OCT scan with clearly defined retinal layers.", "Healthy macula with no pathological findings.", "Retinal architecture intact without fluid or deposits.", "OCT shows normal foveal contour and reflectivity.", "Image consistent with anatomically normal retina."]\} \\ \hline
		\textbf{Link} & \url{https://medmnist.com/} \\ \hline
	\end{tabular}
\end{table}

\begin{table}[h]
	\centering
	\caption{Statistics of FundusJSIEC.}
	\begin{tabular}{|>{\raggedright\arraybackslash}p{2cm}|>{\raggedright\arraybackslash}p{13cm}|}
		\hline
		\multicolumn{2}{|c|}{\textbf{FundusJSIEC \, (Fundus)}~\cite{cen2021automatic}} \\ \hline
		\textbf{Classes} & \{"0.0.Normal": 0, "0.1.Tessellated fundus": 1, "0.2.Large optic cup": 2, "0.3.DR1": 3, "1.0.DR2": 4, "1.1.DR3": 5, "2.0.BRVO": 6, "2.1.CRVO": 7, "3.RAO": 8, "4.Rhegmatogenous RD": 9, "5.0.CSCR": 10, "5.1.VKH disease": 11, "6.Maculopathy": 12, "7.ERM": 13, "8.MH": 14, "9.Pathological myopia": 15, "10.0.Possible glaucoma": 16, "10.1.Optic atrophy": 17, "11.Severe hypertensive retinopathy": 18, "12.Disc swelling and elevation": 19, "13.Dragged Disc": 20, "14.Congenital disc abnormality": 21, "15.0.Retinitis pigmentosa": 22, "15.1.Bietti crystalline dystrophy": 23, "16.Peripheral retinal degeneration and break": 24, "17.Myelinated nerve fiber": 25, "18.Vitreous particles": 26, "19.Fundus neoplasm": 27, "20.Massive hard exudates": 28, "21.Yellow-white spots-flecks": 29, "22.Cotton-wool spots": 30, "23.Vessel tortuosity": 31, "24.Chorioretinal atrophy-coloboma": 32, "25.Preretinal hemorrhage": 33, "26.Fibrosis": 34, "27.Laser Spots": 35, "28.Silicon oil in eye": 36, "29.0.Blur fundus without PDR": 37, "29.1.Blur fundus with suspected PDR": 38\} \\ \hline
		\textbf{Splits} & \{"train": 585, "test": 182, "val": 233\} \\ \hline
		\textbf{Description} & A fundus image dataset comprising 1,000 retinal photographs from the Joint Shantou International Eye Centre, labeled for 39-class classification covering diverse retinal diseases and conditions including diabetic retinopathy, glaucoma, macular disorders, and vascular occlusions. \\ \hline
		\textbf{Prompts} & \{"0.0.Normal": [ "Fundus image showing normal retinal anatomy with no abnormalities.", ... ], "0.1.Tessellated fundus": [ "Fundus image showing tessellated appearance with visible choroidal vessels.", ... ], "0.2.Large optic cup": [ "Fundus image showing an enlarged optic cup relative to the disc.", ... ], "0.3.DR1": [ "Early signs of diabetic retinopathy including microaneurysms observed.", ... ], "1.0.DR2": [ "Moderate non-proliferative diabetic retinopathy with increased hemorrhages.", ... ], "1.1.DR3": [ "Severe non-proliferative DR with extensive hemorrhages and venous beading.", ... ], "2.0.BRVO": [ "Fundus showing sectoral retinal hemorrhages consistent with BRVO.", ... ], "2.1.CRVO": [ "Central retinal vein occlusion with widespread hemorrhages.", ... ], "3.RAO": [ "Retinal artery occlusion evident with pale retina and cherry-red spot.", ... ], "4.Rhegmatogenous RD": [ "Retinal detachment with visible retinal break noted.", ... ], "5.0.CSCR": [ "Serous retinal detachment due to central serous chorioretinopathy.", ... ], "5.1.VKH disease": [ "Vogt-Koyanagi-Harada disease with serous retinal detachments.", ... ], "6.Maculopathy": [ "Macular degeneration with central vision distortion.", ... ], "7.ERM": [ "Epiretinal membrane visible as glistening layer over macula.", ... ], "8.MH": [ "Full-thickness macular hole present in the central retina.", ... ], "9.Pathological myopia": [ "Posterior staphyloma and chorioretinal atrophy from high myopia.", ... ], "10.0.Possible glaucoma": [ "Increased cup-to-disc ratio suggesting glaucoma risk.", ... ], "10.1.Optic atrophy": [ "Pale optic disc indicative of optic nerve atrophy.", ... ], "11.Severe hypertensive retinopathy": [ "Arteriolar narrowing and hemorrhages due to severe hypertension.", ... ], "12.Disc swelling and elevation": [ "Optic disc shows swelling with blurred margins.", ... ], "13.Dragged Disc": [ "Optic disc displaced from its usual position.", ... ], "14.Congenital disc abnormality": [ "Optic disc coloboma or hypoplasia visible.", ... ], "15.0.Retinitis pigmentosa": [ "Bone spicule pigmentation seen in peripheral retina.", ... ], "15.1.Bietti crystalline dystrophy": [ "Crystalline deposits in retina seen in Bietti dystrophy.", ... ], "16.Peripheral retinal degeneration and break": [ "Fundus image showing lattice degeneration in periphery.", ... ], "17.Myelinated nerve fiber": [ "Bright white striated patches along retinal nerve fibers.", ... ], "18.Vitreous particles": [ "Floaters or particulate matter visible in vitreous cavity.", ... ], "19.Fundus neoplasm": [ "Mass lesion noted in retina consistent with neoplasm.", ... ], "20.Massive hard exudates": [ "Dense lipid deposits forming star-shaped pattern.", ... ], "21.Yellow-white spots-flecks": [ "Small yellow flecks dispersed across retina.", ... ], "22.Cotton-wool spots": [ "White fluffy retinal lesions indicating nerve fiber infarction.", ... ], "23.Vessel tortuosity": [ "Retinal vessels appear abnormally twisted and dilated.", ... ], "24.Chorioretinal atrophy-coloboma": [ "Fundus shows area of chorioretinal thinning or coloboma.", ... ], "25.Preretinal hemorrhage": [ "Boat-shaped hemorrhage seen anterior to retina.", ... ], "26.Fibrosis": [ "Fibrotic band formation on retinal surface.", ... ], "27.Laser Spots": [ "Round, white retinal scars from panretinal photocoagulation.", ... ], "28.Silicon oil in eye": [ "Hyperreflective bubbles suggesting silicone oil tamponade.", ... ], "29.0.Blur fundus without PDR": [ "Blurred fundus image without signs of proliferative DR.", ... ], "29.1.Blur fundus with suspected PDR": [ "Poor image quality but possible neovascularization present.", ...]\} \\ \hline
		\textbf{Link} & \url{https://www.kaggle.com/datasets/linchundan/fundusimage1000} \\ \hline
	\end{tabular}
\end{table}

\begin{table}[h]
	\centering
	\caption{Statistics of HAM10000.}
	\begin{tabular}{|>{\raggedright\arraybackslash}p{3cm}|>{\raggedright\arraybackslash}p{11cm}|}
		\hline
		\multicolumn{2}{|c|}{\textbf{HAM10000 \, (Dermatology)}~\cite{tschandl2018ham10000}} \\ \hline
		\textbf{Classes} & \{"akiec": 0, "bcc": 1, "bkl": 2, "df": 3, "mel": 4, "nv": 5, "vasc": 6\} \\ \hline
		\textbf{Splits} & \{"train": 6007, "test": 2000, "val": 2008\} \\ \hline
		\textbf{Description} & A dermatoscopic image dataset comprising 10,015 images of pigmented skin lesions labeled for 7-class classification including melanoma, basal cell carcinoma, melanocytic nevi, and other common skin conditions. \\ \hline
		\textbf{Prompts} & \{"Actinic keratoses and intraepithelial carcinoma / Bowen"s disease": [ "Dermatological image showing actinic keratosis with intraepithelial carcinoma features.", "Clinical presentation of Bowen’s disease in dermatological imaging.", "Skin lesion consistent with actinic keratosis and carcinoma in situ.", "Surface crusting and erythema characteristic of intraepithelial carcinoma.", "Dermatoscopic image revealing early squamous neoplasia." ], "basal cell carcinoma": [ "Dermatological image showing nodular basal cell carcinoma.", "Skin lesion with pearly borders and telangiectasia typical of basal cell carcinoma.", "Basal cell carcinoma presenting as a localized raised skin nodule.", "Clinical image of pigmented basal cell carcinoma on sun-exposed area.", "Dermatoscopic image consistent with basal cell carcinoma diagnosis." ], "benign keratosis-like lesions": [ "Skin image showing benign keratosis with hyperkeratotic surface.", "Dermatological image of seborrheic keratosis-like lesion.", "Lesion with wart-like appearance indicating benign keratosis.", "Benign epidermal growth seen in dermatological photography.", "Crusty, scaly lesion suggesting benign keratosis-like features." ], "dermatofibroma": [ "Firm skin nodule consistent with dermatofibroma.", "Pigmented lesion with central dimpling indicating dermatofibroma.", "Clinical presentation of dermatofibroma on the lower extremities.", "Dermatoscopic features of fibrous skin lesion indicating dermatofibroma.", "Localized benign fibrous growth characteristic of dermatofibroma." ], "melanoma": [ "Asymmetrical pigmented lesion with irregular borders suggesting melanoma.", "Dermatological image showing features consistent with cutaneous melanoma.", "Melanoma with variegated pigmentation and rapid change in size.", "Highly suspicious lesion with ABCD features of melanoma.", "Dermatoscopic image showing atypical pigment network of melanoma." ], "melanocytic nevi": [ "Benign mole with uniform color and symmetry indicating melanocytic nevus.", "Common nevus seen in dermatoscopic examination.", "Flat or raised pigmented lesion consistent with melanocytic nevus.", "Dermatological image of a benign melanocytic nevus.", "Uniform pigmentation and smooth borders suggesting a nevus." ], "vascular lesions": [ "Red or purple lesion indicative of a vascular malformation.", "Dermatological image showing cherry angioma or hemangioma.", "Vascular skin lesion with blanching under pressure.", "Capillary proliferation evident in this vascular lesion.", "Dome-shaped red papule characteristic of vascular skin anomaly."]\} \\ \hline
		\textbf{Link} & \url{https://www.kaggle.com/datasets/kmader/skin-cancer-mnist-ham10000} \\ \hline
	\end{tabular}
\end{table}

\begin{table}[h]
	\centering
	\caption{Statistics of Kvasir.}
	\begin{tabular}{|>{\raggedright\arraybackslash}p{3cm}|>{\raggedright\arraybackslash}p{11cm}|}
		\hline
		\multicolumn{2}{|c|}{\textbf{Kvasir \, (Endoscopy)}~\cite{Pogorelov:2017:KMI:3083187.3083212}} \\ \hline
		\textbf{Classes} & \{"dyed-lifted-polyps": 0, "dyed-resection-margins": 1, "esophagitis": 2, "normal-cecum": 3, "normal-pylorus": 4, "normal-z-line": 5, "polyps": 6, "ulcerative-colitis": 7\} \\ \hline
		\textbf{Splits} & \{"train": 4800, "test": 1600, "val": 1600\} \\ \hline
		\textbf{Description} & A gastrointestinal endoscopy dataset comprising 8,000 images from inside the GI tract, labeled for 8-class classification covering anatomical landmarks (z-line, pylorus, cecum) and pathological findings (esophagitis, polyps, ulcerative colitis, dyed procedures). \\ \hline
		\textbf{Prompts} & \{"dyed-lifted-polyps": [ "Endoscopic image showing dye-enhanced lifted polyps.", "Lifted lesion with dye contrast in colonic mucosa.", "Polyp elevated after submucosal injection captured in image.", "Image highlights dye-assisted mucosal elevation of polyp.", "Chromoscopic view enhancing lifted polyp margin." ], "dyed-resection-margins": [ "Dye highlights post-polypectomy margins.", "Endoscopic image shows dyed resection edges.", "Contrast-enhanced view of resection margins.", "Blue dye outlines treated area after lesion removal.", "Resection boundary stained for clear visualization." ], "esophagitis": [ "Inflammation and erosions indicating esophagitis.", "Mucosal breaks and redness consistent with esophagitis.", "Endoscopy showing linear esophageal erosions.", "Image reflects reflux-related esophageal inflammation.", "Signs of acid-induced mucosal damage in esophagus." ], "normal-cecum": [ "Normal cecum with visible appendiceal orifice.", "Image showing healthy cecal mucosa during colonoscopy.", "Unremarkable cecum with no pathological findings.", "Cecum image shows intact vascular pattern and haustral folds.", "Endoscopic image of normal-appearing cecum." ], "normal-pylorus": [ "Gastric pylorus visualized with no signs of obstruction.", "Normal pyloric channel evident in endoscopy.", "Healthy appearance of the pyloric region.", "Endoscopic view confirms normal pyloric function.", "Pylorus imaged without evidence of pathology." ], "normal-z-line": [ "Z-line shows sharp transition between epithelial types.", "Normal Z-line identified at esophagogastric junction.", "Endoscopic image of intact Z-line anatomy.", "Clear squamocolumnar transition seen in Z-line image.", "Z-line appears regular with no mucosal irregularities." ], "polyps": [ "Polyp seen as raised lesion on mucosal surface.", "Colonic polyp visualized during screening endoscopy.", "Endoscopic snapshot showing benign-appearing polyp.", "Pedunculated or sessile lesion consistent with polyp.", "Protruding mucosal abnormality detected in colon." ], "ulcerative-colitis": [ "Image of inflamed colon consistent with ulcerative colitis.", "Endoscopy shows mucosal edema and ulcerations.", "Colonic inflammation visible in ulcerative colitis patient.", "Loss of vascular pattern and bleeding in UC image.", "Chronic inflammation and granularity in UC lesion."]\} \\ \hline
		\textbf{Link} & \url{https://datasets.simula.no/kvasir/} \\ \hline
	\end{tabular}
\end{table}

\begin{table}[h]
	\centering
	\caption{Statistics of LC25000\_lung.}
	\begin{tabular}{|>{\raggedright\arraybackslash}p{3cm}|>{\raggedright\arraybackslash}p{11cm}|}
		\hline
		\multicolumn{2}{|c|}{\textbf{LC25000\_lung \, (Pathology)}~\cite{borkowski2019lung}} \\ \hline
		\textbf{Classes} & \{"lung\_aca": 0, "lung\_n": 1, "lung\_scc": 2\} \\ \hline
		\textbf{Splits} & \{"train": 9000, "test": 3000, "val": 3000\} \\ \hline
		\textbf{Description} & A histopathological image dataset comprising 15,000 lung tissue images at 768×768 pixels, labeled for 3-class classification to distinguish between lung adenocarcinoma, benign lung tissue, and lung squamous cell carcinoma. \\ \hline
		\textbf{Prompts} & \{"Lung adenocarcinoma": [ "Histopathology slide of Lung adenocarcinoma with glandular patterns", "Histopathology image showing Lung adenocarcinoma cells with mucin production", "Pathology tissue sample of Lung adenocarcinoma with lepidic growth pattern" ], "Benign lung": [ "Histopathology slide of Benign lung tissue with normal alveolar architecture", "Histopathology image showing Benign lung tissue with no abnormal cell growth", "Pathology tissue sample of Benign lung with fibrous tissue and no signs of malignancy" ], "Lung squamous cell carcinoma": [ "Histopathology slide of Lung squamous cell carcinoma with keratin pearls", "Histopathology image showing Lung squamous cell carcinoma cells with intercellular bridges", "Pathology tissue sample of Lung squamous cell carcinoma with squamous differentiation"]\} \\ \hline
		\textbf{Link} & \url{https://github.com/tampapath/lung_colon_image_set} \\ \hline
	\end{tabular}
\end{table}

\begin{table}[h]
	\centering
	\caption{Statistics of NCT\_CRC.}
	\begin{tabular}{|>{\raggedright\arraybackslash}p{3cm}|>{\raggedright\arraybackslash}p{11cm}|}
		\hline
		\multicolumn{2}{|c|}{\textbf{NCT\_CRC \, (Pathology)}~\cite{kather2018100}} \\ \hline
		\textbf{Classes} & \{"ADI": 0, "BACK": 1, "DEB": 2, "LYM": 3, "MUC": 4, "MUS": 5, "NORM": 6, "STR": 7, "TUM": 8\} \\ \hline
		\textbf{Splits} & \{"train": 90000, "test": 7180, "val": 10000\} \\ \hline
		\textbf{Description} & A histopathological image dataset comprising 107,180 H\&E-stained colorectal cancer tissue patches at 224×224 pixels, labeled for 9-class classification including normal tissue, tumor epithelium, stroma, and other tissue types. \\ \hline
		\textbf{Prompts} & \{"ADI": ["a histopathology slide showing Adipose", "histopathology image of Adipose", "pathology tissue showing Adipose", "presence of Adipose tissue on image"], "DEB": ["a histopathology slide showing Debris", "histopathology image of Debris", "pathology tissue showing Debris", "presence of Debris tissue on image"], "BACK": [ "a histopathology slide showing Background region", "histopathology image of Background region", "pathology tissue showing Background region", "presence of Background region on image"], "LYM": ["a histopathology slide showing Lymphocytes", "histopathology image of Lymphocytes", "pathology tissue showing Lymphocytes", "presence of Lymphocytes tissue on image"], "MUC": ["a histopathology slide showing Mucus", "histopathology image of Mucus", "pathology tissue showing Mucus", "presence of Mucus tissue on image"], "MUS": ["a histopathology slide showing Smooth muscle", "histopathology image of Smooth muscle", "pathology tissue showing Smooth muscle", "presence of Smooth muscle tissue on image"], "NORM": ["a histopathology slide showing Normal colon mucosa", "histopathology image of Normal colon mucosa", "pathology tissue showing Normal colon mucosa", "presence of Normal colon mucosa tissue on image"], "STR": ["a histopathology slide showing Cancer-associated stroma", "histopathology image of Cancer-associated stroma", "pathology tissue showing Cancer-associated stroma", "presence of Cancer-associated stroma tissue on image"], "TUM": [ "a histopathology slide showing Colorectal adenocarcinoma epithelium", "histopathology image of Colorectal adenocarcinoma epithelium", "pathology tissue showing Colorectal adenocarcinoma epithelium", "presence of Colorectal adenocarcinoma epithelium tissue on image"]\} \\ \hline
		\textbf{Link} & \url{https://zenodo.org/records/1214456} \\ \hline
	\end{tabular}
\end{table}

\begin{table}[h]
	\centering
	\caption{Statistics of Osteo.}
	\begin{tabular}{|>{\raggedright\arraybackslash}p{3cm}|>{\raggedright\arraybackslash}p{11cm}|}
		\hline
		\multicolumn{2}{|c|}{\textbf{Osteo \, (Pathology)}~\cite{arunachalam2019viable}} \\ \hline
		\textbf{Classes} & \{"Non-Tumor": 0, "Non-Viable-Tumor": 1, "Viable": 2\} \\ \hline
		\textbf{Splits} & \{"train": 653, "test": 221, "val": 217\} \\ \hline
		\textbf{Description} & A histopathological image dataset comprising 1,091 H\&E-stained osteosarcoma tissue tiles at 1024×1024 pixels, labeled for 3-class classification to distinguish between non-tumor tissue, necrotic tumor, and viable tumor regions. \\ \hline
		\textbf{Prompts} & \{"Normal non-tumor": ["a histopathology slide showing Normal non-tumor", "histopathology image of Normal non-tumor", "pathology tissue showing Normal non-tumor", "presence of Normal non-tumor tissue on image"], "Necrotic": ["a histopathology slide showing Necrotic", "histopathology image of Necrotic", "pathology tissue showing Necrotic", "presence of Necrotic tissue on image"], "Tumor": ["a histopathology slide showing Tumor", "histopathology image of Tumor", "pathology tissue showing Tumor", "presence of Tumor tissue on image"]\} \\ \hline
		\textbf{Link} & \url{https://journals.plos.org/plosone/article?id=10.1371/journal.pone.0210706#sec001} \\ \hline
	\end{tabular}
\end{table}

\begin{table}[h]
	\centering
	\caption{Statistics of WCE.}
	\begin{tabular}{|>{\raggedright\arraybackslash}p{3cm}|>{\raggedright\arraybackslash}p{11cm}|}
		\hline
		\multicolumn{2}{|c|}{\textbf{WCE \, (Endoscopy)}~\cite{montalbo2022colon}} \\ \hline
		\textbf{Classes} & \{"normal": 0, "ulcerative\_colitis": 1, "polyps": 2, "esophagitis": 3\} \\ \hline
		\textbf{Splits} & \{"train": 3200, "test": 800, "val": 2000\} \\ \hline
		\textbf{Description} & A curated wireless capsule endoscopy dataset comprising 6,000 gastrointestinal images labeled for 4-class classification to detect normal tissue, ulcerative colitis, polyps, and esophagitis conditions. \\ \hline
		\textbf{Prompts} & \{"normal": [ "Endoscopic view showing normal gastrointestinal mucosa.", "Unremarkable findings with smooth mucosal surface in endoscopy.", "No signs of inflammation or pathology in capsule endoscopy.", "Normal small bowel lining with intact villi visible.", "Capsule endoscopic image displaying normal peristalsis and mucosa." ], "ulcerative\_colitis": [ "Inflamed mucosa with ulceration consistent with ulcerative colitis.", "Endoscopic evidence of continuous mucosal inflammation in colon.", "Pseudopolyps and loss of vascular pattern seen in ulcerative colitis.", "Mucosal erythema and friability noted in colonic endoscopy.", "Capsule endoscopy indicating signs of active ulcerative colitis." ], "polyps": [ "Endoscopic view showing pedunculated colonic polyp.", "Capsule endoscopy revealing multiple small intestinal polyps.", "Raised mucosal lesion consistent with polyp formation.", "Colonic image indicating benign polypoid growth.", "Sessile polyps visualized in gastrointestinal endoscopy." ], "esophagitis": [ "Erosions and mucosal redness seen in esophageal inflammation.", "Endoscopic findings consistent with reflux esophagitis.", "Linear ulcerations in distal esophagus suggestive of esophagitis.", "Capsule imaging showing inflamed esophageal lining.", "Erythema and mucosal breaks typical of esophagitis."]\} \\ \hline
		\textbf{Link} & \url{https://www.kaggle.com/datasets/francismon/curated-colon-dataset-for-deep-learning} \\ \hline
	\end{tabular}
\end{table}

\begin{table}[h]
	\centering
	\caption{Statistics of PCAM.}
	\begin{tabular}{|>{\raggedright\arraybackslash}p{3cm}|>{\raggedright\arraybackslash}p{11cm}|}
		\hline
		\multicolumn{2}{|c|}{\textbf{PCAM \, (Pathology)}~\cite{kawai2023large}} \\ \hline
		\textbf{Classes} & \{"Lymph node": 0, "Lymph node containing metastatic tumor tissue": 1\} \\ \hline
		\textbf{Splits} & \{"train": 262144, "test": 32769, "val": 32769\} \\ \hline
		\textbf{Description} & A histopathology dataset comprising 327,680 color images (96×96 pixels) extracted from lymph node sections, labeled for binary classification to detect presence of metastatic breast cancer tissue. \\ \hline
		\textbf{Prompts} & \{"Lymph node": [ "Histopathology image of normal lymph node.", "Histopathology slide of a healthy lymph node." ], "Lymph node containing metastatic tumor tissue": [ "Histopathology image of a abnormal lymph node with metastatic tumor invasion and disrupted structure.", "Histopathology slide of a abnormal lymph node with metastatic tumor cells."]\} \\ \hline
		\textbf{Link} & \url{https://www.kaggle.com/datasets/andrewmvd/metastatic-tissue-classification-patchcamelyon} \\ \hline
	\end{tabular}
\end{table}

\begin{table}[h]
	\centering
	\caption{Statistics of PADUFES20.}
	\begin{tabular}{|>{\raggedright\arraybackslash}p{3cm}|>{\raggedright\arraybackslash}p{11cm}|}
		\hline
		\multicolumn{2}{|c|}{\textbf{PADUFES20 \, (Dermatology)}~\cite{pacheco2020pad}} \\ \hline
		\textbf{Classes} & \{"ACK": 0, "BCC": 1, "MEL": 2, "NEV": 3, "SCC": 4, "SEK": 5\} \\ \hline
		\textbf{Splits} & \{"train": 1378, "test": 462, "val": 458\} \\ \hline
		\textbf{Description} & A clinical skin lesion dataset comprising 2,298 smartphone-captured images from 1,373 patients, labeled for 6-class classification including three skin cancers (melanoma, basal cell carcinoma, squamous cell carcinoma) and three skin diseases (actinic keratosis, nevus, seborrheic keratosis), with up to 22 clinical features per sample. \\ \hline
		\textbf{Prompts} & \{"Actinic Keratosis": [ "A dermatological image of Actinic Keratosis.", "A diagnostic dermatological image illustrating Actinic Keratosis.", "Scaly lesion on sun-exposed skin suggestive of Actinic Keratosis.", "Actinic damage visible as rough, red or pink patches on the skin.", "Precancerous skin lesion indicating Actinic Keratosis in the image." ], "Basal Cell Carcinoma": [ "A dermatological image of Basal Cell Carcinoma.", "A diagnostic dermatological image illustrating Basal Cell Carcinoma.", "Nodular skin growth with pearly borders typical of Basal Cell Carcinoma.", "Ulcerated lesion in sun-exposed area suggesting BCC.", "Image consistent with slow-growing basal cell malignancy." ], "Melanoma": [ "A dermatological image of Melanoma.", "A diagnostic dermatological image illustrating Melanoma.", "Asymmetric pigmented lesion with irregular borders suggestive of Melanoma.", "Dermoscopic features indicating malignant Melanoma.", "Atypical mole with variegated pigmentation shown in image." ], "Nevus": [ "A dermatological image of Nevus.", "A diagnostic dermatological image illustrating Nevus.", "Benign pigmented lesion with symmetric and regular borders.", "A mole or nevus visible with uniform coloration.", "Common melanocytic nevus identified in skin image." ], "Squamous Cell Carcinoma": [ "A dermatological image of Squamous Cell Carcinoma.", "A diagnostic dermatological image illustrating Squamous Cell Carcinoma.", "Keratotic or ulcerated lesion typical of Squamous Cell Carcinoma.", "Invasive skin cancer with scaly and firm plaque appearance.", "Red, crusted lesion consistent with squamous cell origin." ], "Seborrheic Keratosis": [ "A dermatological image of Seborrheic Keratosis.", "A diagnostic dermatological image illustrating Seborrheic Keratosis.", "Waxy, stuck-on lesion indicative of Seborrheic Keratosis.", "Brown to black papule with sharply demarcated borders.", "Benign epidermal proliferation showing pseudohorn cysts."]\} \\ \hline
		\textbf{Link} & \url{https://smoggy-whimsey-274.notion.site/PAD-UFES-20-c1e0ff8f36814b94b205ee919e2e3ff8} \\ \hline
	\end{tabular}
\end{table}

\begin{table}[h]
	\centering
	\caption{Statistics of RetinalOCT.}
	\begin{tabular}{|>{\raggedright\arraybackslash}p{3cm}|>{\raggedright\arraybackslash}p{11cm}|}
		\hline
		\multicolumn{2}{|c|}{\textbf{RetinalOCT \, (OCT)}~\cite{naren2021oct}} \\ \hline
		\textbf{Classes} & \{"AMD": 0, "CNV": 1, "CSR": 2, "DME": 3, "DR": 4, "DRUSEN": 5, "MH": 6, "NORMAL": 7\} \\ \hline
		\textbf{Splits} & \{"train": 18400, "test": 2800, "val": 2800\} \\ \hline
		\textbf{Description} & A retinal OCT dataset comprising 24,000 high-quality optical coherence tomography images labeled for 8-class classification to detect age-related macular degeneration, choroidal neovascularization, central serous retinopathy, diabetic macular edema, diabetic retinopathy, drusen, macular hole, and normal retinal conditions. \\ \hline
		\textbf{Prompts} & \{"Age-related macular degeneration": [ "A retinal OCT image of Age-related Macular Degeneration.", "OCT scan showing drusen and retinal thinning characteristic of AMD.", "Disruption of retinal layers consistent with age-related changes.", "Retinal image displaying signs of dry or wet AMD.", "Photoreceptor layer degeneration observed in AMD case." ], "Choroidal neovascularization": [ "A retinal OCT image of Choroidal Neovascularization.", "Subretinal fluid accumulation indicating neovascular membranes.", "OCT showing CNV with RPE disruption and hyperreflective lesions.", "New abnormal vessels seen beneath the retina.", "Fluid and hemorrhage consistent with CNV pathology." ], "Central serous retinopathy": [ "A retinal OCT image of Central Serous Retinopathy.", "Serous detachment of neurosensory retina visible on OCT.", "Localized elevation of retinal layers due to fluid accumulation.", "OCT scan reveals subretinal fluid in macular region.", "CSR presentation with dome-shaped retinal detachment." ], "Diabetic macular edema": [ "A retinal OCT image of Diabetic Macular Edema.", "Cystoid spaces within the macula indicating fluid leakage.", "OCT shows thickening and intraretinal cysts in diabetic eye.", "Signs of chronic retinal swelling due to diabetes.", "Diffuse retinal edema in the macular area." ], "Diabetic retinopathy": [ "A retinal OCT image of Diabetic Retinopathy.", "Retinal thickening and microvascular changes observed in OCT.", "Signs of intraretinal hemorrhages and exudates in diabetic eye.", "OCT showing complications of long-standing diabetes.", "Hyperreflective lesions suggesting diabetic microangiopathy." ], "Drusen": [ "A retinal OCT image of Drusen.", "RPE elevation due to drusen deposits beneath retina.", "OCT scan shows scattered drusenoid material in macula.", "Small yellowish deposits characteristic of early AMD.", "Subretinal accumulation of lipoprotein debris visible." ], "Macular hole": [ "A retinal OCT image of Macular Hole.", "Full-thickness defect through the central retina.", "OCT scan shows round foveal discontinuity and retinal cysts.", "Loss of retinal tissue integrity in macular region.", "Retinal splitting or gap consistent with macular hole." ], "Normal": [ "A retinal OCT image of a normal healthy retina.", "Well-preserved retinal layers with no abnormalities.", "No signs of edema, atrophy, or fluid accumulation in OCT.", "Normal macular contour and foveal depression visible.", "OCT scan indicating intact retinal architecture."]\} \\ \hline
		\textbf{Link} & \url{https://www.kaggle.com/datasets/obulisainaren/retinal-oct-c8} \\ \hline
	\end{tabular}
\end{table}

\begin{table}[h]
	\centering
	\caption{Statistics of SIIMACR.}
	\begin{tabular}{|>{\raggedright\arraybackslash}p{3cm}|>{\raggedright\arraybackslash}p{11cm}|}
		\hline
		\multicolumn{2}{|c|}{\textbf{SIIMACR \, (X-Ray)}~\cite{zawacki2019siim}} \\ \hline
		\textbf{Classes} & \{"No\_Finding": 0, "Pneumothorax": 1\} \\ \hline
		\textbf{Splits} & \{"train": 3202, "test": 1069, "val": 1067\} \\ \hline
		\textbf{Description} & A chest X-ray dataset comprising 12,047 DICOM images labeled for binary classification and segmentation to detect pneumothorax (collapsed lung), with 2,669 positive cases containing pixel-level annotations and 9,378 negative cases. \\ \hline
		\textbf{Prompts} & \{"No Finding": [ "The chest radiograph reveals no abnormal findings.", "No pathological signs are evident in this chest X-ray image.", "Chest imaging presents a normal appearance without diagnostic concerns.", "The X-ray of the thorax shows no disease or abnormality." ], "Pneumothorax": [ "The chest radiograph displays signs of pneumothorax with lung collapse.", "X-ray image reveals air in the pleural space indicating pneumothorax.", "Chest X-ray suggests the presence of a pneumothorax.", "Radiographic evidence of unilateral lung collapse consistent with pneumothorax.", "Thoracic X-ray showing typical findings of pneumothorax."],\} \\ \hline
		\textbf{Link} & \url{https://www.kaggle.com/c/siim-acr-pneumothorax-segmentation} \\ \hline
	\end{tabular}
\end{table}

\begin{table}[h]
	\centering
	\caption{Statistics of Acl\_mri.}
	\begin{tabular}{|>{\raggedright\arraybackslash}p{3cm}|>{\raggedright\arraybackslash}p{11cm}|}
		\hline
		\multicolumn{2}{|c|}{\textbf{Acl\_mri \, (MRI)}~\cite{bien2018deep}} \\ \hline
		\textbf{Classes} & \{"no": 0, "yes": 1\} \\ \hline
		\textbf{Splits} & \{"train": 725, "test": 208, "val": 88\} \\ \hline
		\textbf{Description} & A knee MRI dataset comprising 1,021 MRI examinations for binary classification of ACL (anterior cruciate ligament) tears, derived from Stanford ML Group's MRNet dataset. The original MRNet dataset consists of 1,370 knee MRI exams performed at Stanford University Medical Center (2001-2012) with 319 (23.3\%) ACL tears confirmed through manual extraction from clinical reports. \\ \hline
		\textbf{Prompts} & \{"no": ["Normal appearance of the knee in MRI imaging.", "MRI scan showing no abnormalities in the knee.", "The knee MRI indicates a normal condition.", "MRI of the knee displaying typical anatomy and no pathologic findings.", "Normal knee structure as revealed by magnetic resonance imaging.", "Magnetic resonance image of the knee showing healthy structural integrity.", "The MRI results confirm normal knee anatomy without any abnormal findings.", "Diagnostic MRI reveals the knee to be in normal condition.", "Knee MRI, which shows no signs of pathology, indicating a normal state.", "MRI examination of the knee presents normal anatomical features.."], "yes": ["ACL pathology is evident in the MRI imaging of the knee.", "The MRI scan reveals a pathology in the anterior cruciate ligament of the knee.", "Pathological changes in the knee ACL are detected in the MRI.", "The knee MRI results indicate an ACL disorder.", "An anterior cruciate ligament pathology is diagnosed through the knee MRI.", "MRI of the knee shows compromised integrity of the ACL.", "The knee anterior cruciate ligament exhibits pathological abnormalities in the MRI scan.", "MRI findings suggest a disease process in the knee anterior cruciate ligament.", "The imaging of the knee reveals a pathological condition of the ACL.", "A disorder in the knee’s anterior cruciate ligament is visible on MRI."]\} \\ \hline
		\textbf{Link} & \url{https://github.com/ahmedbesbes/mrnet} \\ \hline
	\end{tabular}
\end{table}

\begin{table}[h]
	\centering
	\caption{Statistics of Breast\_us.}
	\begin{tabular}{|>{\raggedright\arraybackslash}p{3cm}|>{\raggedright\arraybackslash}p{11cm}|}
		\hline
		\multicolumn{2}{|c|}{\textbf{Breast\_us \, (Ultrasound)}~\cite{al2020dataset}} \\ \hline
		\textbf{Classes} & \{"benign": 0, "malignant": 1\} \\ \hline
		\textbf{Splits} & \{"train": 561, "test": 156, "val": 63\} \\ \hline
		\textbf{Description} & A breast ultrasound dataset comprising 780 images from 600 female patients, labeled for binary classification between benign and malignant breast lesions, derived from the Al-Dhabyani et al. breast ultrasound images dataset. \\ \hline
		\textbf{Prompts} & \{"benign": ["Benign lesions are visible in the ultrasound imaging of the breast.", "The ultrasound scan shows a benign pathology in the breast tissue.", "Breast ultrasound reveals features consistent with benign conditions.", "Benign findings indicated in the breast by ultrasonic imaging.", "Ultrasonography of the breast suggests a benign anomaly.", "The breast sonographic examination displays benign characteristics.", "Evidence of benign abnormalities observed in ultrasonic breast images.", "Breast imaging via ultrasound depicting benign results.", "Non-malignant indications identified in breast ultrasound.", "Ultrasound of the breast demonstrates non-cancerous features."], "malignant": ["Lesion (migraine) is identified in the ultrasound imaging of the breast.", "In the ultrasound image, a malignant lesion is evident in the breast tissue.", "Breast ultrasound reveals presence of a lesion characterized as malignant.", "The ultrasound study of the breast shows a malignant lesion.", "Malignant lesion appears in the breast as seen in ultrasound image.", "Breast ultrasonographic image displays a malignant lesion.", "Ultrasound diagnostic of the breast indicates a malignant lesion.", "Malignant lesion in breast detected by ultrasound imaging.", "Breast ultrasound imaging reveals a malignancy in the lesion.", "A malignant lesion is showcased in the ultrasound image of the breast."]\} \\ \hline
		\textbf{Link} & \url{https://pubmed.ncbi.nlm.nih.gov/31867417/} \\ \hline
	\end{tabular}
\end{table}

\begin{table}[h]
	\centering
	\caption{Statistics of Breastmnist.}
	\begin{tabular}{|>{\raggedright\arraybackslash}p{3cm}|>{\raggedright\arraybackslash}p{11cm}|}
		\hline
		\multicolumn{2}{|c|}{\textbf{Breastmnist \, (Ultrasound)}~\cite{yang2023medmnist}} \\ \hline
		\textbf{Classes} & \{"malignant": 0, "normal, benign": 1\} \\ \hline
		\textbf{Splits} & \{"train": 546, "test": 156, "val": 78\} \\ \hline
		\textbf{Description} & A breast ultrasound dataset comprising 780 images pre-processed to 28×28 pixels from the MedMNIST collection, labeled for binary classification to distinguish between malignant and normal/benign breast conditions. \\ \hline
		\textbf{Prompts} & \{"malignant": [ "Breast ultrasound or mammogram showing malignant lesion.", "Irregular mass with spiculated margins and hypoechoic center.", "Malignant features including asymmetry, distortion, or microcalcifications.", "Radiologic appearance consistent with breast carcinoma.", "Image indicates presence of aggressive or invasive breast tumor." ], "normal, benign": [ "Image showing normal breast tissue or benign lesion.", "Well-circumscribed mass with benign features on imaging.", "Cystic or fibroadenomatous structure observed.", "No suspicious features indicating malignancy.", "Breast image consistent with benign or normal classification."]\} \\ \hline
		\textbf{Link} & \url{https://medmnist.com/} \\ \hline
	\end{tabular}
\end{table}

\begin{table}[h]
	\centering
	\caption{Statistics of Five\_retina.}
	\begin{tabular}{|>{\raggedright\arraybackslash}p{3cm}|>{\raggedright\arraybackslash}p{11cm}|}
		\hline
		\multicolumn{2}{|c|}{\textbf{Five\_retina \, (Fundus)}~\cite{jin2022fives}} \\ \hline
		\textbf{Classes} & \{"A": 0, "D": 1, "G": 2, "N": 3\} \\ \hline
		\textbf{Splits} & \{"train": 600, "test": 200, "val": 0\} \\ \hline
		\textbf{Description} & A fundus image dataset comprising 800 retinal photographs labeled for 4-class classification to detect age-related macular degeneration, diabetic retinopathy, glaucoma, and normal conditions, derived from the FIVES dataset. \\ \hline
		\textbf{Prompts} & \{"A": [ "Age-related macular degeneration: This fundus image shows drusen deposits.", "The image reveals geographic atrophy, a sign of age-related macular degeneration.", "Fundus image displays pigmentary changes associated with age-related macular degeneration.", "Age-related macular degeneration: This eye fundus image shows subretinal fluid.", "The macula appears distorted, indicative of age-related macular degeneration." ], "D": [ "Diabetic retinopathy: This fundus image shows microaneurysms.", "The image displays retinal hemorrhages, a sign of diabetic retinopathy.", "Diabetic retinopathy: Fundus image reveals cotton wool spots.", "This eye fundus image shows neovascularization, characteristic of diabetic retinopathy.", "The retina appears with hard exudates, indicating diabetic retinopathy." ], "G": [ "Glaucoma: This fundus image shows optic nerve damage.", "The image displays increased cup-to-disc ratio, a sign of glaucoma.", "Glaucoma: The fundus image reveals a pale optic disc.", "This eye fundus image shows visual field loss, typical of glaucoma.", "The optic nerve appears excavated, indicating glaucoma." ], "N": [ "Normal: This fundus image shows a healthy optic nerve with no signs of disease.", "The optic disc and cup appear normal, indicating no abnormalities.", "Normal: The fundus image reveals no retinal damage, suggesting a healthy eye.", "This eye fundus image is normal, with no evidence of any disease.", "The image shows a normal macula and retina, indicating a healthy eye."]\} \\ \hline
		\textbf{Link} & \url{https://www.kaggle.com/datasets/mohamedbounechada/fives-a-fundus-image} \\ \hline
	\end{tabular}
\end{table}

\begin{table}[h]
	\centering
	\caption{Statistics of Rsna\_pneumonia.}
	\begin{tabular}{|>{\raggedright\arraybackslash}p{3cm}|>{\raggedright\arraybackslash}p{11cm}|}
		\hline
		\multicolumn{2}{|c|}{\textbf{Rsna\_pneumonia \, (X-Ray)}~\cite{shih2019augmenting}} \\ \hline
		\textbf{Classes} & \{"Normal": 0, "Pneumonia": 1\} \\ \hline
		\textbf{Splits} & \{"train": 18136, "test": 6046, "val": 6045\} \\ \hline
		\textbf{Description} & A chest X-ray dataset comprising 30,227 frontal-view radiographs labeled for binary classification to detect pneumonia, derived from the RSNA Pneumonia Detection Challenge 2018 based on the NIH chest X-ray database. \\ \hline
		\textbf{Prompts} & \{"Normal": ["No pathological findings are evident in the chest X-ray image.", "The chest radiograph shows no abnormalities.", "There are no detectable abnormalities in the X-ray of the chest.", "The X-ray examination of the chest reveals no significant findings.", "No diagnostic abnormalities are present in the chest X-ray.", "The radiographic chest image indicates an absence of pathological conditions.", "No findings of concern appear in the chest imaging from the X-ray.", "The chest X-ray is clear of any detectable conditions.", "There are no visible pathological signs in the X-ray of the chest.", "The chest X-ray image is free from any abnormalities."] , "Pneumonia": ["Pneumonia is indicated in this chest X-ray image.", "This radiograph of the chest shows characteristics typical of pneumonia.", "Signs of pneumonia are evident in the X-ray of the chest.", "The chest radiograph reveals pulmonary patterns consistent with pneumonia.", "Evidence of pneumonia is present in this thoracic X-ray.", "This X-ray of the chest displays diagnostic features of pneumonia.", "Manifestations of pneumonia appear in this chest X-ray.", "Pneumonia is diagnosed in the displayed chest X-ray.", "The chest X-ray examination suggests the presence of pneumonia.", "Chest X-ray analysis indicates pneumonia in this image."]\} \\ \hline
		\textbf{Link} & \url{https://www.rsna.org/rsnai/ai-image-challenge/rsna-pneumonia-detection-challenge-2018} \\ \hline
	\end{tabular}
\end{table}

\begin{table}[h]
	\centering
	\caption{Statistics of Kather\_et\_al\_2016.}
	\begin{tabular}{|>{\raggedright\arraybackslash}p{3cm}|>{\raggedright\arraybackslash}p{11cm}|}
		\hline
		\multicolumn{2}{|c|}{\textbf{Kather\_et\_al\_2016 \, (Pathology)}~\cite{kather2016multi}} \\ \hline
		\textbf{Classes} & \{"adipose": 0, "complex": 1, "debris": 2, "empty": 3, "lympho": 4, "mucosa": 5, "stroma": 6, "tumor": 7\} \\ \hline
		\textbf{Splits} & \{"train": 600, "test": 200, "val": 200\} \\ \hline
		\textbf{Description} & A colorectal cancer histology dataset comprising 5,000 H\&E-stained tissue images (150×150 pixels) labeled for 8-class classification of tissue types including tumor, stroma, adipose, lymphocytes, mucosa, debris, complex regions, and background. \\ \hline
		\textbf{Prompts} & \{"adipose": [ "Histological image of adipose tissue with large, clear fat vacuoles.", "Loose connective tissue with round adipocytes and empty cytoplasm.", "Adipose region characterized by lipid-filled cells with peripheral nuclei.", "Low cellularity area consistent with fatty tissue morphology." ], "complex": [ "Complex tissue architecture with mixed components visible.", "Histology showing overlapping regions of tumor, stroma, and mucosa.", "Tissue with heterogeneous morphology and ambiguous classification.", "Complex structure with poorly defined boundaries among tissue types." ], "debris": [ "Cellular debris and necrotic material observed in tissue section.", "Histopathological image with fragmented and non-viable cell remnants.", "Non-specific granular content consistent with necrotic debris.", "Area filled with breakdown products of dead cells." ], "empty": [ "Blank or artifact region with no discernible tissue structure.", "Empty region of the histology slide, possibly out of focus.", "Image field with no relevant biological tissue.", "Uninformative area lacking cellular or matrix components." ], "lympho": [ "Dense aggregation of small round lymphocytes visible in tissue.", "Lymphoid infiltrate forming a focal or diffuse pattern.", "Image showing sheets of immune cells consistent with lymphoid tissue.", "Compact clusters of small, dark nuclei typical of lymphocytes." ], "mucosa": [ "Intact mucosal lining with regular epithelial cell architecture.", "Histological section showing villous or glandular mucosa.", "Mucosa with columnar epithelial cells and underlying lamina propria.", "Well-organized mucosal layer from intestinal lining visible." ], "stroma": [ "Fibrous stroma composed of spindle-shaped cells and collagen.", "Supportive connective tissue surrounding tumor glands or normal mucosa.", "Stromal component with moderate cellularity and extracellular matrix.", "Histology showing stromal tissue separating epithelial components." ], "tumor": [ "Malignant epithelial cells forming irregular glandular structures.", "Tumor region with increased nuclear atypia and mitotic activity.", "Histological evidence of invasive carcinoma.", "Neoplastic cells disrupting normal tissue architecture."]\} \\ \hline
		\textbf{Link} & \url{https://pubmed.ncbi.nlm.nih.gov/27306927/} \\ \hline
	\end{tabular}
\end{table}

\begin{table}[h]
	\centering
	\caption{Statistics of Kather\_et\_al\_2018.}
	\begin{tabular}{|>{\raggedright\arraybackslash}p{3cm}|>{\raggedright\arraybackslash}p{11cm}|}
		\hline
		\multicolumn{2}{|c|}{\textbf{Kather\_et\_all\_2018 \, (Pathology)}~\cite{kather2019predicting}} \\ \hline
		\textbf{Classes} & \{"adipose": 0, "background": 1, "cancer\_stroma": 2, "debris": 3, "lymphocytes": 4, "mucus": 5, "normal\_mucosa": 6, "smooth\_muscle": 7, "tumor": 8\} \\ \hline
		\textbf{Splits} & \{"train": 1080, "test": 360, "val": 360\} \\ \hline
		\textbf{Description} & A colorectal cancer histology dataset comprising 1,800 H\&E-stained tissue images (224×224 pixels) labeled for 9-class classification, derived from the NCT-CRC-HE-100K dataset used for survival prediction in colorectal cancer. \\ \hline
		\textbf{Prompts} & \{"adipose": [ "Adipose tissue characterized by clear cytoplasm and round vacuoles.", "Large, pale cells with peripheral nuclei indicating fat cells.", "Lipid-rich adipocytes forming low-density tissue matrix.", "Histology image consistent with subcutaneous or visceral fat." ], "background": [ "Slide background or artifact region without biological tissue.", "Non-tissue area with uniform color and no histological content.", "Image containing scanner artifacts or background noise.", "Region outside annotated tissue areas on histology slide." ], "cancer\_stroma": [ "Desmoplastic stroma associated with invasive carcinoma.", "Fibroblastic reaction surrounding malignant epithelial cells.", "Dense extracellular matrix and spindle cells indicating tumor stroma.", "Stromal remodeling and increased collagen in cancer-associated areas." ], "debris": [ "Histology image containing necrotic cell fragments and granular debris.", "Tissue degradation byproducts lacking viable nuclei.", "Necrosis-associated debris visible in tumor or inflammatory zones.", "Non-viable cell material scattered throughout the section." ], "lymphocytes": [ "Image showing dense lymphocytic infiltration in stromal regions.", "Small round cells with dark nuclei consistent with lymphocytes.", "Chronic inflammatory infiltrate composed of lymphoid cells.", "Histological evidence of immune cell accumulation." ], "mucus": [ "Mucous lakes or extracellular mucin evident in tissue.", "Histology image showing pools of pale mucin within tumor.", "Mucinous substance surrounding or separating epithelial cells.", "Mucin-producing cells releasing extracellular material." ], "normal\_mucosa": [ "Normal mucosal epithelium with organized architecture.", "Intact glandular structures and uniform epithelial cells.", "Healthy mucosal lining without atypia or inflammation.", "Columnar epithelium forming regular crypts and surface lining." ], "smooth\_muscle": [ "Smooth muscle bundles with elongated nuclei and eosinophilic cytoplasm.", "Spindle-shaped muscle fibers arranged in parallel arrays.", "Muscularis propria layer visible in gastrointestinal wall.", "Histological features consistent with visceral smooth muscle." ], "tumor": [ "Malignant glandular structures invading surrounding stroma.", "Neoplastic epithelial cells with nuclear pleomorphism and mitoses.", "Tumor area showing disrupted tissue organization.", "Atypical cells forming irregular nests and sheets in tumor region."]\} \\ \hline
		\textbf{Link} & \url{https://journals.plos.org/plosmedicine/article?id=10.1371/journal.pmed.1002730} \\ \hline
	\end{tabular}
\end{table}

\begin{table}[h]
	\centering
	\caption{Statistics of kather\_et\_al\_2018\_val7k.}
	\begin{tabular}{|>{\raggedright\arraybackslash}p{3cm}|>{\raggedright\arraybackslash}p{11cm}|}
		\hline
		\multicolumn{2}{|c|}{\textbf{kather\_et\_al\_2018\_val7k \, (Pathology)}~\cite{kather2019predicting}} \\ \hline
		\textbf{Classes} & \{"adipose": 0, "background": 1, "cancer\_stroma": 2, "debris": 3, "lymphocytes": 4, "mucus": 5, "normal\_mucosa": 6, "smooth\_muscle": 7, "tumor": 8\}  \\ \hline
		\textbf{Splits} & \{"train": 796, "test": 259, "val": 259\} \\ \hline
		\textbf{Description} & A colorectal cancer histology validation dataset comprising 1,314 H\&E-stained tissue images (224×224 pixels) labeled for 9-class classification, subset of the CRC-VAL-HE-7K external validation set from Kather et al. 2018. \\ \hline
		\textbf{Prompts} & \{"adipose": [ "Adipose tissue composed of round, clear vacuolated fat cells.", "Fatty tissue region with low cellularity and large cytoplasmic droplets.", "Histological image of normal adipose structure.", "Pale-stained adipocytes with peripheral flattened nuclei." ], "background": [ "Non-tissue background region on histology slide.", "Artifact zone with no biological relevance.", "Uniform area outside annotated tissue section.", "Scanner-related or preparation-related background field." ], "cancer\_stroma": [ "Fibrotic stroma adjacent to invasive tumor glands.", "Reactive stromal changes in response to cancer invasion.", "Dense connective tissue seen in tumor microenvironment.", "Image showing desmoplastic stromal remodeling in cancer." ], "debris": [ "Histological area filled with fragmented, necrotic material.", "Cellular debris lacking structural organization.", "Breakdown products and inflammatory residue in tissue.", "Zone of necrosis or degeneration visible in slide." ], "lymphocytes": [ "Clusters of small, round, dark-stained immune cells.", "Lymphoid cell aggregates infiltrating tissue.", "Dense lymphocytic presence in inflammatory or tumor context.", "Histology image of lymphocyte-rich region." ], "mucus": [ "Pale, extracellular mucin pools indicating mucus secretion.", "Mucous material accumulating between glands.", "Mucinous phenotype evident in epithelial tissue.", "Histological evidence of mucus-producing cells." ], "normal\_mucosa": [ "Healthy mucosal layer with orderly epithelial arrangement.", "Normal glandular tissue showing intact morphology.", "Regular mucosal pattern with crypts and basal nuclei.", "Non-dysplastic mucosa with no inflammatory or neoplastic changes." ], "smooth\_muscle": [ "Smooth muscle tissue with parallel, eosinophilic fibers.", "Image shows characteristic elongated nuclei in muscle layer.", "Spindle-shaped cells forming contractile tissue.", "Muscularis propria seen in gastrointestinal section." ], "tumor": [ "Malignant epithelial structures with glandular distortion.", "Tumor cells infiltrating beyond normal boundaries.", "High-grade neoplastic area with cellular atypia.", "Histopathology showing irregular tumor cell clusters."]\} \\ \hline
		\textbf{Link} & \url{https://journals.plos.org/plosmedicine/article?id=10.1371/journal.pmed.1002730} \\ \hline
	\end{tabular}
\end{table}

\begin{table}[h]
	\centering
	\caption{Statistics of Organamnist.}
	\begin{tabular}{|>{\raggedright\arraybackslash}p{3cm}|>{\raggedright\arraybackslash}p{11cm}|}
		\hline
		\multicolumn{2}{|c|}{\textbf{Organamnist \, (CT)}~\cite{yang2023medmnist}} \\ \hline
		\textbf{Classes} & \{"bladder": 0, "femur-left": 1, "femur-right": 2, "heart": 3, "kidney-left": 4, "kidney-right": 5, "liver": 6, "lung-left": 7, "lung-right": 8, "pancreas": 9, "spleen": 10\}  \\ \hline
		\textbf{Splits} & \{"train": 34561, "test": 17778, "val": 6491\} \\ \hline
		\textbf{Description} & An abdominal CT dataset comprising 58,830 images pre-processed to 28×28 pixels from the MedMNIST collection, labeled for 11-class classification to identify different organs including bladder, femurs, heart, kidneys, liver, lungs, pancreas, and spleen. \\ \hline
		\textbf{Prompts} & \{"bladder": [ "Axial scan showing the anatomical structure of the bladder.", "Cross-sectional image highlighting the urinary bladder.", "Bladder located in the pelvic cavity with visible lumen.", "Medical image revealing the outline of the bladder organ.", "Well-defined bladder structure seen in imaging slice." ], "femur-left": [ "Image showing the left femur bone in a coronal or axial view.", "Anatomical scan of the left thigh region revealing the femur.", "Clear visualization of the left femoral shaft and head.", "Left femur bone structure visible in radiographic image.", "Image displays long bone morphology of the left femur." ], "femur-right": [ "Scan highlighting the right femur in skeletal imaging.", "Right femoral anatomy shown in cross-sectional view.", "Radiographic image capturing right femur shaft and head.", "Right femur bone visible with cortical and trabecular detail.", "Right thigh bone visualized in diagnostic image." ], "heart": [ "Cross-sectional scan revealing the chambers of the heart.", "Image showing cardiac silhouette in thoracic cavity.", "Heart anatomy including ventricles and atria visible.", "Axial image displaying the heart and surrounding vessels.", "Cardiac structure visualized in thoracic imaging." ], "kidney-left": [ "Left kidney visualized with clear renal cortex and medulla.", "Cross-section showing left kidney"s internal anatomy.", "Renal parenchyma and collecting system in the left kidney.", "Medical image showing the left kidney in abdominal cavity.", "Left-sided renal structure observed in scan." ], "kidney-right": [ "Right kidney outlined with distinct internal structures.", "Imaging slice highlighting the right kidney in situ.", "Right renal anatomy with visible cortex and calyces.", "Image showing normal positioning of the right kidney.", "Cross-section of right kidney with identifiable landmarks." ], "liver": [ "Liver appearing in the upper right abdominal quadrant.", "Anatomical image showing liver lobes and surface.", "Large organ in right hypochondrium identified as liver.", "Cross-section of liver with vascular supply visible.", "Liver seen adjacent to diaphragm and stomach." ], "lung-left": [ "Left lung shown in thoracic imaging with visible lobes.", "Pulmonary fields of the left lung appearing aerated.", "Left lung anatomy seen near cardiac border.", "Image showing bronchial and alveolar structures of the left lung.", "Lung field visualized on the left side of thoracic cavity." ], "lung-right": [ "Right lung visualized with three lobes in imaging slice.", "Thoracic scan showing right pulmonary anatomy.", "Right lung field with visible vascular markings.", "Image highlights air-filled right lung structure.", "Radiologic scan showing right-sided pulmonary region." ], "pancreas": [ "Cross-sectional image showing pancreas in epigastric region.", "Pancreas visualized as elongated soft tissue structure.", "Axial scan highlighting pancreatic head and tail.", "Pancreatic tissue visible posterior to the stomach.", "Organ with endocrine and exocrine function imaged here." ], "spleen": [ "Spleen seen in left upper quadrant adjacent to stomach.", "Homogeneous soft tissue mass representing spleen.", "Splenic parenchyma visualized in abdominal imaging.", "Image showing spleen with smooth borders and hilum.", "Spleen observed lateral to the left kidney in scan."]\} \\ \hline
		\textbf{Link} & \url{https://medmnist.com/} \\ \hline
	\end{tabular}
\end{table}

\begin{table}[h]
	\centering
	\caption{Statistics of Skin\_cancer.}
	\begin{tabular}{|>{\raggedright\arraybackslash}p{3cm}|>{\raggedright\arraybackslash}p{11cm}|}
		\hline
		\multicolumn{2}{|c|}{\textbf{Skin\_cancer \, (Pathology)}~\cite{kriegsmann2022deep}} \\ \hline
		\textbf{Classes} & \{"nontumor\_skin\_necrosis\_necrosis": 0, "nontumor\_skin\_muscle\_skeletal": 1, "nontumor\_skin\_sweatglands\_sweatglands": 2, "nontumor\_skin\_vessel\_vessel": 3, "nontumor\_skin\_elastosis\_elastosis": 4, "nontumor\_skin\_chondraltissue\_chondraltissue": 5, "nontumor\_skin\_hairfollicle\_hairfollicle": 6, "nontumor\_skin\_epidermis\_epidermis": 7, "nontumor\_skin\_nerves\_nerves": 8, "nontumor\_skin\_subcutis\_subcutis": 9, "nontumor\_skin\_dermis\_dermis": 10, "nontumor\_skin\_sebaceousglands\_sebaceousglands": 11, "tumor\_skin\_epithelial\_sqcc": 12, "tumor\_skin\_melanoma\_melanoma": 13, "tumor\_skin\_epithelial\_bcc": 14, "tumor\_skin\_naevus\_naevus": 15\} \\ \hline
		\textbf{Splits} & \{"train": 3815, "test": 1211, "val": 1081\} \\ \hline
		\textbf{Description} & A skin histopathology dataset comprising 6,107 image tiles (100×100 $\mu$m, 395×395 pixels) labeled for 16-class classification of skin tissue structures and neoplasms, including 12 non-tumor and 4 tumor categories (BCC, SqCC, melanoma, naevus). \\ \hline
		\textbf{Prompts} & \{"nontumor\_skin\_necrosis\_necrosis": ["a histopathology slide showing Necrosis", ...], "nontumor\_skin\_muscle\_skeletal": ["a histopathology slide showing Skeletal muscle", ...], "nontumor\_skin\_sweatglands\_sweatglands": ["a histopathology slide showing Eccrine sweat glands", ...], "nontumor\_skin\_vessel\_vessel": ["a histopathology slide showing Vessels", ...], "nontumor\_skin\_elastosis\_elastosis": ["a histopathology slide showing Elastosis", ...], "nontumor\_skin\_chondraltissue\_chondraltissue": ["a histopathology slide showing Chondral tissue", ...], "nontumor\_skin\_hairfollicle\_hairfollicle": ["a histopathology slide showing Hair follicle", ...], "nontumor\_skin\_epidermis\_epidermis": ["a histopathology slide showing Epidermis", ...], "nontumor\_skin\_nerves\_nerves": ["a histopathology slide showing Nerves", ...], "nontumor\_skin\_subcutis\_subcutis": ["a histopathology slide showing Subcutis", ...], "nontumor\_skin\_dermis\_dermis": ["a histopathology slide showing Dermis", ...], "nontumor\_skin\_sebaceousglands\_sebaceousglands": ["a histopathology slide showing Sebaceous glands", ...], "tumor\_skin\_epithelial\_sqcc": ["a histopathology slide showing Squamous-cell carcinoma", ...], "tumor\_skin\_melanoma\_melanoma": ["a histopathology slide showing Melanoma in-situ", ...], "tumor\_skin\_epithelial\_bcc": ["a histopathology slide showing Basal-cell carcinoma", ...], "tumor\_skin\_naevus\_naevus": ["a histopathology slide showing Naevus", ...]\} \\ \hline
		\textbf{Link} & \url{https://heidata.uni-heidelberg.de/dataset.xhtml?persistentId=doi:10.11588/data/7QCR8S} \\ \hline
	\end{tabular}
\end{table}

\begin{table}[h]
	\centering
	\caption{Statistics of Tang\_et\_al\_2019.}
	\begin{tabular}{|>{\raggedright\arraybackslash}p{3cm}|>{\raggedright\arraybackslash}p{11cm}|}
		\hline
		\multicolumn{2}{|c|}{\textbf{Tang\_et\_al\_2019 \, (Pathology)}~\cite{tang2019interpretable}} \\ \hline
		\textbf{Classes} & \{"caa": 0, "cored": 1, "diffuse": 2, "negative": 3\} \\ \hline
		\textbf{Splits} & \{"train": 295, "test": 98, "val": 98\} \\ \hline
		\textbf{Description} & An Alzheimer's disease histopathology dataset comprising 491 brain tissue images (256×256 pixels) labeled for 4-class classification of amyloid pathologies: cerebral amyloid angiopathy (CAA), cored plaques, diffuse plaques, and negative controls. \\ \hline
		\textbf{Prompts} & \{"caa": ["Amyloid deposits observed along cerebral blood vessels, consistent with CAA.", "Cerebral amyloid angiopathy identified by vascular A$\beta$ accumulation.", "Perivascular amyloid staining visible in histological image.", "Image reveals amyloid-laden vessel walls typical of CAA.", "Fluorescence pattern indicating cerebral vascular amyloid."], "cored": ["Cored amyloid plaque with dense central region observed.", "Image shows amyloid core surrounded by neuritic halo.", "Compact amyloid deposit with distinct core visible.", "Dense-core plaque characteristic of advanced amyloid pathology.", "Amyloid staining reveals central core structure in tissue."], "diffuse": ["Diffuse amyloid plaque lacking a compact core.", "Widespread amyloid deposition with amorphous appearance.", "Image shows loosely distributed amyloid without dense center.", "Non-cored, spread-out amyloid staining observed.", "Diffuse A$\beta$ aggregates visible in parenchymal tissue." ], "negative": [ "No amyloid deposits detected in image.", "Negative control or healthy tissue with no A$\beta$ accumulation.", "Image shows absence of pathological protein aggregates.", "No staining patterns indicative of amyloid pathology.", "Histological section free of amyloid burden."]\} \\ \hline
		\textbf{Link} & \url{https://zenodo.org/records/1470797} \\ \hline
	\end{tabular}
\end{table}

\begin{table}[h]
	\centering
	
	\caption{Statistics of Wong\_et\_al\_2022.}
	\begin{tabular}{|>{\raggedright\arraybackslash}p{3cm}|>{\raggedright\arraybackslash}p{11cm}|}
		\hline
		\multicolumn{2}{|c|}{\textbf{Wong\_et\_al\_2022 \, (Pathology)}~\cite{wong2022deep}} \\ \hline
		\textbf{Classes} & \{"caa": 0, "cored": 1, "diffuse": 2, "negative": 3\} \\ \hline
		\textbf{Splits} & \{"train": 480, "test": 160, "val": 160\} \\ \hline
		\textbf{Description} & An Alzheimer's disease histopathology dataset comprising 800 brain tissue images (256×256 pixels) labeled for 4-class classification of amyloid pathologies: CAA, cored plaques, diffuse plaques, and negative controls, derived from multi-expert annotations. \\ \hline
		\textbf{Prompts} & \{"caa": [ "Cerebral blood vessels showing amyloid deposition consistent with CAA.", "Histopathology image of cerebral amyloid angiopathy with vascular A$\beta$.", "Perivascular amyloid plaques observed, indicating CAA.", "Fluorescence highlights amyloid along vessel walls typical of CAA.", "Image shows blood vessel-associated amyloid staining." ], "cored": [ "Dense-cored amyloid plaque with a compact central region.", "Image shows characteristic cored plaque with bright center and surrounding halo.", "Dense amyloid accumulation forming classic cored structure.", "Cored plaque evident with central bright core under staining.", "Histological section reveals cored amyloid deposition." ], "diffuse": [ "Diffuse amyloid plaque without a defined core.", "Amorphous and widespread amyloid deposits visible.", "Image shows non-compact, scattered amyloid staining.", "Diffuse A$\beta$ aggregation lacking central densification.", "Spread-out amyloid pathology seen in brain parenchyma." ], "negative": [ "No amyloid deposits visible in this histological section.", "Negative for amyloid plaque staining.", "Image shows no evidence of cerebral amyloid pathology.", "Brain tissue appears free of amyloid accumulation.", "Normal background staining without pathological features."]\} \\ \hline
		\textbf{Link} & \url{https://osf.io/xh2jd/} \\ \hline
	\end{tabular}
	\label{zeroend}
\end{table}

\clearpage
\newpage
\subsubsection{Cross-modal retrieval}

\begin{table}[!h]
	\centering
	\caption{Statistics of cross-modal retrieval dataset.}
	\begin{tabular}{|>{\raggedright\arraybackslash}p{2cm}|>{\raggedright\arraybackslash}p{11cm}|}
		\hline
		\multicolumn{2}{|c|}{\textbf{Bookset}~\cite{gamper2021multiple}} \\ \hline
		\textbf{Splits} & \{"train": -, "test": 4,270, "val": -\} \\ \hline
		\textbf{Description} & The BookSet collection consists of 4,265 paired images and textual descriptions, sourced from figures and corresponding captions within ten academic textbooks. Our research employed this BookSet collection to assess cross-modal retrieval capabilities.  \\
		\hline
		\textbf{Link} & \url{https://warwick.ac.uk/fac/cross_fac/tia/data/arch} \\ \hline
		\multicolumn{2}{|c|}{\textbf{MedTrinity-25M subset}~\cite{xie2024medtrinity}} \\ \hline
		\textbf{Splits} & \{"train": -, "test": 3,771, "val": -\} \\ \hline
		\textbf{Description} & MedTrinity-25M is a large-scale medical multimodal dataset containing over 25 million images across 10 modalities with comprehensive annotations for 65+ diseases. The dataset provides both global textual descriptions and detailed local annotations, enabling large-scale pre-training of medical AI foundation models and supporting diverse tasks including captioning, classification, and segmentation. However, it contains a large number of similar images. We only randomly select 3,771 images as a subset to evaluate MMKD-CLIP. \\
		\hline
		\textbf{Link} & \url{https://huggingface.co/datasets/UCSC-VLAA/MedTrinity-25M} \\ \hline
	\end{tabular}
\end{table}

\subsubsection{Supervised cancer diagnosis}
\begin{table}[!h]
	\centering
	\caption{Statistics of supervised cancer diagnosis dataset.}
	\begin{tabular}{|>{\raggedright\arraybackslash}p{2.2cm}|>{\raggedright\arraybackslash}p{10.8cm}|}
		\hline
		\multicolumn{2}{|c|}{\textbf{Camelyon16}~\cite{bejnordi2017diagnostic}} \\ \hline
		\textbf{Patch number} & 4,809,940 \\ \hline
		\textbf{Classes} & \{"normal": 0, "tumor": 1\} \\ \hline
		\textbf{Splits} & \{"train": 270, "test": 129, "val": -\} \\ \hline
		\textbf{Description} & Camelyon16 is a histopathology dataset containing 399 whole-slide images of sentinel lymph nodes collected from two Dutch medical centers, designed for detecting lymph node metastases with training data including 270 WSIs and test data of 129 WSIs. \\ \hline
		\textbf{Link} & \url{https://camelyon16.grand-challenge.org/} \\ \hline
		\multicolumn{2}{|c|}{\textbf{TCGA-NSCLC}~\cite{weinstein2013cancer}} \\ \hline
		\textbf{Patch number} & 4,377,051 \\ \hline
		\textbf{Classes} & \{"LUAD": 0, "LUSC": 1\} \\ \hline
		\textbf{Splits} & \{"train": 836, "test": 210, "val": -\} \\ \hline
		\textbf{Description} & TCGA-NSCLC contains 1,046 histopathology slides of Non-Small Cell Lung Cancer, including 534 lung adenocarcinoma and 512 lung squamous cell carcinoma cases. \\ \hline
		\textbf{Link} & \url{https://www.cancer.gov/ccg/research/genome-sequencing/tcga} \\ \hline
		\multicolumn{2}{|c|}{\textbf{BRACS7/BRACS3}~\cite{brancati2022bracs}} \\ \hline
		\textbf{Patch number} & 45,390 \\ \hline
		\textbf{Classes-3} & \{"Benign Tumors": 0, "Atypical Tumors": 1, "Malignant Tumors": 2\} \\ \hline
		\textbf{Classes-7} & \{"Normal": 0, "Pathological Benign": 1, "Usual Ductal Hyperplasia": 2, "Flat Epithelial Atypia": 3, "Atypical Ductal Hyperplasia": 4, "Ductal Carcinoma In Situ": 5, "Invasive Carcinoma": 6\} \\ \hline
		\textbf{Splits} & \{"train": 599, "test": 166, "val": 89\} \\ \hline
		\textbf{Description} & A dataset of hematoxylin and eosin histopathological images for automated detection/classification of breast tumors. \\ \hline
		\textbf{Link} & \url{https://www.bracs.icar.cnr.it/} \\ \hline
	\end{tabular}
\end{table}

\clearpage
\subsubsection{Visual question answering}

\begin{table}[h]
	\centering
	\caption{Statistics of VQA dataset.}
	\begin{tabular}{|>{\raggedright\arraybackslash}p{2cm}|>{\raggedright\arraybackslash}p{11cm}|}
		\hline
		\multicolumn{2}{|c|}{\textbf{VQA-RAD}~\cite{lau2018dataset}} \\ \hline
		\textbf{Splits} & \{"train": 2,298, "test": 766, "val": 451\} \\ \hline
		\textbf{Description} & The VQA-RAD collection was created through manual curation, consisting of 3,515 question-answer combinations in which medical professionals pose natural language inquiries regarding radiological images and supply corresponding reference responses.  \\
		\hline
		\textbf{Link} & \url{https://osf.io/89kps/} \\ \hline
		\multicolumn{2}{|c|}{\textbf{SLAKE}~\cite{liu2021slake}} \\ \hline
		\textbf{Splits} & \{"train": 3,690, "test": 1,229, "val": 1,061\} \\ \hline
		\textbf{Description} & A dual-language radiological visual question answering collection comprising 642 images and 14,000 questions, from which only the English portion was utilized in this study. \\
		\hline
		\textbf{Link} & \url{https://www.med-vqa.com/slake/} \\ \hline
	\end{tabular}
\end{table}

\subsubsection{Survival prediction}

\begin{table}[h]
	\centering
	\caption{Statistics of survival prediction dataset.}
	\begin{tabular}{|>{\raggedright\arraybackslash}p{3cm}|>{\raggedright\arraybackslash}p{10cm}|}
		\hline
		\multicolumn{2}{|c|}{\textbf{TCGA-UT}~\cite{komura2022universal}} \\ \hline
		\textbf{Splits} & 5-fold cross validation \\ \hline
		\textbf{Description} & The TCGA-UT dataset is a large-scale collection of histopathological image patches from human cancer tissues, comprising 1,608,060 patch images sourced from hematoxylin-eosin stained tissue specimens spanning 32 distinct categories of solid tumor malignancies. The study used 138,220 image patches covering 12 types of cancer.  \\
		\hline
		\textbf{Patch number:} &   \\
		\textbf{BLCA} & 9,990 \\
		\textbf{BRCA} & 23,690 \\  
		\textbf{CESC} & 6,270 \\  
		\textbf{HNSC} & 11,790 \\  
		\textbf{KIRC} & 11,650 \\  
		\textbf{LUAD} & 16,460 \\  
		\textbf{LUSC} & 16,560 \\  
		\textbf{MESO} & 2,090 \\  
		\textbf{PAAD} & 4,090 \\  
		\textbf{SARC} & 13,480 \\  
		\textbf{STAD} & 9,670 \\  
		\textbf{UCEC} & 12,480 \\  
		\hline
		\textbf{Link} & \url{https://zenodo.org/records/5889558} \\ \hline
	\end{tabular}
\end{table}

\subsection{Quantitative experimental results}

\begin{table}[htbp]
	\centering
	\caption{Zero-shot classification task results across 9 medical imaging modalities. \textbf{Bold} indicates best result, \underline{underline} indicates second best result. 95\% CI is included in parentheses.}
	\label{tab:Zero-shot classification}%
	\scriptsize
	\setlength{\tabcolsep}{1.5pt}
	\renewcommand{\arraystretch}{1.16}
	
	\begin{tabular}{l|cccccccccc}
		\toprule
		Dataset & BMCA & BiomedCLIP & GenMedClip & MedCLIP & PLIP & PMCCLIP & PUBMEDCLIP & QuiltNet & UniMedCLIP & MMKD-CLIP \\
		\midrule
		\multicolumn{11}{c}{X-Ray} \\
		\midrule
		\multirow{2}[1]{*}{CovidCXR4} & 57.59 & 61.72 & 57.53 & 54.95 & 43.95 & 57.9  & 55.73 & 59.55 & \textbf{64.13} & \underline{62.77} \\
		& (56.30,58.77) & (60.49,62.91) & (56.35,58.75) & (53.73,56.23) & (42.63,45.12) & (56.65,59.06) & (54.54,56.87) & (58.38,60.69) & (62.99,65.25) & (61.52,63.95) \\
		\multirow{2}[0]{*}{DDSM} & 61.5  & 78.58 & 70.22 & 61.7  & 46.7  & 73.89 & 59.04 & \underline{84.69} & 66.98 & \textbf{85.85} \\
		& (59.78,63.24) & (76.99,80.06) & (68.50,71.71) & (59.75,63.55) & (44.83,48.55) & (72.31,75.43) & (57.12,60.88) & (83.39,85.91) & (65.17,68.82) & (84.54,86.98) \\
		\multirow{2}[0]{*}{NLMTB} & 79.06 & 91.2  & 85.32 & 32.63 & 67.8  & 83.42 & 55.88 & \textbf{95.26} & 80.71 & \underline{93.84} \\
		& (73.85,83.76) & (87.77,94.32) & (80.54,90.24) & (27.76,37.85) & (62.90,72.87) & (79.22,87.33) & (49.62,62.21) & (93.22,97.00) & (74.92,86.30) & (91.04,96.20) \\
		\multirow{2}[0]{*}{SIIMACR} & 71.15 & 77.51 & 73.84 & 85.2  & 49.91 & 67.37 & 61.37 & 74.55 & \textbf{92.86} & \underline{90.48} \\
		& (67.95,74.23) & (74.62,80.39) & (70.77,76.98) & (82.77,87.64) & (46.25,53.26) & (64.10,70.89) & (57.75,64.86) & (71.47,77.45) & (91.16,94.34) & (88.55,92.26) \\
		\multirow{2}[0]{*}{rsna\_pneum} & 82.47 & \underline{86.67} & 84.06 & 46.64 & 37.11 & 86.13 & 73.41 & \textbf{86.94} & 80.61 & 84.32 \\
		& (81.33,83.57) & (85.68,87.53) & (83.05,85.07) & (44.97,48.28) & (35.60,38.65) & (85.19,87.09) & (71.95,74.72) & (86.06,87.78) & (79.45,81.82) & (83.28,85.30) \\

		\multirow{2}[1]{*}{Average} & 70.35 & 79.13 & 74.19 & 56.22 & 49.1  & 73.74 & 61.09 & \underline{80.20} & 77.06 & \textbf{83.45} \\ 
		& (67.84,72.72) & (77.11,81.04) & (71.84,76.55) & (53.80,58.71) & (46.44,51.69) & (71.49,75.96) & (58.20,63.91) & (78.50,81.77) & (74.74,79.31) & (81.79,84.94) \\
		\midrule
		\multicolumn{11}{c}{Fundus} \\
		\midrule
		\multirow{2}[1]{*}{DRD} & 64.07 & 63.84 & 60.64 & 53.21 & 49.7  & 65.2  & 52.78 & 63.96 & \underline{68.39} & \textbf{71.29} \\
		& (62.55,65.44) & (62.40,65.18) & (59.10,62.07) & (51.46,54.85) & (48.27,51.21) & (63.87,66.39) & (51.32,54.33) & (62.58,65.25) & (67.08,69.63) & (70.18,72.43) \\
		\multirow{2}[0]{*}{FundusJSIEC} & 74.19 & 82.05 & 78.48 & 56.4  & 59.85 & 81.52 & 62.13 & 82.41 & \underline{87.58} & \textbf{89.34} \\
		& (71.66,76.60) & (79.99,84.03) & (76.52,80.56) & (53.10,59.87) & (56.36,63.45) & (79.06,84.24) & (58.93,64.93) & (80.11,84.51) & (85.86,89.63) & (87.79,90.92) \\
		\multirow{2}[0]{*}{five\_retina} & 63.3  & 66.5  & 82.26 & 47.04 & 48.83 & 82.89 & 47.04 & 75.09 & \textbf{91.09} & \underline{87.14} \\
		& (58.67,67.30) & (61.53,71.17) & (79.20,85.07) & (41.64,52.07) & (44.17,53.89) & (80.46,85.19) & (41.67,52.12) & (71.69,78.53) & (88.46,93.51) & (84.11,90.15) \\
		\multirow{2}[1]{*}{Average} & 67.19 & 70.79 & 73.79 & 52.22 & 52.79 & 76.54 & 53.98 & 73.82 & \underline{82.35} & \textbf{82.59} \\
		& (64.29,69.78) & (67.97,73.46) & (71.61,75.90) & (48.73,55.60) & (49.60,56.18) & (74.47,78.60) & (50.64,57.13) & (71.46,76.10) & (80.47,84.26) & (80.70,84.50) \\
		\midrule
		\multicolumn{11}{c}{Pathology} \\
		\midrule
		\multirow{2}[1]{*}{BACH} & 69.87 & 75.67 & 84.09 & 43.59 & 72.07 & 62.44 & 47.6  & \textbf{86.40} & 82.89 & \underline{86.03} \\
		& (63.95,75.58) & (69.40,81.80) & (79.53,88.05) & (37.80,49.55) & (65.64,78.08) & (55.12,69.66) & (41.02,54.58) & (82.47,90.46) & (77.59,87.69) & (82.29,89.77) \\
		\multirow{2}[0]{*}{LC25000\_col} & 96.52 & 98.72 & 96.46 & 57.18 & 98.52 & \underline{99.02} & 82.69 & 98.89 & \textbf{99.37} & 98.89 \\
		& (95.82,97.16) & (98.36,99.07) & (95.69,97.12) & (54.57,59.70) & (98.13,98.87) & (98.68,99.33) & (80.82,84.42) & (98.55,99.20) & (99.14,99.59) & (98.50,99.21) \\
		\multirow{2}[0]{*}{LC25000\_lng} & 91.79 & 86.69 & 91.18 & 45.88 & 91.69 & 92.01 & 44.43 & 89.15 & \textbf{92.63} & \underline{92.61} \\
		& (91.16,92.45) & (85.84,87.53) & (90.49,91.88) & (44.45,47.25) & (90.91,92.44) & (91.42,92.59) & (43.30,45.60) & (88.46,89.92) & (91.99,93.33) & (91.98,93.29) \\
		\multirow{2}[0]{*}{NCT-CRC} & 85.64 & \textbf{91.88} & 84.15 & 49.08 & \underline{89.99} & 84.19 & 63.32 & 86.43 & 89.49 & 89.09 \\
		& (85.31,85.95) & (91.62,92.15) & (83.68,84.65) & (48.51,49.65) & (89.68,90.33) & (83.87,84.51) & (62.85,63.81) & (85.95,86.86) & (89.14,89.84) & (88.74,89.41) \\
		\multirow{2}[0]{*}{Osteo} & 65.93 & 76.86 & 75.36 & 38.23 & 78.9  & 61.11 & 41.5  & 79.07 & \underline{90.05} & \textbf{90.20} \\
		& (61.20,70.29) & (71.13,82.07) & (69.00,80.64) & (33.77,42.54) & (73.87,83.42) & (55.60,66.64) & (36.58,47.01) & (73.96,83.95) & (86.38,93.19) & (86.59,93.33) \\
		\multirow{2}[0]{*}{PCAM} & \underline{82.71} & 78.14 & 71.74 & 64.45 & \textbf{83.98} & 76.33 & 55.67 & 76.82 & 60.83 & 80.99 \\
		& (82.29,83.17) & (77.69,78.61) & (71.20,72.29) & (63.90,65.01) & (83.54,84.40) & (75.82,76.82) & (55.08,56.29) & (76.31,77.31) & (60.20,61.42) & (80.52,81.46) \\
		\multirow{2}[0]{*}{kather\_et\_al\_2016} & 80.85 & \underline{81.19} & 72.25 & 43.09 & \textbf{86.22} & 75.22 & 70.62 & 75.41 & 67.23 & 80.01 \\
		& (77.89,83.63) & (77.96,84.18) & (68.84,75.87) & (39.14,47.13) & (83.44,88.77) & (71.98,78.38) & (67.17,73.91) & (72.40,78.30) & (63.20,70.81) & (76.73,82.97) \\
		\multirow{2}[0]{*}{kather\_et\_al\_2018} & 91.57 & 90.96 & 78.48 & 49.49 & \textbf{93.68} & \underline{92.64} & 70.43 & 87.53 & 84.2  & 90.67 \\
		& (90.04,92.90) & (89.43,92.36) & (76.31,81.03) & (46.28,52.52) & (92.60,94.76) & (91.37,93.80) & (68.26,72.64) & (85.94,89.13) & (82.22,86.17) & (88.91,92.19) \\
		\multirow{2}[0]{*}{kather\_2018\_val7k} & 85.39 & 87.61 & 79    & 49.58 & \textbf{90.97} & \underline{90.87} & 70.03 & 80.83 & 85.98 & 88.85 \\
		& (83.39,87.53) & (85.44,89.59) & (76.59,81.33) & (46.01,53.27) & (89.66,92.31) & (89.09,92.80) & (67.31,72.53) & (78.91,82.77) & (83.87,87.83) & (86.98,90.65) \\
		\multirow{2}[0]{*}{skin\_cancer} & 77.9  & 77.71 & \textbf{89.89} & 50.09 & 84.21 & 81.71 & 51.76 & 82.63 & \underline{87.85} & 87.24 \\
		& (77.49,78.35) & (77.22,78.23) & (89.48,90.24) & (49.52,50.64) & (83.93,84.49) & (81.31,82.10) & (51.29,52.26) & (82.26,82.98) & (87.57,88.13) & (86.89,87.59) \\
		\multirow{2}[0]{*}{tang\_2019} & 81.28 & \underline{82.36} & 70.87 & 46.12 & 46.55 & 78.09 & 71.38 & \textbf{84.02} & 70.34 & 77.44 \\
		& (76.24,85.87) & (77.01,86.77) & (66.56,76.08) & (38.95,52.46) & (40.43,51.73) & (72.32,83.67) & (66.53,76.65) & (78.82,89.00) & (64.52,75.57) & (72.45,82.02) \\
		\multirow{2}[0]{*}{wong\_2022} & 63.82 & 80.15 & 60.38 & 48.48 & 54.05 & \underline{80.17} & 61.72 & \textbf{82.41} & 75.67 & 77.34 \\
		& (58.53,68.64) & (75.88,84.24) & (54.74,66.02) & (41.94,54.50) & (49.05,58.98) & (75.89,84.21) & (57.08,66.23) & (78.64,86.01) & (72.13,79.26) & (73.33,81.17) \\
		\multirow{2}[1]{*}{Average} & 81.11 & 83.99 & 79.49 & 48.77 & 80.9  & 81.15 & 60.93 & \underline{84.13} & 82.21 & \textbf{86.61} \\
		& (78.61,83.46) & (81.41,86.38) & (76.84,82.10) & (45.40,52.02) & (78.41,83.22) & (78.54,83.71) & (58.11,63.83) & (81.89,86.32) & (79.83,84.40) & (84.49,88.59) \\
		\midrule
		\multicolumn{11}{c}{Endoscopy} \\
		\midrule
		\multirow{2}[1]{*}{Kvasir} & 90.09 & 92.11 & 89.88 & 47.48 & 68.82 & 87.74 & 61.71 & \underline{92.72} & 90.49 & \textbf{93.79} \\
		& (89.52,90.67) & (91.51,92.63) & (89.28,90.41) & (45.95,49.04) & (67.69,69.93) & (86.98,88.44) & (60.41,62.99) & (92.27,93.18) & (89.91,91.04) & (93.35,94.22) \\
		\multirow{2}[0]{*}{WCE} & 93.63 & \textbf{95.29} & 92.61 & 52.23 & 62.34 & 88.26 & 61.57 & \underline{94.75} & 87.82 & 94.57 \\
		& (92.66,94.51) & (94.43,96.06) & (91.44,93.63) & (49.98,54.62) & (59.90,64.67) & (86.95,89.37) & (59.22,64.09) & (93.85,95.56) & (86.39,89.13) & (93.62,95.47) \\
		\multirow{2}[1]{*}{Average} & 91.86 & 93.7  & 91.25 & 49.86 & 65.58 & 88    & 61.64 & \underline{93.73} & 89.16 & \textbf{94.18} \\
		& (91.09,92.59) & (92.97,94.34) & (90.36,92.02) & (47.96,51.83) & (63.80,67.30) & (86.97,88.90) & (59.81,63.54) & (93.06,94.37) & (88.15,90.08) & (93.49,94.84) \\
		\bottomrule
	\end{tabular}%
	
	\vspace{2pt}
	\begin{flushright}
		\footnotesize Continued on next page
	\end{flushright}
\end{table}%

\clearpage

\begin{table}[htbp]
	\centering
	\scriptsize
	\setlength{\tabcolsep}{1.5pt}
	\renewcommand{\arraystretch}{1.2}
	
	\begin{tabular}{l|cccccccccc}
		\toprule
		Dataset & BMCA & BiomedCLIP & GenMedClip & MedCLIP & PLIP & PMCCLIP & PUBMEDCLIP & QuiltNet & UniMedCLIP & MMKD-CLIP \\
		\midrule
		\multicolumn{11}{c}{CT} \\
		\midrule
		\multirow{2}[1]{*}{BrainTumorCT} & 85.82 & 95.24 & 92.18 & 44.58 & 19.04 & 91.61 & 45.81 & \underline{96.17} & 96.03 & \textbf{97.85} \\
		& (83.34,88.19) & (93.85,96.41) & (90.38,93.68) & (41.04,48.27) & (16.17,21.77) & (89.79,93.27) & (42.12,49.54) & (95.02,97.20) & (94.75,97.17) & (97.06,98.57) \\
		\multirow{2}[0]{*}{CT\_axial} & 82.13 & 80.76 & 78.18 & 45.72 & 64.46 & 82.58 & \underline{84.37} & 76.99 & 84.19 & \textbf{85.49} \\
		& (80.61,83.66) & (79.17,82.34) & (76.48,80.01) & (43.25,48.17) & (62.31,66.44) & (81.14,84.03) & (83.29,85.54) & (75.18,78.83) & (82.70,85.72) & (84.08,86.91) \\
		\multirow{2}[0]{*}{CT\_coronal} & 75.99 & 75.69 & 75    & 46.54 & 57.46 & 71.98 & 76.49 & 72.99 & \underline{76.53} & \textbf{77.98} \\
		& (74.27,77.58) & (73.72,77.51) & (73.02,76.98) & (43.94,49.24) & (55.00,60.05) & (70.00,74.07) & (74.85,78.13) & (71.03,74.95) & (74.63,78.24) & (76.12,79.72) \\
		\multirow{2}[0]{*}{CT\_sagittal} & 78.98 & 75.39 & 77.39 & 45.38 & 56.56 & 75.33 & 76.09 & 74.07 & \underline{79.93} & \textbf{81.72} \\
		& (77.40,80.59) & (73.66,76.91) & (75.77,79.06) & (42.87,47.95) & (54.29,58.76) & (73.35,77.05) & (74.49,77.74) & (72.14,76.09) & (78.29,81.65) & (79.99,83.31) \\
		\multirow{2}[0]{*}{CovidCT3A} & 73.73 & 88.27 & 81.46 & 48.72 & 43.99 & 86.24 & 68.25 & 87.37 & \underline{90.55} & \textbf{91.80} \\
		& (73.34,74.10) & (88.02,88.50) & (81.11,81.83) & (48.26,49.19) & (43.58,44.44) & (85.99,86.50) & (67.82,68.66) & (87.08,87.64) & (90.33,90.78) & (91.58,92.02) \\
		\multirow{2}[0]{*}{organamnist} & 77.8  & 76.73 & 73.46 & 45.57 & 51.65 & 76.81 & 78.73 & 74.28 & \textbf{82.07} & \underline{79.86} \\
		& (77.49,78.11) & (76.42,77.06) & (73.08,73.83) & (45.16,46.03) & (51.25,52.04) & (76.46,77.12) & (78.45,79.00) & (73.93,74.63) & (81.81,82.33) & (79.54,80.16) \\
		\multirow{2}[1]{*}{Average} & 79.07 & 82.01 & 79.61 & 46.08 & 48.86 & 80.76 & 71.62 & 80.31 & \underline{84.88} & \textbf{85.78} \\
		& (77.74,80.37) & (80.81,83.12) & (78.31,80.90) & (44.09,48.14) & (47.10,50.58) & (79.45,82.01) & (70.17,73.10) & (79.06,81.56) & (83.75,85.98) & (84.73,86.78) \\
		\midrule
		\multicolumn{11}{c}{MRI} \\
		\midrule
		\multirow{2}[1]{*}{BrainTumorMRI} & \textbf{99.48} & 98.73 & 99.29 & 60.65 & 85.81 & 98.49 & 84.45 & 99.12 & 94.73 & \underline{99.46} \\
		& (99.19,99.72) & (98.16,99.25) & (98.92,99.60) & (57.24,63.98) & (83.24,88.08) & (97.92,99.00) & (81.84,86.66) & (98.71,99.47) & (93.20,96.07) & (99.08,99.77) \\
		\multirow{2}[0]{*}{BrainTumorMRI2} & 79.23 & 83.34 & 87.38 & 53.64 & 54.75 & \underline{87.44} & 67.2  & 84.08 & 86.02 & \textbf{87.55} \\
		& (77.82,80.46) & (82.10,84.48) & (86.36,88.31) & (51.77,55.46) & (52.72,56.53) & (86.19,88.59) & (65.47,68.94) & (83.07,85.10) & (84.91,87.05) & (86.45,88.60) \\
		\multirow{2}[0]{*}{acl\_mri} & 49.89 & 47.89 & 65.24 & 68.04 & 52.85 & 67.12 & 64.68 & 50.62 & \textbf{85.82} & \underline{83.28} \\
		& (41.29,57.62) & (40.10,55.71) & (57.66,72.71) & (60.40,75.12) & (44.84,61.18) & (59.99,74.70) & (57.12,72.18) & (42.82,58.47) & (80.25,90.93) & (77.31,88.53) \\
		\multirow{2}[1]{*}{Average} & 76.2  & 76.65 & 83.97 & 60.77 & 64.47 & 84.35 & 72.11 & 77.94 & \underline{88.86} & \textbf{90.09} \\
		& (72.77,79.27) & (73.45,79.81) & (80.98,86.87) & (56.47,64.86) & (60.27,68.59) & (81.37,87.43) & (68.14,75.93) & (74.87,81.01) & (86.12,91.35) & (87.61,92.30) \\
		\midrule
		\multicolumn{11}{c}{Ultrasound} \\
		\midrule
		\multirow{2}[1]{*}{BUSBRA} & 58.38 & \underline{70.26} & 58.99 & 47.43 & 52.15 & 53.24 & 61.75 & 69.2  & 45.37 & \textbf{70.92} \\
		& (52.43,64.17) & (63.97,76.22) & (52.46,65.74) & (41.33,54.09) & (45.54,58.29) & (47.25,60.16) & (54.92,67.92) & (63.17,75.23) & (39.31,51.70) & (65.13,76.73) \\
		\multirow{2}[0]{*}{breast\_us} & 62.87 & 68.12 & 65.57 & 54.94 & 40.34 & 68.76 & 36.63 & \textbf{72.61} & 66.31 & \underline{72.40} \\
		& (51.40,73.39) & (58.17,77.38) & (54.73,76.48) & (45.60,64.73) & (31.82,50.27) & (59.17,78.72) & (28.22,46.05) & (63.51,81.81) & (57.34,75.52) & (62.71,81.51) \\
		\multirow{2}[0]{*}{breastmnist} & 67.4  & \underline{78.74} & 76.82 & 67.18 & 60.67 & 66.08 & 55.85 & \textbf{83.00} & 62.03 & 73.08 \\
		& (58.12,75.78) & (70.54,86.53) & (67.91,85.09) & (57.29,75.73) & (50.96,70.23) & (56.09,75.14) & (45.92,65.57) & (74.88,90.62) & (52.24,71.31) & (62.86,82.65) \\
		\multirow{2}[1]{*}{Average} & 62.88 & \underline{72.37} & 67.13 & 56.52 & 51.05 & 62.69 & 51.41 & \textbf{74.94} & 57.9  & 72.13 \\
		& (53.98,71.11) & (64.22,80.04) & (58.37,75.77) & (48.07,64.85) & (42.77,59.60) & (54.17,71.34) & (43.02,59.85) & (67.19,82.55) & (49.63,66.18) & (63.57,80.30) \\
		\midrule
		\multicolumn{11}{c}{Dermatology} \\
		\midrule
		\multirow{2}[1]{*}{HAM10000} & \textbf{78.26} & \underline{73.23} & 67.24 & 49.62 & 59.2  & 71.37 & 65.44 & 68.09 & 63.89 & 71.74 \\
		& (76.07,80.28) & (71.10,75.25) & (65.17,69.19) & (47.00,52.28) & (56.96,61.28) & (69.28,73.18) & (62.93,67.89) & (65.08,70.78) & (61.50,66.15) & (69.21,74.02) \\
		\multirow{2}[0]{*}{PADUFES20} & \textbf{83.39} & \underline{81.22} & 77.15 & 54.44 & 59.86 & 68.48 & 70.08 & 73.17 & 71.98 & 80.79 \\
		& (81.36,85.35) & (78.92,83.28) & (74.53,79.54) & (49.87,59.15) & (55.95,63.40) & (65.07,71.82) & (67.09,72.83) & (70.36,75.77) & (68.89,74.96) & (78.73,82.71) \\
		\multirow{2}[1]{*}{Average} & \textbf{80.83} & \underline{77.23} & 72.19 & 52.03 & 59.53 & 69.92 & 67.76 & 70.63 & 67.94 & 76.26 \\
		& (78.72,82.82) & (75.01,79.27) & (69.85,74.36) & (48.44,55.71) & (56.45,62.34) & (67.18,72.50) & (65.01,70.36) & (67.72,73.28) & (65.19,70.56) & (73.97,78.37) \\
		\midrule
		\multicolumn{11}{c}{OCT} \\
		\midrule
		\multirow{2}[1]{*}{OCTMNIST} & 71.28 & 80.16 & 85.36 & 57.78 & 53.46 & 92.42 & 68.89 & 83.57 & \underline{94.32} & \textbf{95.10} \\
		& (69.49,73.35) & (79.00,81.31) & (84.13,86.56) & (55.86,59.73) & (51.26,55.76) & (91.60,93.11) & (67.01,70.86) & (82.40,84.81) & (93.65,95.00) & (94.41,95.75) \\
		\multirow{2}[0]{*}{RetinalOCT} & 55.71 & 68.43 & 71.17 & 54.71 & 56.69 & \textbf{81.90} & 52.14 & 73.49 & 77.29 & \underline{79.71} \\
		& (54.75,56.64) & (67.57,69.30) & (70.41,71.91) & (53.70,55.85) & (55.63,57.75) & (81.20,82.57) & (50.98,53.16) & (72.72,74.32) & (76.51,78.04) & (79.06,80.39) \\
		\multirow{2}[1]{*}{Average} & 63.49 & 74.3  & 78.27 & 56.25 & 55.07 & \underline{87.16} & 60.52 & 78.53 & 85.81 & \textbf{87.40} \\
		& (62.12,65.00) & (73.29,75.30) & (77.27,79.24) & (54.78,57.79) & (53.44,56.76) & (86.40,87.84) & (59.00,62.01) & (77.56,79.57) & (85.08,86.52) & (86.74,88.07) \\
		\bottomrule
	\end{tabular}%
\end{table}%

\newpage

\begin{table}[htbp]
	\centering
	\caption{Linear probe classification task results across 9 different modalities (1\% training data). \textbf{Bold} indicates best result, \underline{underline} indicates second best result. 95\% CI is included in parentheses.}
	\vspace{0.3cm}
	\label{tab:linear_probe_results}%
	\scriptsize
	\setlength{\tabcolsep}{1.5pt}
	\renewcommand{\arraystretch}{1.2}
	
	\begin{tabular}{l|cccccccccc}
		\toprule
		Dataset & BMCA & BiomedCLIP & GenMedClip & MedCLIP & PLIP & PMCCLIP & PUBMEDCLIP & QuiltNet & UniMedCLIP & MMKD-CLIP \\
		\midrule
		\multicolumn{11}{c}{X-Ray} \\
		\midrule
		\multirow{2}[1]{*}{CovidCXR4} & 60.43 & 58.44 & 56.35 & 56.51 & 59.66 & 57.28 & 55.01 & 68.51 & \textbf{72.73} & \underline{70.81} \\
		& (59.25,61.60) & (57.27,59.63) & (55.05,57.54) & (55.31,57.78) & (58.37,60.86) & (56.03,58.43) & (53.83,56.26) & (67.37,69.62) & (71.64,73.79) & (69.64,71.94) \\
		\multirow{2}[0]{*}{DDSM} & \textbf{95.18} & 92.23 & 88.35 & 66.75 & 88.06 & 90.09 & 80.78 & 93.1  & 94.36 & \underline{95.15} \\
		& (94.51,95.75) & (91.41,93.03) & (87.13,89.45) & (64.96,68.50) & (87.01,88.94) & (89.18,90.93) & (79.49,82.06) & (92.35,93.86) & (93.67,95.04) & (94.50,95.83) \\
		\multirow{2}[0]{*}{NLMTB} & 98.45 & 94.55 & 93.11 & 87.46 & 90.09 & 95.11 & 92.25 & 92.19 & \textbf{99.88} & \underline{99.36} \\
		& (97.64,99.14) & (91.95,96.68) & (90.84,95.10) & (83.14,91.45) & (86.95,92.75) & (93.34,96.65) & (89.67,94.48) & (89.43,94.71) & (99.74,99.98) & (98.98,99.66) \\
		\multirow{2}[0]{*}{SIIMACR} & 70.41 & 71.83 & 69.65 & \underline{82.48} & 61.72 & 65.68 & 60.36 & 79.39 & 78.5  & \textbf{87.46} \\
		& (67.19,73.56) & (68.60,75.20) & (66.37,72.63) & (79.96,84.92) & (58.21,65.18) & (62.34,68.74) & (57.26,64.06) & (76.63,82.25) & (75.71,81.20) & (85.17,89.71) \\
		\multirow{2}[0]{*}{Rsna\_pneumonia} & 83.94 & 86.29 & 85.9  & \textbf{88.51} & 79.01 & 82.41 & 80.78 & \underline{88.32} & 88.17 & 87.35 \\
		& (82.89,84.98) & (85.34,87.22) & (84.91,86.82) & (87.65,89.32) & (77.80,80.19) & (81.28,83.50) & (79.66,81.88) & (87.44,89.12) & (87.33,89.04) & (86.45,88.25) \\
		\multirow{2}[1]{*}{Average} & 81.68 & 80.67 & 78.67 & 76.34 & 75.71 & 78.11 & 73.84 & 84.3  & \underline{86.73} & \textbf{88.03} \\
		& (80.29,83.00) & (78.91,82.35) & (76.86,80.31) & (74.20,78.39) & (73.67,77.58) & (76.44,79.65) & (71.98,75.75) & (82.64,85.91) & (85.62,87.81) & (86.95,89.08) \\
		\midrule
		\multicolumn{11}{c}{Fundus} \\
		\midrule
		\multirow{2}[1]{*}{DRD} & 63.83 & \underline{64.18} & 61.05 & 55.25 & 59    & 59    & 58.82 & 63.21 & 63.26 & \textbf{65.41} \\
		& (62.51,65.13) & (62.71,65.51) & (59.54,62.51) & (53.72,56.69) & (57.49,60.48) & (57.55,60.41) & (57.24,60.25) & (61.81,64.64) & (61.94,64.67) & (64.13,66.77) \\
		\multirow{2}[0]{*}{FundusJSIEC} & 86.87 & \underline{87.33} & 80.45 & 72.1  & 82.06 & 83.44 & 79.48 & 85.66 & 87.21 & \textbf{90.37} \\
		& (84.69,88.89) & (85.64,89.03) & (78.10,82.98) & (69.30,74.92) & (79.76,84.74) & (81.35,85.45) & (76.69,82.10) & (83.65,87.80) & (84.85,89.43) & (88.61,91.79) \\
		\multirow{2}[0]{*}{Five\_retina} & 68.26 & 56.83 & \underline{76.87} & 54.94 & 52.29 & 73.43 & 64.7  & 58.92 & 71.75 & \textbf{83.52} \\
		& (64.18,72.49) & (52.19,61.64) & (72.83,80.61) & (49.89,60.01) & (47.24,57.45) & (69.54,77.28) & (60.52,69.21) & (54.66,63.60) & (67.08,76.08) & (80.33,86.77) \\
		\multirow{2}[1]{*}{Average} & 72.99 & 69.45 & 72.79 & 60.76 & 64.45 & 71.96 & 67.67 & 69.26 & \underline{74.07} & \textbf{79.76} \\
		& (70.46,75.51) & (66.85,72.06) & (70.16,75.36) & (57.64,63.87) & (61.49,67.56) & (69.48,74.38) & (64.82,70.52) & (66.71,72.01) & (71.29,76.73) & (77.69,81.78) \\
		\midrule
		\multicolumn{11}{c}{Pathology} \\
		\midrule
		\multirow{2}[1]{*}{BACH} & 71.08 & 69.95 & \underline{74.64} & 58.55 & 62.99 & 57.21 & 60.28 & 61.35 & 72.53 & \textbf{76.61} \\
		& (65.15,76.97) & (62.36,77.10) & (67.22,81.36) & (51.35,65.94) & (55.55,69.66) & (48.86,65.15) & (53.06,67.07) & (53.95,68.95) & (65.62,79.08) & (69.97,82.64) \\
		\multirow{2}[0]{*}{LC25000\_colon} & 99.27 & 99.54 & 98.91 & 88.23 & \underline{99.91} & 99.89 & 99.1  & \textbf{99.93} & 99.87 & 99.77 \\
		& (98.97,99.51) & (99.37,99.68) & (98.59,99.18) & (86.73,89.63) & (99.86,99.96) & (99.83,99.95) & (98.74,99.42) & (99.84,99.98) & (99.77,99.94) & (99.66,99.86) \\
		\multirow{2}[0]{*}{LC25000\_lung} & 96.63 & 96.69 & 96.12 & 82.95 & 96.87 & 97.05 & 93.52 & \underline{97.91} & \textbf{98.16} & 97.74 \\
		& (96.22,97.04) & (96.22,97.15) & (95.62,96.55) & (82.02,83.90) & (96.51,97.23) & (96.65,97.43) & (92.99,94.10) & (97.53,98.25) & (97.84,98.49) & (97.40,98.08) \\
		\multirow{2}[0]{*}{NCT-CRC} & 98.84 & \underline{99.36} & 98.65 & 93.4  & \textbf{99.44} & 99.09 & 97.87 & 99.32 & 99.35 & 99.08 \\
		& (98.73,98.95) & (99.29,99.44) & (98.53,98.77) & (93.06,93.79) & (99.37,99.51) & (99.00,99.18) & (97.72,98.01) & (99.24,99.41) & (99.26,99.43) & (98.98,99.18) \\
		\multirow{2}[0]{*}{Osteo} & 72.63 & 74.19 & 65.76 & 40.38 & 71.7  & \underline{79.59} & 66.15 & 70.04 & \textbf{80.97} & 76.54 \\
		& (67.69,77.45) & (69.37,78.88) & (58.99,71.61) & (34.98,46.32) & (66.93,76.56) & (74.78,84.34) & (61.60,70.99) & (65.39,74.80) & (77.06,84.71) & (72.02,80.61) \\
		\multirow{2}[0]{*}{PCAM} & 93.24 & 91.56 & 91.1  & 84.74 & \underline{93.83} & 91.13 & 89.63 & 91.25 & \textbf{94.03} & 93.03 \\
		& (93.00,93.48) & (91.24,91.84) & (90.78,91.40) & (84.33,85.14) & (93.58,94.07) & (90.83,91.42) & (89.30,89.95) & (90.93,91.55) & (93.79,94.26) & (92.77,93.29) \\
		\multirow{2}[0]{*}{Kather\_et\_al\_2016} & 89.03 & 90    & 88.92 & 70.76 & \textbf{92.11} & 89.91 & 85.71 & 88.64 & \underline{91.83} & 89.67 \\
		& (86.53,91.19) & (87.34,92.28) & (86.53,91.17) & (66.85,74.48) & (89.77,94.00) & (87.08,92.45) & (83.02,88.17) & (85.68,91.25) & (88.79,94.14) & (86.82,92.12) \\
		\multirow{2}[0]{*}{Kather\_et\_al\_2018} & \underline{93.90} & 93.02 & 84.65 & 60.52 & 92.41 & 85.86 & 81.07 & 91.49 & 93.58 & \textbf{95.04} \\
		& (92.70,94.99) & (91.68,94.32) & (82.52,86.75) & (57.56,63.39) & (91.23,93.66) & (83.88,87.76) & (79.27,82.88) & (89.88,92.93) & (92.12,94.98) & (93.98,96.14) \\
		\multirow{2}[0]{*}{Kather\_2018\_val7k} & 91.97 & 92.76 & 91.94 & 77.82 & \textbf{95.71} & 89.46 & 92.71 & \underline{95.31} & 90.42 & 92.86 \\
		& (90.15,93.72) & (90.98,94.48) & (90.20,93.53) & (74.38,81.21) & (94.46,96.88) & (86.91,91.77) & (90.92,94.45) & (93.97,96.58) & (87.71,93.01) & (90.86,94.66) \\
		\multirow{2}[0]{*}{Skin\_cancer} & 97.56 & 97.62 & 97.65 & 84.51 & 98.14 & 98.13 & 95.77 & \underline{98.19} & \textbf{98.59} & 98 \\
		& (97.41,97.69) & (97.39,97.81) & (97.51,97.79) & (84.05,84.97) & (98.06,98.21) & (98.03,98.23) & (95.64,95.90) & (98.10,98.29) & (98.50,98.66) & (97.90,98.10) \\
		\multirow{2}[0]{*}{Tang\_et\_al\_2019} & 78.85 & \underline{82.01} & 69.06 & 45.1  & 73.24 & 77.77 & \textbf{82.47} & 81.08 & 71.07 & 72 \\
		& (73.18,84.63) & (75.47,87.33) & (63.55,74.89) & (38.97,51.28) & (66.55,79.69) & (71.16,84.02) & (77.03,87.54) & (75.23,85.75) & (63.65,77.14) & (65.15,78.08) \\
		\multirow{2}[0]{*}{Wong\_et\_al\_2022} & 66.05 & 66.13 & \underline{69.81} & 50.64 & 58.61 & 60.22 & 60.22 & 62.61 & 67.1  & \textbf{73.03} \\
		& (61.03,71.10) & (60.94,70.98) & (65.09,74.62) & (45.73,55.77) & (53.42,63.96) & (55.11,65.66) & (54.82,65.71) & (57.35,67.63) & (61.88,72.34) & (68.20,77.62) \\
		\multirow{2}[1]{*}{Average} & 87.42 & 87.74 & 85.6  & 69.8  & 86.25 & 85.44 & 83.71 & 86.43 & \underline{88.12} & \textbf{88.61} \\
		& (85.06,89.73) & (85.14,90.11) & (82.93,88.14) & (66.67,72.98) & (83.77,88.61) & (82.68,88.11) & (81.18,86.18) & (83.93,88.78) & (85.50,90.51) & (86.14,90.87) \\
		\midrule
		\multicolumn{11}{c}{Endoscopy} \\
		\midrule
		\multirow{2}[1]{*}{Kvasir} & \textbf{95.87} & 93.78 & 94.15 & 74.5  & 90.5  & 93.46 & 88.24 & 92.62 & \underline{94.68} & 94.45 \\
		& (95.53,96.18) & (93.25,94.31) & (93.65,94.59) & (73.34,75.68) & (90.00,91.04) & (92.95,93.95) & (87.53,88.88) & (92.04,93.15) & (94.28,95.07) & (94.09,94.81) \\
		\multirow{2}[0]{*}{WCE} & \textbf{99.14} & 95.33 & \underline{98.05} & 87.24 & 90.63 & 97.13 & 92.18 & 95.12 & 97.07 & 97.9 \\
		& (98.78,99.44) & (94.43,96.32) & (97.49,98.57) & (85.65,88.83) & (89.36,91.84) & (96.39,97.70) & (90.78,93.39) & (94.22,96.04) & (96.43,97.72) & (97.27,98.44) \\
		\multirow{2}[1]{*}{Average} & \textbf{97.50} & 94.56 & 96.1  & 80.87 & 90.56 & 95.29 & 90.21 & 93.87 & 95.88 & \underline{96.18} \\
		& (97.16,97.81) & (93.84,95.32) & (95.57,96.58) & (79.50,82.25) & (89.68,91.44) & (94.67,95.82) & (89.15,91.14) & (93.13,94.59) & (95.36,96.40) & (95.68,96.63) \\
		\bottomrule
	\end{tabular}%
	
	\vspace{2pt}
	\begin{flushright}
		\footnotesize Continued on next page
	\end{flushright}
\end{table}%

\clearpage

\begin{table}[htbp]
	\centering
	\scriptsize
	\setlength{\tabcolsep}{1.5pt}
	\renewcommand{\arraystretch}{1.2}
	
	\begin{tabular}{l|cccccccccc}
		\toprule
		Dataset & BMCA & BiomedCLIP & GenMedClip & MedCLIP & PLIP & PMCCLIP & PUBMEDCLIP & QuiltNet & UniMedCLIP & MMKD-CLIP \\
		\midrule
		\multicolumn{11}{c}{CT} \\
		\midrule
		\multirow{2}[1]{*}{BrainTumorCT} & 95.79 & 97.75 & 95.71 & 85.22 & 91.05 & 95.74 & 90.04 & 95.92 & \textbf{98.97} & \underline{98.20} \\
		& (94.60,96.92) & (96.92,98.45) & (94.36,96.74) & (82.62,87.83) & (89.03,92.77) & (94.49,96.93) & (87.87,92.00) & (94.62,97.01) & (98.40,99.40) & (97.50,98.83) \\
		\multirow{2}[0]{*}{CT\_axial} & 81.91 & 74.48 & 72.78 & 73.13 & 75.96 & \underline{83.29} & \textbf{83.77} & 74.75 & 80.13 & 75.81 \\
		& (80.28,83.50) & (72.60,76.32) & (70.74,74.73) & (71.05,74.83) & (74.59,77.26) & (81.84,84.72) & (82.54,84.99) & (72.91,76.55) & (78.31,81.81) & (73.99,77.69) \\
		\multirow{2}[0]{*}{CT\_coronal} & 72.37 & 68.04 & \textbf{79.30} & 71.94 & 73.39 & \underline{76.86} & 76.53 & 71.09 & 75.85 & 74.64 \\
		& (70.32,74.34) & (65.69,70.29) & (77.37,81.25) & (69.84,74.02) & (71.72,74.83) & (74.84,78.78) & (74.32,78.49) & (69.20,73.14) & (73.91,77.77) & (72.82,76.62) \\
		\multirow{2}[0]{*}{CT\_sagittal} & 80.04 & 75.2  & 79.75 & 69.45 & 75.49 & 81.48 & 77.15 & 75.08 & \textbf{83.04} & \underline{81.87} \\
		& (78.25,81.70) & (73.27,77.30) & (77.93,81.56) & (67.27,71.50) & (73.89,76.97) & (79.80,83.23) & (75.40,79.00) & (73.14,76.95) & (81.30,84.69) & (79.96,83.51) \\
		\multirow{2}[0]{*}{CovidCT3A} & 90.95 & 94.86 & 92.13 & 86.12 & 75.48 & 93.59 & 81    & 94.95 & \textbf{96.46} & \underline{96.38} \\
		& (90.71,91.20) & (94.68,95.04) & (91.90,92.37) & (85.80,86.43) & (75.10,75.88) & (93.39,93.79) & (80.64,81.37) & (94.77,95.12) & (96.30,96.59) & (96.22,96.52) \\
		\multirow{2}[0]{*}{Organamnist} & 96.02 & 95.96 & 94.7  & 93.35 & 93.52 & \textbf{97.20} & 94.03 & 95.16 & \underline{96.60} & 95.52 \\
		& (95.89,96.14) & (95.80,96.11) & (94.54,94.86) & (93.14,93.54) & (93.34,93.69) & (97.11,97.28) & (93.89,94.17) & (94.99,95.31) & (96.48,96.72) & (95.38,95.66) \\
		\multirow{2}[1]{*}{Average} & 86.18 & 84.38 & 85.73 & 79.87 & 80.82 & \underline{88.03} & 83.75 & 84.49 & \textbf{88.51} & 87.07 \\
		& (85.01,87.30) & (83.16,85.59) & (84.47,86.92) & (78.29,81.36) & (79.61,81.90) & (86.91,89.12) & (82.44,85.00) & (83.27,85.68) & (87.45,89.50) & (85.98,88.14) \\
		\midrule
		\multicolumn{11}{c}{MRI} \\
		\midrule
		\multirow{2}[1]{*}{BrainTumorMRI} & \textbf{99.82} & 99.61 & \underline{99.73} & 87.33 & 95.74 & 95.93 & 90.33 & 99.52 & 98.19 & 99.67 \\
		& (99.67,99.93) & (99.37,99.81) & (99.48,99.91) & (84.87,89.58) & (94.55,96.74) & (94.63,97.05) & (88.34,92.09) & (99.25,99.76) & (97.48,98.77) & (99.42,99.84) \\
		\multirow{2}[0]{*}{BrainTumorMRI2} & 91.98 & 91.15 & 93.03 & 78.63 & 83.66 & 92.17 & 85.53 & 87.66 & \textbf{94.44} & \underline{94.03} \\
		& (91.00,92.84) & (89.98,92.22) & (92.19,93.87) & (76.99,80.41) & (82.20,85.09) & (91.27,93.09) & (84.26,86.76) & (86.27,89.01) & (93.61,95.20) & (93.17,94.92) \\
		\multirow{2}[0]{*}{Acl\_mri} & 52.47 & 73.17 & \underline{74.17} & 58.05 & 38.65 & 52.43 & 51.72 & 69.22 & 71.27 & \textbf{89.62} \\
		& (44.13,60.47) & (65.86,80.14) & (66.80,80.29) & (49.89,65.75) & (30.58,46.40) & (43.87,60.41) & (43.07,59.53) & (61.80,76.08) & (64.18,78.14) & (85.06,93.52) \\
		\multirow{2}[1]{*}{Average} & 81.43 & 87.98 & \underline{88.98} & 74.67 & 72.68 & 80.18 & 75.86 & 85.47 & 87.97 & \textbf{94.44} \\
		& (78.27,84.42) & (85.07,90.72) & (86.16,91.36) & (70.59,78.58) & (69.11,76.08) & (76.59,83.52) & (71.89,79.46) & (82.44,88.28) & (85.09,90.70) & (92.55,96.09) \\
		\midrule
		\multicolumn{11}{c}{Ultrasound} \\
		\midrule
		\multirow{2}[1]{*}{BUSBRA} & 56.05 & \textbf{65.36} & 51.1  & \underline{62.05} & 51.56 & 54.36 & 56.29 & 55.11 & 59.31 & 59.64 \\
		& (49.76,62.19) & (59.46,71.46) & (44.57,57.19) & (56.02,68.32) & (44.81,57.97) & (48.40,59.52) & (49.43,62.53) & (49.13,61.25) & (52.96,65.05) & (53.39,65.52) \\
		\multirow{2}[0]{*}{Breast\_us} & 55.79 & \underline{63.79} & \textbf{64.25} & 54.17 & 60.84 & 54.68 & 54.39 & 42.76 & 54.85 & 59.44 \\
		& (46.16,65.59) & (54.15,73.35) & (54.46,73.29) & (44.29,63.08) & (50.62,70.30) & (43.55,65.39) & (43.59,64.67) & (32.57,52.97) & (45.10,64.23) & (49.29,69.77) \\
		\multirow{2}[0]{*}{Breastmnist} & 59.98 & 63.26 & 54.59 & \underline{75.44} & 69.21 & \textbf{76.98} & 64.95 & 72.7  & 56.77 & 67.84 \\
		& (48.66,69.95) & (53.43,72.92) & (43.78,65.07) & (66.55,83.20) & (57.95,80.04) & (67.81,84.88) & (53.03,74.73) & (63.71,80.95) & (46.50,66.31) & (57.38,77.64) \\
		\multirow{2}[1]{*}{Average} & 57.28 & \textbf{64.14} & 56.65 & \underline{63.88} & 60.54 & 62.01 & 58.54 & 56.86 & 56.97 & 62.31 \\
		& (48.19,65.91) & (55.68,72.57) & (47.61,65.18) & (55.62,71.53) & (51.13,69.43) & (53.25,69.93) & (48.68,67.31) & (48.47,65.06) & (48.19,65.20) & (53.35,70.98) \\
		\midrule
		\multicolumn{11}{c}{Dermatology} \\
		\midrule
		\multirow{2}[1]{*}{HAM10000} & \textbf{77.20} & 70.4  & 67.77 & 62.14 & 72.28 & 73.86 & 65.4  & 70.04 & 69.93 & \underline{74.53} \\
		& (74.90,79.28) & (67.97,72.92) & (65.40,70.32) & (59.50,64.72) & (70.07,74.43) & (71.76,75.88) & (62.91,67.77) & (67.32,72.39) & (67.30,72.38) & (72.12,76.57) \\
		\multirow{2}[0]{*}{PADUFES20} & 66.85 & \underline{67.03} & \textbf{71.15} & 55.46 & 55.95 & 52.9  & 60.7  & 62.51 & 55.09 & 59.96 \\
		& (63.31,70.47) & (62.95,71.13) & (67.95,74.16) & (51.41,59.80) & (52.09,59.61) & (49.13,56.57) & (57.09,64.50) & (58.39,66.36) & (51.07,59.13) & (56.42,63.34) \\
		\multirow{2}[1]{*}{Average} & \textbf{72.03} & 68.72 & \underline{69.46} & 58.8  & 64.11 & 63.38 & 63.05 & 66.27 & 62.51 & 67.24 \\
		& (69.11,74.87) & (65.46,72.03) & (66.68,72.24) & (55.45,62.26) & (61.08,67.02) & (60.44,66.22) & (60.00,66.14) & (62.86,69.37) & (59.19,65.75) & (64.27,69.96) \\
		\midrule
		\multicolumn{11}{c}{OCT} \\
		\midrule
		\multirow{2}[1]{*}{OCTMNIST} & 91.66 & 94.5  & 93.18 & 86.13 & 80    & 89.98 & 92.11 & 96.4  & \textbf{97.86} & \underline{97.51} \\
		& (90.60,92.64) & (93.57,95.34) & (92.21,94.12) & (84.65,87.51) & (78.43,81.57) & (88.78,91.18) & (90.94,93.16) & (95.72,97.08) & (97.30,98.34) & (96.90,98.04) \\
		\multirow{2}[0]{*}{RetinalOCT} & 93.31 & 89.97 & 94.48 & 84.27 & 87.54 & 93.71 & 90.18 & 92.52 & \textbf{96.55} & \underline{96.37} \\
		& (92.90,93.69) & (89.35,90.60) & (94.10,94.87) & (83.48,85.01) & (86.99,88.12) & (93.34,94.07) & (89.74,90.62) & (91.98,93.00) & (96.24,96.85) & (96.05,96.66) \\
		\multirow{2}[1]{*}{Average} & 92.48 & 92.23 & 93.83 & 85.2  & 83.77 & 91.85 & 91.15 & 94.46 & \textbf{97.20} & \underline{96.94} \\
		& (91.75,93.17) & (91.46,92.97) & (93.15,94.49) & (84.06,86.26) & (82.71,84.84) & (91.06,92.62) & (90.34,91.89) & (93.85,95.04) & (96.77,97.59) & (96.47,97.35) \\
		\bottomrule
	\end{tabular}%
\end{table}%

\newpage

\clearpage

\begin{table}[htbp]
	\centering
	\caption{Linear probe classification task results across 9 different modalities (10\% training data). \textbf{Bold} indicates best result, \underline{underline} indicates second best result. 95\% CI is included in parentheses.}
	\vspace{0.3cm}
	\label{tab:linear_probe_10_results}%
	\scriptsize
	\setlength{\tabcolsep}{1.5pt}
	\renewcommand{\arraystretch}{1.2}
	
	\begin{tabular}{l|cccccccccc}
		\toprule
		Dataset & BMCA & BiomedCLIP & GenMedClip & MedCLIP & PLIP & PMCCLIP & PUBMEDCLIP & QuiltNet & UniMedCLIP & MMKD-CLIP \\
		\midrule
		\multicolumn{11}{c}{X-Ray} \\
		\midrule
		\multirow{2}[1]{*}{CovidCXR4} & 61.52 & 58.46 & 58.67 & 56.18 & 58.82 & 68.68 & 61.97 & 67.12 & \underline{72.74} & \textbf{75.39} \\
		& (60.31,62.72) & (57.22,59.63) & (57.45,59.77) & (55.05,57.45) & (57.50,60.04) & (67.63,69.80) & (60.75,63.11) & (65.86,68.17) & (71.69,73.81) & (74.40,76.45) \\
		\multirow{2}[0]{*}{DDSM} & \textbf{97.43} & 95    & 93.7  & 87.95 & 92.41 & 96.39 & 91.81 & 96.09 & 96.84 & \underline{96.85} \\
		& (96.99,97.79) & (94.40,95.61) & (92.85,94.46) & (86.87,88.93) & (91.56,93.16) & (95.94,96.84) & (90.98,92.62) & (95.56,96.63) & (96.39,97.32) & (96.40,97.29) \\
		\multirow{2}[0]{*}{NLMTB} & 99.08 & 99.06 & 97.74 & 88.86 & 96.46 & 99.22 & 95.59 & 98.14 & \textbf{99.94} & \underline{99.76} \\
		& (98.07,99.72) & (98.39,99.62) & (96.42,98.83) & (84.79,92.70) & (94.49,98.11) & (98.50,99.75) & (93.83,97.18) & (97.18,99.00) & (99.86,99.99) & (99.57,99.90) \\
		\multirow{2}[0]{*}{SIIMACR} & 75    & 83.3  & 80.49 & 91.18 & 70.85 & 76.63 & 74.56 & 84.84 & \textbf{92.74} & \underline{91.90} \\
		& (72.01,77.91) & (80.90,85.80) & (77.77,83.02) & (89.46,92.80) & (67.83,74.11) & (73.72,79.31) & (71.24,77.52) & (82.55,87.18) & (90.98,94.27) & (90.06,93.45) \\
		\multirow{2}[0]{*}{Rsna\_pneumonia} & 86.56 & 88.79 & 87.74 & \underline{90.26} & 81.8  & 86.34 & 85.21 & 89.18 & \textbf{90.71} & 90.1 \\
		& (85.61,87.45) & (87.92,89.60) & (86.86,88.63) & (89.49,90.98) & (80.66,82.91) & (85.38,87.30) & (84.23,86.23) & (88.35,89.96) & (89.95,91.44) & (89.32,90.85) \\
		\multirow{2}[1]{*}{Average} & 83.92 & 84.92 & 83.67 & 82.88 & 80.07 & 85.45 & 81.83 & 87.07 & \underline{90.59} & \textbf{90.80} \\
		& (82.60,85.12) & (83.77,86.05) & (82.27,84.94) & (81.13,84.57) & (78.41,81.66) & (84.24,86.60) & (80.21,83.33) & (85.90,88.19) & (89.78,91.37) & (89.95,91.59) \\
		\midrule
		\multicolumn{11}{c}{Fundus} \\
		\midrule
		\multirow{2}[1]{*}{DRD} & 75.94 & 71.83 & 70.84 & 67.23 & 70.21 & 74.97 & 67.36 & 72.99 & \underline{76.81} & \textbf{78.33} \\
		& (74.79,77.02) & (70.57,73.14) & (69.61,72.02) & (65.70,68.48) & (68.89,71.39) & (73.92,76.04) & (65.92,68.60) & (71.82,74.21) & (75.70,77.84) & (77.36,79.29) \\
		\multirow{2}[0]{*}{FundusJSIEC} & \underline{89.46} & 87.61 & 84.48 & 71.57 & 82.5  & 86.93 & 80.67 & 85.17 & 89.19 & \textbf{91.25} \\
		& (87.82,90.89) & (85.54,89.12) & (82.64,86.53) & (68.39,75.18) & (79.88,85.12) & (84.39,89.21) & (77.49,83.65) & (83.18,86.95) & (87.02,91.02) & (89.80,92.51) \\
		\multirow{2}[0]{*}{Five\_retina} & 70    & 58.19 & 78.86 & 57.99 & 66.02 & \underline{80.25} & 74.6  & 62.81 & 79.78 & \textbf{84.17} \\
		& (65.81,74.22) & (53.49,63.26) & (75.10,82.37) & (53.02,62.63) & (61.34,70.86) & (76.63,83.40) & (70.64,78.49) & (58.44,67.04) & (75.74,83.48) & (80.98,87.38) \\
		\multirow{2}[1]{*}{Average} & 78.47 & 72.55 & 78.06 & 65.6  & 72.91 & 80.72 & 74.21 & 73.66 & \underline{81.93} & \textbf{84.58} \\
		& (76.14,80.71) & (69.87,75.17) & (75.78,80.31) & (62.37,68.76) & (70.04,75.79) & (78.32,82.89) & (71.35,76.91) & (71.15,76.07) & (79.49,84.11) & (82.71,86.39) \\
		\midrule
		\multicolumn{11}{c}{Pathology} \\
		\midrule
		\multirow{2}[1]{*}{BACH} & 82.43 & 82.77 & 82.47 & 65.67 & 79.4  & 74.21 & 68.57 & 76.64 & \underline{85.00} & \textbf{86.85} \\
		& (76.52,87.91) & (76.33,87.93) & (77.16,87.44) & (57.65,72.64) & (73.94,85.00) & (68.06,81.08) & (61.42,75.50) & (70.83,82.59) & (79.78,89.70) & (82.05,91.36) \\
		\multirow{2}[0]{*}{LC25000\_colon} & 99.84 & 99.93 & 99.69 & 99.66 & 99.98 & \textbf{99.99} & 99.46 & \underline{99.98} & 99.98 & 99.98 \\
		& (99.71,99.94) & (99.88,99.97) & (99.52,99.82) & (99.53,99.78) & (99.95,99.99) & (99.98,100.00) & (99.20,99.69) & (99.96,100.00) & (99.95,99.99) & (99.95,99.99) \\
		\multirow{2}[0]{*}{LC25000\_lung} & 99.04 & 99.18 & 98.49 & 93.43 & 99.26 & \underline{99.42} & 98.08 & 99.3  & \textbf{99.47} & 99.38 \\
		& (98.83,99.24) & (98.97,99.37) & (98.19,98.77) & (92.76,94.09) & (99.07,99.42) & (99.28,99.55) & (97.76,98.38) & (99.11,99.48) & (99.34,99.61) & (99.21,99.53) \\
		\multirow{2}[0]{*}{NCT-CRC} & 99.48 & 99.45 & 98.95 & 98.19 & \textbf{99.66} & 99.23 & 99.33 & \underline{99.55} & 99.45 & 99.45 \\
		& (99.41,99.55) & (99.38,99.52) & (98.84,99.06) & (98.03,98.35) & (99.61,99.71) & (99.13,99.32) & (99.25,99.41) & (99.48,99.62) & (99.38,99.53) & (99.37,99.52) \\
		\multirow{2}[0]{*}{Osteo} & 92.95 & \textbf{95.72} & 91.12 & 75.81 & 93.05 & 94.97 & 90.68 & 94.47 & 94.58 & \underline{95.16} \\
		& (90.36,95.06) & (93.46,97.61) & (87.96,93.87) & (71.10,80.53) & (90.41,95.29) & (92.51,96.81) & (87.52,93.20) & (91.80,96.39) & (92.02,96.91) & (92.58,97.10) \\
		\multirow{2}[0]{*}{PCAM} & \underline{94.37} & 92.27 & 92.31 & 88.76 & \textbf{95.05} & 92.87 & 91.89 & 92.47 & 93.77 & 93.72 \\
		& (94.14,94.60) & (91.98,92.55) & (92.03,92.59) & (88.40,89.13) & (94.82,95.26) & (92.60,93.13) & (91.61,92.17) & (92.18,92.73) & (93.53,94.03) & (93.48,93.97) \\
		\multirow{2}[0]{*}{Kather\_et\_al\_2016} & 94.8  & 95.99 & 92.97 & 82.07 & \textbf{96.72} & \underline{96.16} & 93.47 & 92.79 & 96.02 & 95.3 \\
		& (93.28,96.20) & (94.60,97.08) & (90.85,94.90) & (78.87,85.20) & (95.56,97.65) & (94.70,97.32) & (91.61,95.11) & (90.83,94.60) & (94.11,97.51) & (93.42,96.81) \\
		\multirow{2}[0]{*}{Kather\_et\_al\_2018} & \textbf{98.01} & 97.33 & 94.99 & 82.66 & 97.2  & 97.43 & 92.5  & 97.16 & \underline{98.01} & 97.55 \\
		& (97.48,98.54) & (96.55,98.07) & (93.89,96.11) & (80.42,84.88) & (96.43,97.93) & (96.74,98.07) & (91.34,93.60) & (96.32,97.88) & (97.28,98.64) & (96.62,98.38) \\
		\multirow{2}[0]{*}{Kather\_2018\_val7k} & 96.6  & \textbf{98.86} & 97.22 & 86.81 & 98.29 & 97.13 & 95.31 & 98.36 & 98.37 & \underline{98.49} \\
		& (95.74,97.45) & (98.31,99.32) & (96.32,98.01) & (84.00,89.65) & (97.75,98.81) & (96.17,98.01) & (93.79,96.77) & (97.71,99.02) & (97.72,98.96) & (97.82,99.02) \\
		\multirow{2}[0]{*}{Skin\_cancer} & 99.19 & 99.19 & 99.03 & 96.1  & \textbf{99.38} & 99.26 & 98.45 & 99.26 & \underline{99.36} & 99.22 \\
		& (99.15,99.24) & (99.14,99.23) & (98.97,99.09) & (95.94,96.26) & (99.35,99.42) & (99.21,99.30) & (98.38,98.51) & (99.21,99.30) & (99.31,99.40) & (99.17,99.26) \\
		\multirow{2}[0]{*}{Tang\_et\_al\_2019} & 83.04 & \textbf{88.19} & 80.29 & 72.32 & 76.09 & 81.25 & 76.62 & \underline{83.68} & 79.6  & 82.35 \\
		& (76.65,88.57) & (82.06,93.34) & (74.57,85.76) & (65.33,78.85) & (68.77,82.96) & (74.46,87.51) & (70.33,82.36) & (77.26,89.22) & (73.25,85.25) & (76.19,87.76) \\
		\multirow{2}[0]{*}{Wong\_et\_al\_2022} & 83.69 & 83.81 & 81.47 & 73.17 & 75.44 & 81.81 & 77.3  & 81.72 & \textbf{85.25} & \underline{84.52} \\
		& (79.91,87.02) & (79.68,87.69) & (77.15,85.68) & (68.32,77.87) & (70.99,79.84) & (77.13,86.13) & (72.66,81.54) & (77.16,85.47) & (81.26,88.82) & (80.50,88.25) \\
		\multirow{2}[1]{*}{Average} & 93.62 & \textbf{94.39} & 92.42 & 84.55 & 92.46 & 92.81 & 90.14 & 92.95 & 94.07 & \underline{94.33} \\
		& (91.76,95.28) & (92.53,95.97) & (90.45,94.26) & (81.70,87.27) & (90.55,94.27) & (90.83,94.68) & (87.91,92.19) & (90.99,94.69) & (92.24,95.70) & (92.53,95.91) \\
		\midrule
		\multicolumn{11}{c}{Endoscopy} \\
		\midrule
		\multirow{2}[1]{*}{Kvasir} & \textbf{98.44} & 98.05 & 97.51 & 89.12 & 94.77 & 97.45 & 95.8  & \underline{98.08} & 97.93 & 97.82 \\
		& (98.20,98.67) & (97.76,98.35) & (97.21,97.82) & (88.32,89.94) & (94.34,95.18) & (97.13,97.74) & (95.43,96.16) & (97.81,98.34) & (97.65,98.21) & (97.51,98.11) \\
		\multirow{2}[0]{*}{WCE} & \textbf{99.65} & \underline{99.21} & 98.05 & 94.66 & 95.3  & 99.01 & 96.4  & 99.01 & 99.17 & 99.19 \\
		& (99.41,99.82) & (98.87,99.50) & (97.49,98.55) & (93.62,95.57) & (94.47,95.96) & (98.70,99.30) & (95.70,97.05) & (98.65,99.35) & (98.85,99.45) & (98.80,99.52) \\
		\multirow{2}[1]{*}{Average} & \textbf{99.05} & \underline{98.63} & 97.78 & 91.89 & 95.03 & 98.23 & 96.1  & 98.54 & 98.55 & 98.5 \\
		& (98.81,99.24) & (98.31,98.92) & (97.35,98.19) & (90.97,92.75) & (94.40,95.57) & (97.91,98.52) & (95.57,96.61) & (98.23,98.84) & (98.25,98.83) & (98.16,98.82) \\
		\bottomrule
	\end{tabular}%
	
	\vspace{2pt}
	\begin{flushright}
		\footnotesize Continued on next page
	\end{flushright}
\end{table}%

\clearpage

\begin{table}[htbp]
	\centering
	\scriptsize
	\setlength{\tabcolsep}{1.5pt}
	\renewcommand{\arraystretch}{1.2}
	
	\begin{tabular}{l|cccccccccc}
		\toprule
		Dataset & BMCA & BiomedCLIP & GenMedClip & MedCLIP & PLIP & PMCCLIP & PUBMEDCLIP & QuiltNet & UniMedCLIP & MMKD-CLIP \\
		\midrule
		\multicolumn{11}{c}{CT} \\
		\midrule
		\multirow{2}[1]{*}{BrainTumorCT} & 98.18 & 98.33 & 97.7  & 87.52 & 93.69 & 96.25 & 94.69 & 98.31 & \textbf{99.47} & \underline{99.42} \\
		& (97.42,98.86) & (97.57,98.93) & (96.73,98.46) & (85.25,89.83) & (91.99,95.16) & (94.83,97.51) & (93.23,96.02) & (97.52,98.96) & (99.14,99.74) & (99.02,99.73) \\
		\multirow{2}[0]{*}{CT\_axial} & \underline{94.31} & 89.71 & 89.05 & 81.34 & 88.88 & \textbf{94.74} & 92.31 & 88.71 & 91.8  & 89.94 \\
		& (93.54,94.96) & (88.33,90.87) & (87.79,90.25) & (79.46,83.12) & (87.88,89.77) & (94.08,95.35) & (91.34,93.19) & (87.44,89.88) & (90.76,92.74) & (88.79,91.03) \\
		\multirow{2}[0]{*}{CT\_coronal} & \underline{87.88} & 83.8  & 85.42 & 77.57 & 82.34 & \textbf{89.28} & 85.18 & 83.16 & 86.94 & 83.28 \\
		& (86.56,89.15) & (81.90,85.44) & (83.77,86.89) & (75.60,79.58) & (80.97,83.66) & (87.96,90.41) & (83.54,86.51) & (81.42,84.89) & (85.37,88.31) & (81.34,84.89) \\
		\multirow{2}[0]{*}{CT\_sagittal} & 87.87 & 84.22 & 86.52 & 75.22 & 84.17 & \textbf{90.80} & 83.5  & 83.76 & \underline{89.21} & 86.55 \\
		& (86.37,89.16) & (82.57,85.66) & (84.96,87.99) & (73.23,77.18) & (82.82,85.38) & (89.75,91.75) & (82.02,84.88) & (82.04,85.40) & (87.70,90.46) & (84.99,87.98) \\
		\multirow{2}[0]{*}{CovidCT3A} & 96.98 & 96.59 & 96.56 & 94.22 & 90.86 & 96.62 & 91.29 & 96.85 & \textbf{97.61} & \underline{97.38} \\
		& (96.84,97.13) & (96.44,96.74) & (96.40,96.70) & (94.03,94.40) & (90.63,91.11) & (96.49,96.75) & (91.08,91.53) & (96.72,96.99) & (97.49,97.73) & (97.25,97.52) \\
		\multirow{2}[0]{*}{Organamnist} & 98.48 & \underline{98.56} & 97.46 & 97.69 & 97.2  & \textbf{98.96} & 98.08 & 98.39 & 98.54 & 98.09 \\
		& (98.41,98.55) & (98.49,98.63) & (97.35,97.56) & (97.59,97.79) & (97.10,97.29) & (98.91,99.01) & (98.00,98.16) & (98.31,98.48) & (98.47,98.61) & (98.00,98.17) \\
		\multirow{2}[1]{*}{Average} & \underline{93.95} & 91.87 & 92.12 & 85.59 & 89.52 & \textbf{94.44} & 90.84 & 91.53 & 93.93 & 92.44 \\
		& (93.19,94.64) & (90.88,92.71) & (91.17,92.98) & (84.19,86.98) & (88.56,90.40) & (93.67,95.13) & (89.87,91.71) & (90.57,92.43) & (93.16,94.60) & (91.56,93.22) \\
		\midrule
		\multicolumn{11}{c}{MRI} \\
		\midrule
		\multirow{2}[1]{*}{BrainTumorMRI} & 99.87 & 99.83 & \underline{99.88} & 90.36 & 96.17 & 98.09 & 94.78 & \textbf{99.89} & 99.83 & 99.88 \\
		& (99.75,99.95) & (99.67,99.95) & (99.75,99.98) & (88.29,92.19) & (95.03,97.20) & (97.35,98.71) & (93.40,95.90) & (99.78,99.97) & (99.65,99.96) & (99.75,99.97) \\
		\multirow{2}[0]{*}{BrainTumorMRI2} & 96.78 & 97.02 & 96.65 & 90.12 & 88.07 & 95.09 & 91    & 97.04 & \textbf{98.28} & \underline{98.01} \\
		& (96.14,97.31) & (96.45,97.56) & (96.05,97.18) & (89.08,91.17) & (86.96,89.19) & (94.29,95.79) & (89.98,91.97) & (96.44,97.58) & (97.86,98.65) & (97.55,98.42) \\
		\multirow{2}[0]{*}{Acl\_mri} & 75.93 & 68.17 & \underline{83.06} & 64.56 & 49.96 & 73.62 & 59.66 & 72.32 & 80.65 & \textbf{94.51} \\
		& (69.05,82.16) & (60.29,75.22) & (77.11,88.41) & (57.00,71.68) & (41.89,57.94) & (65.95,80.40) & (51.23,66.92) & (65.35,78.99) & (74.64,86.24) & (91.12,97.06) \\
		\multirow{2}[1]{*}{Average} & 90.86 & 88.34 & \underline{93.20} & 81.68 & 78.07 & 88.93 & 81.81 & 89.75 & 92.92 & \textbf{97.46} \\
		& (88.32,93.14) & (85.47,90.91) & (90.97,95.19) & (78.12,85.01) & (74.63,81.44) & (85.86,91.63) & (78.20,84.93) & (87.19,92.18) & (90.72,94.95) & (96.14,98.48) \\
		\midrule
		\multicolumn{11}{c}{Ultrasound} \\
		\midrule
		\multirow{2}[1]{*}{BUSBRA} & 62.28 & \textbf{77.00} & 66.7  & 63.25 & 57.66 & 61.08 & 61.58 & 64.29 & 66.33 & \underline{68.77} \\
		& (56.22,68.10) & (71.67,82.09) & (60.63,72.28) & (56.42,69.59) & (51.16,63.53) & (55.11,66.35) & (55.81,67.65) & (58.57,70.18) & (60.63,71.85) & (63.03,74.59) \\
		\multirow{2}[0]{*}{Breast\_us} & 69.71 & \textbf{78.97} & 69.03 & 69.38 & 68.49 & \underline{78.68} & 69.81 & 71.93 & 68.53 & 71.52 \\
		& (59.91,79.33) & (70.09,87.33) & (59.16,78.17) & (59.61,78.66) & (58.92,77.65) & (70.77,86.10) & (60.82,79.05) & (62.28,81.46) & (57.39,78.82) & (60.80,81.07) \\
		\multirow{2}[0]{*}{Breastmnist} & 75.52 & 68.21 & 67.36 & \textbf{83.88} & 72.45 & 75.58 & 70.51 & 75.75 & 73.64 & \underline{76.40} \\
		& (66.42,84.35) & (58.07,78.01) & (58.37,76.88) & (77.14,89.77) & (62.27,82.95) & (66.64,83.65) & (61.27,79.61) & (67.08,83.03) & (63.90,82.13) & (67.61,83.76) \\
		\multirow{2}[1]{*}{Average} & 69.17 & \textbf{74.73} & 67.69 & 72.17 & 66.2  & 71.78 & 67.3  & 70.66 & 69.5  & \underline{72.23} \\
		& (60.85,77.26) & (66.61,82.48) & (59.39,75.78) & (64.39,79.34) & (57.45,74.71) & (64.17,78.70) & (59.30,75.44) & (62.64,78.23) & (60.64,77.60) & (63.81,79.81) \\
		\midrule
		\multicolumn{11}{c}{Dermatology} \\
		\midrule
		\multirow{2}[1]{*}{HAM10000} & \textbf{89.46} & 88.08 & 84.57 & 70.97 & 83.94 & 85.54 & 81.74 & \underline{88.09} & 86.66 & 87.56 \\
		& (88.14,90.57) & (86.46,89.49) & (83.16,86.03) & (68.45,73.51) & (82.26,85.46) & (83.21,87.69) & (79.77,83.52) & (86.54,89.50) & (84.91,88.16) & (86.13,88.91) \\
		\multirow{2}[0]{*}{PADUFES20} & \textbf{82.51} & \underline{80.36} & 78.94 & 59.78 & 68.78 & 69.83 & 70.34 & 79.77 & 71.59 & 76.97 \\
		& (80.48,84.51) & (77.73,83.04) & (76.21,81.64) & (56.00,64.00) & (65.63,71.74) & (65.65,74.24) & (67.17,73.36) & (77.00,82.39) & (67.83,75.15) & (73.78,79.84) \\
		\multirow{2}[1]{*}{Average} & \textbf{85.98} & \underline{84.22} & 81.76 & 65.38 & 76.36 & 77.69 & 76.04 & 83.93 & 79.12 & 82.26 \\
		& (84.31,87.54) & (82.09,86.26) & (79.69,83.83) & (62.23,68.75) & (73.95,78.60) & (74.43,80.96) & (73.47,78.44) & (81.77,85.94) & (76.37,81.65) & (79.96,84.37) \\
		\midrule
		\multicolumn{11}{c}{OCT} \\
		\midrule
		\multirow{2}[1]{*}{OCTMNIST} & 96.29 & 96.08 & 97.23 & 95.73 & 91.88 & 93.78 & 96.37 & \underline{98.14} & 97.41 & \textbf{98.42} \\
		& (95.53,96.91) & (95.30,96.76) & (96.57,97.81) & (94.92,96.49) & (90.68,92.98) & (92.88,94.65) & (95.58,97.07) & (97.59,98.63) & (96.83,97.93) & (97.94,98.81) \\
		\multirow{2}[0]{*}{RetinalOCT} & 97.83 & 97.94 & 97.92 & 95.06 & 95.38 & 98.15 & 95.85 & 98.22 & \textbf{98.80} & \underline{98.73} \\
		& (97.60,98.06) & (97.69,98.19) & (97.67,98.16) & (94.64,95.46) & (95.04,95.71) & (97.92,98.36) & (95.52,96.18) & (98.00,98.43) & (98.60,98.97) & (98.53,98.91) \\
		\multirow{2}[1]{*}{Average} & 97.06 & 97.01 & 97.58 & 95.4  & 93.63 & 95.97 & 96.11 & \underline{98.18} & 98.11 & \textbf{98.57} \\
		& (96.57,97.48) & (96.50,97.47) & (97.12,97.98) & (94.78,95.97) & (92.86,94.34) & (95.40,96.51) & (95.55,96.63) & (97.80,98.53) & (97.72,98.45) & (98.24,98.86) \\
		\bottomrule
	\end{tabular}%
\end{table}%

\newpage

\clearpage

\newgeometry{left=0.7in,right=0.7in,top=0.6in,bottom=0.6in}

\begin{table}[htbp]
	\centering
	\caption{Linear probe classification task results across 9 different modalities (100\% training data). \textbf{Bold} indicates best result, \underline{underline} indicates second best result. 95\% CI is included in parentheses.}
	\vspace{0.3cm}
	\label{tab:linear_probe_100_results}%
	\scriptsize
	\setlength{\tabcolsep}{1.5pt}
	\renewcommand{\arraystretch}{1.2}
	
\begin{tabular}{l|cccccccccc}
		\toprule
		Dataset & BMCA & BiomedCLIP & GenMedClip & MedCLIP & PLIP & PMCCLIP & PUBMEDCLIP & QuiltNet & UniMedCLIP & MMKD-CLIP \\
		\midrule
		\multicolumn{11}{c}{X-Ray} \\
		\midrule
		\multirow{2}[1]{*}{CovidCXR4} & 64.31 & 60.31 & 63.41 & 63.14 & 59.89 & 67.44 & 62.33 & 67.94 & \underline{69.82} & \textbf{74.40} \\
		& (63.01,65.46) & (59.09,61.52) & (62.16,64.58) & (61.99,64.35) & (58.71,61.06) & (66.32,68.54) & (61.15,63.46) & (66.70,69.03) & (68.69,70.87) & (73.43,75.48) \\
		\multirow{2}[0]{*}{DDSM} & \textbf{98.64} & 95.75 & 96.79 & 93.82 & 96.14 & \underline{97.86} & 96.02 & 96.76 & 97.52 & 97.44 \\
		& (98.34,98.88) & (95.21,96.30) & (96.24,97.25) & (93.15,94.47) & (95.61,96.61) & (97.46,98.20) & (95.53,96.51) & (96.25,97.26) & (97.13,97.93) & (97.06,97.79) \\
		\multirow{2}[0]{*}{NLMTB} & 99.81 & \textbf{99.99} & 99.91 & 97.88 & 99.73 & 99.95 & 98.17 & 99.89 & \underline{99.98} & 99.98 \\
		& (99.53,99.97) & (99.97,100.00) & (99.81,99.98) & (96.63,98.85) & (99.46,99.93) & (99.87,100.00) & (97.32,98.97) & (99.76,99.98) & (99.95,100.00) & (99.95,100.00) \\
\bottomrule
	\end{tabular}%
	
	\vspace{2pt}
	\begin{flushright}
		\footnotesize Continued on next page
	\end{flushright}
\end{table}%

\clearpage

\begin{table}[htbp]
	\centering
	\scriptsize
	\setlength{\tabcolsep}{1.5pt}
	\renewcommand{\arraystretch}{1.2}
	
	\begin{tabular}{l|cccccccccc}
		\toprule
		Dataset & BMCA & BiomedCLIP & GenMedClip & MedCLIP & PLIP & PMCCLIP & PUBMEDCLIP & QuiltNet & UniMedCLIP & MMKD-CLIP \\
		\midrule
		\multicolumn{11}{c}{CT} \\
		\midrule
		\multirow{2}[1]{*}{BrainTumorCT} & 99.27 & 99.4  & 99.08 & 97.14 & 97.5  & 98.2  & 96.63 & 99.59 & \underline{99.81} & \textbf{99.83} \\
		& (98.79,99.65) & (98.94,99.72) & (98.53,99.51) & (96.17,98.01) & (96.57,98.26) & (97.37,98.88) & (95.53,97.59) & (99.32,99.80) & (99.66,99.92) & (99.68,99.94) \\
		\multirow{2}[0]{*}{CT\_axial} & 98.02 & \underline{98.19} & 96.06 & 93.52 & 95.43 & \textbf{98.22} & 97.03 & 97.91 & 98.01 & 97.36 \\
		& (97.69,98.32) & (97.75,98.60) & (95.35,96.71) & (92.42,94.49) & (94.81,95.99) & (97.90,98.51) & (96.53,97.48) & (97.44,98.31) & (97.55,98.40) & (96.81,97.88) \\
		\multirow{2}[0]{*}{CT\_coronal} & 95.7  & 96.02 & 92.27 & 88.24 & 91.58 & \textbf{96.23} & 92.79 & 95.65 & \underline{96.08} & 94.45 \\
		& (94.97,96.37) & (95.15,96.78) & (91.12,93.28) & (86.78,89.62) & (90.55,92.50) & (95.59,96.78) & (91.85,93.66) & (94.81,96.40) & (95.34,96.73) & (93.46,95.34) \\
\bottomrule
	\end{tabular}%
\end{table}%

\begin{table}[h]
	\centering
	
	\caption{Cross-modal retrieval task result on Bookset.
		\label{cross1}
		 \textbf{Bold} indicates best result, \underline{underline} indicates second best result. 95\% CI is included in parentheses.}

	\begin{tabular}{c|ccc}
		\hline
		\multicolumn{4}{c}{Image-to-Text} \\
		\hline
		Model & R@1 & R@10 & R@50 \\
		\hline
		BMCA  & 3.63 (3.09, 4.19) & 12.79 (11.85, 13.82) & 24.89 (23.58, 26.21) \\
		BioMedCLIP & 5.78 (5.13, 6.53) & 20.68 (19.46, 21.90) & 38.00 (36.49, 39.44) \\
		GenMedClip & 6.86 (6.14, 7.61) & 25.25 (23.93, 26.58) & \textbf{50.52} (49.02, 52.08) \\
		MedCLIP & 0.02 (0.00, 0.07) & 0.31 (0.16, 0.47) & 1.03 (0.75, 1.33) \\
		PLIP  & 1.36 (1.01, 1.73) & 8.22 (7.35, 9.09) & 21.09 (19.91, 22.27) \\
		PMCCLIP & 2.84 (2.37, 3.33) & 13.08 (12.06, 14.12) & 27.71 (26.35, 29.02) \\
		PUBMEDCLIP & 0.07 (0.00, 0.16) & 0.61 (0.37, 0.84) & 2.03 (1.62, 2.48) \\
		QuiltNet & \underline{8.17} (7.35, 8.99) & 25.73 (24.45, 27.07) & 49.69 (48.22, 51.24) \\
		UniMedCLIP & 7.18 (6.44, 7.99) & \underline{26.43} (25.18, 27.75) & \underline{49.63} (48.06, 51.20) \\
		MMKD-CLIP & \textbf{8.45} (7.66, 9.23) & \textbf{26.61} (25.41, 27.99) & 45.54 (44.07, 47.05) \\
		\hline
		\multicolumn{4}{c}{Text-to-Image} \\
		\hline
		Model& R@1 & R@10 & R@50 \\
		\hline
		BMCA  & 3.52 (3.02, 4.08) & 13.35 (12.36, 14.38) & 28.92 (27.56, 30.16) \\
		BioMedCLIP & 5.87 (5.18, 6.60) & 21.74 (20.51, 22.86) & 43.52 (41.99, 44.97) \\
		GenMedClip & 6.84 (6.09, 7.63) & 25.22 (23.93, 26.49) & 49.12 (47.61, 50.73) \\
		MedCLIP & 0.02 (0.00, 0.07) & 0.21 (0.09, 0.35) & 1.21 (0.91, 1.55) \\
		PLIP  & 1.26 (0.94, 1.62) & 7.90 (7.12, 8.69) & 22.33 (21.08, 23.68) \\
		PMCCLIP & 2.89 (2.39, 3.42) & 11.86 (10.87, 12.83) & 27.41 (26.06, 28.78) \\
		PUBMEDCLIP & 0.09 (0.02, 0.19) & 0.84 (0.59, 1.10) & 3.19 (2.72, 3.70) \\
		QuiltNet & \underline{7.37} (6.58, 8.17) & 25.80 (24.47, 27.14) & 49.22 (47.75, 50.75) \\
		UniMedCLIP & 6.93 (6.21, 7.68) & \underline{26.68} (25.36, 28.06) & \underline{52.36} (50.82, 53.79) \\
		MMKD-CLIP & \textbf{7.81} (7.05, 8.55) & \textbf{27.49} (26.18, 28.81) & \textbf{52.95} (51.52, 54.38) \\
		\hline
	\end{tabular}
\end{table}

\begin{table}[h]
	\centering
	
	\caption{Cross-modal retrieval task result on MedTrinity-25M-subset. \textbf{Bold} indicates best result, \label{cross2}
		\underline{underline} indicates second best result. 95\% CI is included in parentheses.}

	\begin{tabular}{c|ccc}
		\hline
		\multicolumn{4}{c}{Image-to-Text} \\
		\hline
		Model & R@1 & R@10 & R@50 \\
		\hline
		BMCA         & \textbf{0.91} (0.66, 1.19) & \underline{4.19} (3.61, 4.78) & 14.21 (13.24, 15.19) \\
		BioMedCLIP   & \underline{0.55} (0.34, 0.79) & 4.05 (3.48, 4.63) & \underline{15.06} (14.02, 16.07) \\
		GenMedClip   & 0.49 (0.30, 0.72) & 3.44 (2.91, 3.97) & 14.66 (13.66, 15.70) \\
		MedCLIP      & 0.00 (0.00, 0.00) & 0.34 (0.19, 0.51) & 1.14 (0.85, 1.47) \\
		PLIP         & 0.23 (0.11, 0.38) & 1.83 (1.49, 2.23) & 8.59 (7.82, 9.41) \\
		PMCCLIP      & 0.49 (0.30, 0.70) & 3.22 (2.74, 3.74) & 12.66 (11.75, 13.64) \\
		PUBMEDCLIP   & 0.25 (0.13, 0.43) & 1.63 (1.28, 2.02) & 7.49 (6.78, 8.22) \\
		QuiltNet     & 0.38 (0.21, 0.57) & 3.85 (3.31, 4.40) & 13.73 (12.75, 14.75) \\
		UniMedCLIP   & 0.30 (0.15, 0.47) & 3.28 (2.80, 3.80) & 13.21 (12.28, 14.17) \\
		MMKD-CLIP    & 0.38 (0.23, 0.57) & \textbf{4.20} (3.68, 4.80) & \textbf{16.63} (15.62, 17.72) \\
		\hline
		\multicolumn{4}{c}{Text-to-Image} \\
		\hline
		Model& R@1 & R@10 & R@50 \\
		\hline
		BMCA & 0.47 (0.28, 0.66) & 3.59 (3.10, 4.14) & \underline{13.67} (12.73, 14.66) \\
		BioMedCLIP & 0.51 (0.32, 0.74) & 3.19 (2.68, 3.72) & 13.64 (12.64, 14.60) \\
		GenMedClip & 0.29 (0.15, 0.45) & 2.95 (2.49, 3.44) & 10.97 (10.11, 11.92) \\
		MedCLIP & 0.02 (0.00, 0.06) & 0.24 (0.11, 0.38) & 1.06 (0.79, 1.36) \\
		PLIP & 0.17 (0.06, 0.30) & 1.72 (1.36, 2.10) & 6.95 (6.27, 7.71) \\
		PMCCLIP & 0.43 (0.26, 0.62) & 3.21 (2.72, 3.76) & 11.60 (10.67, 12.52) \\
		PUBMEDCLIP & 0.25 (0.13, 0.38) & 2.01 (1.64, 2.42) & 8.06 (7.27, 8.82) \\
		QuiltNet & 0.51 (0.34, 0.72) & 3.43 (2.95, 3.93) & 13.15 (12.15, 14.09) \\
		UniMedCLIP & \underline{0.51} (0.30, 0.72) & \underline{3.74} (3.23, 4.27) & 13.50 (12.60, 14.51) \\
		MMKD-CLIP & \textbf{0.68} (0.47, 0.91) & \textbf{3.76} (3.21, 4.27) & \textbf{15.16} (14.15, 16.11) \\
		\hline
	\end{tabular}
\end{table}

\begin{table}[htbp]
	\centering
	\caption{Visual question answering task results on SLAKE and VQA-RAD dataset. \textbf{Bold} indicates best result, \underline{underline} indicates second best result. 95\% CI is included in parentheses.}
	\label{tab:vqa}%
	\scriptsize
	\setlength{\tabcolsep}{1.5pt}
	\renewcommand{\arraystretch}{1.2}
	\adjustbox{width=\textwidth,center}{
		\begin{tabular}{c|cccccccccc}
			\toprule
			Metric & BMCA & BiomedCLIP & GenMedClip & MedCLIP & PLIP & PMCCLIP & PUBMEDCLIP & QuiltNet & UniMedCLIP & MMKD-CLIP \\
			\midrule
			\multicolumn{11}{c}{SLAKE} \\
			\midrule
			\multirow{2}[1]{*}{Closed Accuracy} & 82.21 & 82.69 & 81.73 & 80.96 & 82.17 & \underline{82.93} & 81.97 & 81.01 & 81.73 & \textbf{83.65} \\
			& (78.85,85.58) & (79.09,86.06) & (77.88,85.58) & (77.35,84.82) & (78.55,85.55) & (79.33,86.54) & (78.36,85.34) & (77.16,84.86) & (78.13,85.34) & (80.04,87.02) \\
			\multirow{2}[1]{*}{Open Accuracy} & 76.12 & 75.66 & 76.28 & \textbf{78.17} & 77.55 & 77.83 & \underline{78.14} & 77.98 & 76.74 & 77.98 \\
			& (73.02,79.38) & (72.25,79.22) & (72.87,79.53) & (74.77,81.42) & (74.46,80.65) & (74.57,80.93) & (74.88,81.40) & (74.57,81.24) & (73.64,80.00) & (74.73,81.09) \\
			\multirow{2}[1]{*}{Overall Accuracy} & 78.51 & 78.42 & 78.42 & 79.26 & 79.36 & \underline{79.83} & 79.64 & 79.17 & 78.70 & \textbf{80.21} \\
			& (76.06,80.78) & (76.06,80.87) & (75.97,80.96) & (76.91,81.72) & (77.00,81.81) & (77.47,82.28) & (77.10,82.00) & (76.81,81.53) & (76.06,81.06) & (77.95,82.56) \\
\end{tabular}}%
\end{table}%

\begin{table}[htbp]
	\centering
	\caption{Cancer diagnosis task result on BRACS, NSCLC and Camelyon dataset. \textbf{Bold} indicates best result, \underline{underline} indicates second best result. 95\% CI is included in parentheses.}
	\label{cd}%
	\scriptsize
	\setlength{\tabcolsep}{1.5pt}
	\renewcommand{\arraystretch}{1.2}
	\adjustbox{width=\textwidth,center}{
	\begin{tabular}{c|cccccccccc}
		\toprule
		Metric & BMCA & BiomedCLIP & GenMedClip & MedCLIP & PLIP & PMCCLIP & PUBMEDCLIP & QuiltNet & UniMedCLIP & MMKD-CLIP \\
		\midrule
		\multirow{2}[1]{*}{BRACS-3} 
		& 80.00 & 84.06 & 85.31 & 73.15 & 79.26 & 83.25 & 75.69 & 85.89 & \underline{86.44} & \textbf{87.42} \\
		& (74.41,84.88) & (79.62,88.63) & (80.59,89.78) & (67.36,78.62) & (74.09,84.17) & (78.55,87.53) & (69.74,81.52) & (81.34,90.06) & (81.74,90.76) & (82.75,91.55) \\
		\midrule
		\multirow{2}[1]{*}{BRACS-7} 
		& 81.47 & 84.13 & 81.72 & 73.12 & 82.03 & 83.70 & 77.53 & 86.35 & \underline{88.07} & \textbf{88.13} \\
		& (77.71,85.06) & (80.71,87.33) & (78.38,85.02) & (68.75,77.18) & (79.14,85.15) & (80.21,87.02) & (73.67,81.17) & (83.37,89.22) & (85.50,90.42) & (85.41,90.50) \\
		\multirow{2}[1]{*}{NSCLC} 
		& 93.28 & \underline{97.05} & 94.34 & 85.95 & 93.73 & 94.27 & 89.42 & 95.46 & 95.72 & \textbf{97.28} \\
		& (89.68,96.38) & (94.70,98.74) & (90.90,97.10) & (80.78,90.64) & (90.39,96.47) & (90.86,96.90) & (84.44,93.69) & (92.47,97.76) & (92.91,98.02) & (94.94,99.02) \\
		\multirow{2}[1]{*}{Camelyon} 
		& 88.34 & 83.14 & \textbf{90.73} & 79.95 & 88.68 & 85.11 & 81.05 & 88.76 & \underline{90.03} & 89.17 \\
		& (81.51,94.43) & (75.20,90.67) & (84.62,96.05) & (71.29,87.80) & (82.25,94.04) & (77.39,91.62) & (71.77,89.55) & (81.27,95.06) & (83.23,95.75) & (82.71,94.79) \\
		\midrule

	\end{tabular}}%
\end{table}%

\begin{table}[h]
	\centering
	\caption{Survival prediction task result using PLIP. Mean±std is presented.}
	\label{sp1}
	\adjustbox{width=\textwidth,center}{
		\begin{tabular}{c|ccc|ccc|ccc|ccc}
			\hline
			\multirow{2}[1]{*}{Dataset}  & \multicolumn{3}{c}{C-Index} & \multicolumn{3}{c}{C-Index$_{td}$} & \multicolumn{3}{c}{IBS} & \multicolumn{3}{c}{INBLL} \\
			\cline{2-13}
			& I+T & I & T & I+T & I & T & I+T & I & T & I+T & I & T \\
			\hline
			BLCA & \shortstack{61.24\\(±4.80)} & \shortstack{57.70\\(±2.50)} & \shortstack{58.17\\(±4.92)} & \shortstack{61.28\\(±4.91)} & \shortstack{57.44\\(±2.26)} & \shortstack{57.97\\(±5.12)} & \shortstack{23.48\\(±3.49)} & \shortstack{37.50\\(±2.31)} & \shortstack{33.99\\(±5.18)} & \shortstack{68.79\\(±12.02)} & \shortstack{116.76\\(±12.31)} & \shortstack{108.34\\(±26.16)} \\
			BRCA & \shortstack{74.20\\(±3.62)} & \shortstack{59.44\\(±2.78)} & \shortstack{72.56\\(±4.28)} & \shortstack{74.16\\(±3.35)} & \shortstack{57.71\\(±3.66)} & \shortstack{72.69\\(±4.13)} & \shortstack{25.89\\(±6.08)} & \shortstack{23.91\\(±6.94)} & \shortstack{19.43\\(±2.33)} & \shortstack{83.65\\(±24.09)} & \shortstack{71.52\\(±24.23)} & \shortstack{57.93\\(±7.54)} \\
			CESC & \shortstack{73.81\\(±6.03)} & \shortstack{68.21\\(±8.09)} & \shortstack{65.57\\(±8.63)} & \shortstack{73.15\\(±6.54)} & \shortstack{68.54\\(±8.80)} & \shortstack{65.76\\(±9.94)} & \shortstack{30.24\\(±6.68)} & \shortstack{21.62\\(±4.15)} & \shortstack{25.79\\(±5.76)} & \shortstack{99.84\\(±27.65)} & \shortstack{64.76\\(±12.93)} & \shortstack{84.09\\(±34.09)} \\
			HNSC & \shortstack{62.96\\(±3.63)} & \shortstack{60.72\\(±6.05)} & \shortstack{59.47\\(±3.58)} & \shortstack{62.40\\(±3.43)} & \shortstack{60.84\\(±6.16)} & \shortstack{59.43\\(±3.78)} & \shortstack{31.27\\(±12.17)} & \shortstack{32.84\\(±1.52)} & \shortstack{28.20\\(±3.63)} & \shortstack{163.91\\(±165.47)} & \shortstack{99.13\\(±11.01)} & \shortstack{85.25\\(±15.30)} \\
\hline
		\end{tabular}
	}
\end{table}

\clearpage
\begin{table}[h]
	\centering
	\caption{Survival prediction task result using MMKD-CLIP. Mean±std is presented.}
	\label{sp2}
	\adjustbox{width=\textwidth,center}{
		\begin{tabular}{c|ccc|ccc|ccc|ccc}
			\hline
			\multirow{2}[1]{*}{Dataset}  & \multicolumn{3}{c}{C-Index} & \multicolumn{3}{c}{C-Index$_{td}$} & \multicolumn{3}{c}{IBS} & \multicolumn{3}{c}{INBLL} \\
			\cline{2-13}
			& I+T & I & T & I+T & I & T & I+T & I & T & I+T & I & T \\
        \hline
		BLCA & \shortstack{63.83\\(±3.58)} & \shortstack{59.97\\(±4.70)} & \shortstack{58.68\\(±8.51)} & \shortstack{63.81\\(±3.62)} & \shortstack{59.71\\(±4.61)} & \shortstack{58.09\\(±8.99)} & \shortstack{24.05\\(±2.09)} & \shortstack{38.44\\(±5.42)} & \shortstack{36.92\\(±4.85)} & \shortstack{68.48\\(±5.69)} & \shortstack{116.86\\(±21.09)} & \shortstack{130.15\\(±20.41)} \\

		BRCA & \shortstack{73.15\\(±3.65)} & \shortstack{62.61\\(±5.60)} & \shortstack{71.94\\(±5.48)} & \shortstack{72.65\\(±4.48)} & \shortstack{62.39\\(±5.34)} & \shortstack{71.20\\(±4.72)} & \shortstack{30.45\\(±11.52)} & \shortstack{23.04\\(±2.23)} & \shortstack{18.68\\(±3.55)} & \shortstack{97.65\\(±48.98)} & \shortstack{84.05\\(±28.86)} & \shortstack{58.47\\(±13.36)} \\

		CESC & \shortstack{76.52\\(±4.78)} & \shortstack{66.43\\(±9.42)} & \shortstack{67.12\\(±10.51)} & \shortstack{77.12\\(±6.60)} & \shortstack{65.34\\(±9.66)} & \shortstack{64.75\\(±9.01)} & \shortstack{21.84\\(±3.98)} & \shortstack{19.50\\(±2.09)} & \shortstack{26.25\\(±1.45)} & \shortstack{125.41\\(±131.73)} & \shortstack{58.39\\(±5.75)} & \shortstack{82.61\\(±5.67)} \\

		HNSC & \shortstack{66.99\\(±4.32)} & \shortstack{61.25\\(±4.68)} & \shortstack{64.08\\(±2.15)} & \shortstack{67.29\\(±4.19)} & \shortstack{61.54\\(±4.62)} & \shortstack{62.36\\(±2.56)} & \shortstack{23.08\\(±2.55)} & \shortstack{29.89\\(±3.21)} & \shortstack{30.66\\(±2.02)} & \shortstack{67.01\\(±8.69)} & \shortstack{87.32\\(±11.13)} & \shortstack{96.36\\(±7.25)} \\
\hline
		\end{tabular}
	}
\end{table}

\begin{table}[h]
	\centering
	\caption{Ablation study based on zero-shot image classification. \textbf{Bold} indicates best result, \underline{underline} indicates second best result. 95\% CI is included in parentheses.}
	\label{ab}
	\begin{tabular}{c|cccc}
		\hline
		& Only Pretraining & MMKD-CLIP w/o FD Loss & MMKD-CLIP w/o ICL Loss & MMKD-CLIP \\ \hline
		Overall Dataset & 84.26 (81.89, 86.45) & \underline{84.69} (82.41, 86.81) & 84.64 (82.31, 86.76) & \textbf{84.78} (82.51, 86.91) \\ 
		X-Ray            & 82.65 (80.81, 84.32) & \underline{82.99} (81.23, 84.57) & 82.52 (80.64, 84.26) & \textbf{83.45} (81.78, 84.94) \\ 
		Pathology        & 85.63 (83.34, 87.73) & \underline{86.43} (84.29, 88.43) & 86.41 (84.19, 88.44) & \textbf{86.61} (84.49, 88.59) \\ 
		CT               & 84.53 (83.33, 85.69) & \underline{85.59} (84.51, 86.61) & 85.06 (83.92, 86.14) & \textbf{85.78} (84.72, 86.78) \\ 
		MRI              & 88.90 (86.17, 91.35) & 89.58 (86.96, 91.87) & \underline{89.83} (87.21, 92.11) & \textbf{90.09} (87.61, 92.30) \\ \hline
	\end{tabular}
\end{table}

\end{document}